\documentclass[10pt,twocolumn,letterpaper]{article}

\usepackage[applications]{wacv}      
\usepackage{graphicx}
\usepackage{comment}
\usepackage{booktabs}

\usepackage[accsupp]{axessibility}  

\usepackage{multirow}
\usepackage{amsmath}
\usepackage{amssymb}
\usepackage{float}
\usepackage[table]{xcolor}
\usepackage{colortbl}
\usepackage{markdown}
\usepackage{makecell}
\usepackage{subcaption}
\usepackage[export]{adjustbox}

\usepackage[textsize=tiny]{todonotes}
\setuptodonotes{inline, bordercolor=white, backgroundcolor=yellow!21}

\usepackage{enumitem} 

\usepackage{pifont}

\newcommand{\lname}{CrossLiDAR}
\newcommand{\name}{\lname}
\newcommand{\lidar}{LiDAR}
\newcolumntype{H}{@{}>{\setbox0=\hbox\bgroup}c<{\egroup}}
\newcommand{\titleicon}{%
  \raisebox{-0.15em}{%
    \includegraphics[height=1.25em]{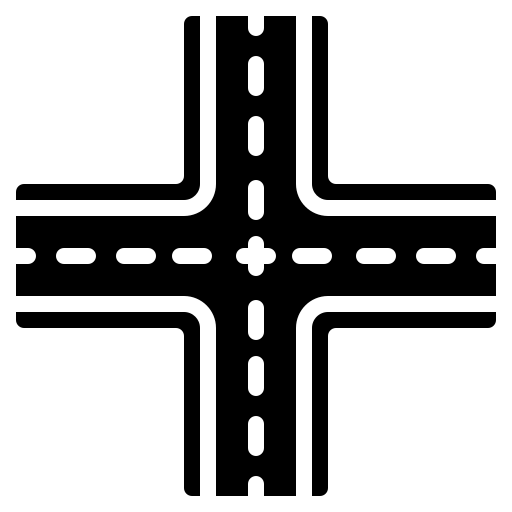}%
  }%
}

\newcommand{\red}[1]{{\color{black}#1}}

\definecolor{wacvblue}{rgb}{0.21,0.49,0.74}
\usepackage[pagebackref,breaklinks,colorlinks,allcolors=wacvblue]{hyperref}

\def\wacvPaperID{1971} 
\def\confName{WACV}
\def\confYear{2027}

\title{%
  \texorpdfstring{\titleicon\ \lname{}: Controllable Multi-Sensor LiDAR Sequence Generation}%
  {\lname{}: Controllable Multi-Sensor LiDAR Sequence Generation}%
}

\author{%
  \parbox{\textwidth}{\centering
    {Khanh Nguyen\textsuperscript{$\dagger$}}\thanks{Equal contribution.}\quad
    {Thomas Walker\textsuperscript{$\P$}\footnotemark[1]}\quad
    {Hichem Telli\textsuperscript{$\P$}}\footnotemark[1]\;\thanks{The corresponding author: tellihicham@univ-biskra.dz}\\[0.2cm]
    {Carlos de la Torre-Ortiz\textsuperscript{$\P$}}\footnotemark[1]\quad
    {Stefano Fiorini\textsuperscript{$\P$}}\footnotemark[1]\quad
    {Salah Eddine Bekhouche\textsuperscript{$\S$}}\\[2.5pt]
    {Mohammed-En-Nadhir Zighem\textsuperscript{$\P$}}\quad
    {Wassim Swaileh\textsuperscript{$\P$}}\quad
    {Tinghuai Wang\textsuperscript{$\P$}}\\[2.5pt]
    {\rm
      \textsuperscript{$\dagger$}Computer Vision Center, Universitat Aut\`onoma de Barcelona, Spain \\
      \textsuperscript{$\S$} University of the Basque Country, Spain  \quad 
      \textsuperscript{$\P$} Independent Researcher\\[2pt]
    }
  }%
}

\begin{document}
\maketitle

\begin{abstract}  
High-fidelity LiDAR simulation remains a bottleneck for scalable development and validation of autonomous systems.
Existing generators can synthesize realistic scans, but they are typically constrained to a single sensor configuration, offer limited control over structure and geometry, and drift in long-horizon 4D generation.
We present \lname, a unified transformer-based framework capable of generating controllable and spatio-temporally consistent multi-sensor LiDAR sequences.
Our approach leverages a rectified flow objective with a Multi-Modal Diffusion Transformer (MM-DiT) backbone, within a compact latent space of range images.
To support multiple LiDAR sensors, an Inter-sensor Module enables 
interaction across sensors, while a ControlNet-style branch provides structural conditioning (e.g., BEV priors or 3D bounding boxes).
For modeling the temporal consistency, \lname{} generates short $N$-frame  sequences and extends them autoregressively to improve long-horizon coherence.
Our model significantly outperforms state-of-the-art baselines, achieving an average improvement of 29.56\% in JSD and 15.36\% in MMD evaluated on nuScenes and the Waymo Dataset.
\lname{} bridges the gap to modern autonomous-driving simulation by enabling calibration-consistent multi-sensor LiDAR rollouts over long horizons for controllable, scenario-based testing.
\end{abstract}

\begin{figure}[htb!]
    \centering
    \includegraphics[width=\linewidth]{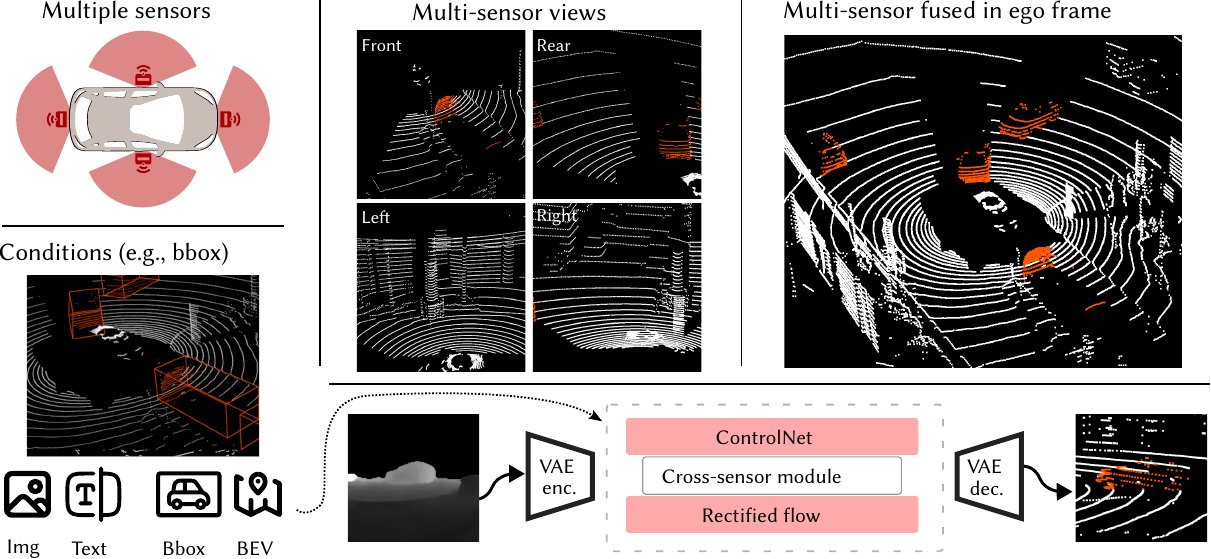}
    \caption{%
        \textbf{\name{}} jointly generates synchronized LiDAR sequences for multi-sensor setups.
        \textbf{Top:} given multiple sensors mounted around the ego vehicle, \name{} synthesizes geometrically aligned per-sensor views (front/rear/left/right) that can be fused into a consistent ego-frame point cloud, while supporting structural conditions such as 3D bounding boxes.
        \textbf{Bottom:} our pipeline operates in a VAE-compressed range-image latent space; a rectified-flow module integrates an Inter-sensor Module for sensor-aware generation and
        a ControlNet branch for structural conditioning using BEV maps with projected 3D bounding boxes,
        yielding controllable multi-sensor LiDAR sequences.
    }%
    \label{fig:teaser}
\end{figure}

\section{Introduction}%
\label{sec:intro}

Autonomous systems have become fundamental in modern robotics and industrial innovation. 
These systems, ranging from self-driving vehicles~\cite{shi2020pv,yin2021center} to surveying drones~\cite{zhang2014loam, resop2019drone}, often rely heavily on \lidar{} to navigate complex 3D environments.
By providing high-fidelity geometric data, LiDAR enables precise perception; however, it is constrained by the high cost of the sensors and logistical complexity of real-world data collection.
Furthermore, capturing critical ``corner cases''~\cite{zhao2025corner}, such as high-impact collisions or extreme meteorological conditions, remains inherently dangerous and rare in physical datasets~\cite{wang2024survey}. 

Traditional simulation toolkits~\cite{dosovitskiy2017carla,manivasagam2020lidarsim,li2022pcgen,montalvo2024spin} require labor-intensive manual scene modeling or existing real-world scans, limiting their scalability.
Therefore, research has shifted toward generative models to automate this process, despite the inherent challenges of \lidar{} data: high sparsity, non-uniformity, and intrinsic geometric structure. Early generators built on Generative Adversarial Networks (GANs)~\cite{goodfellow2020generative,caccia2019deep} and Variational Autoencoders (VAEs)~\cite{kingma2013auto} often produced ``fuzzy'' geometries and detail loss.

While score-based models like LiDARGen~\cite{zyrianov2022learning} improved fidelity, they suffered from slow sampling speed and poor long-range performance. More recently, UltraLiDAR~\cite{xiong2023ultralidar} utilized VQ-VAEs~\cite{van2017neural} in the Bird’s-Eye-View (BEV) to enhance realism. However, processing sparse BEV representations often results in significant computational waste on empty voxels, leading to inefficient generation. 
Alternatively, RangeLDM~\cite{hu2024rangeldm} and LiDMs~\cite{ran2024towards} leverage Latent Diffusion Models (LDMs)~\cite{rombach2022high} in the range-view to improve speed. However, these frameworks are primarily restricted to fixed sensor architectures and input dimensions, preventing cross-sensor scalability. Furthermore, they lack fine-grained structural guidance (e.g., ControlNet~\cite{zhang2023adding}), rendering them impractical for real-world usage, where side \lidar{} proves to be more affordable.
More recently, LiDARCrafter~\cite{liang2026lidarcrafter} extended LiDAR generation to 4D scene modeling by employing a tri-branch diffusion framework that generates temporally ordered 3D LiDAR sequences, with time serving as the fourth dimension.
While it achieves impressive temporal consistency, the model remains limited 
to single-LiDAR sensor configurations.
In modern autonomous driving architectures, however, 
configurations with multiple LiDAR sensors
have become the industry standard~\cite{sun2020scalability}.
This shift is driven by the move toward cost-effective sensor redundancy, where multiple low-profile units augment or replace a single high-performance rooftop scanner to eliminate blind spots.
Unlike single-sensor setups, these distributed arrays introduce non-trivial geometric disparities and complex overlapping fields of view.
Existing 4D models typically lack the architectural mechanisms to perform cross-sensor fusion.

To address these limitations, we introduce \lname{}, a rectified flow model~\cite{liu2022flow} that departs from conventional U-Net-based architectures~\cite{ronneberger2015u} in favor of a transformer-based design: the Multi-Modal Diffusion Transformer (MM-DiT)~\cite{esser2024scaling}, see Figure~\ref{fig:teaser}.
By leveraging the superior modeling capacity and flexibility of transformers~\cite{vaswani2017attention}, \lname{} captures the intricate geometric dependencies of LiDAR sequences with significantly higher fidelity than previous methods.
%
Central to our framework is a specialized \emph{Inter-sensor Module} designed to decouple the generation process from specific hardware constraints. This module enables our model to handle multiple LiDAR sensors in parallel, modeling their interactions and ensuring geometric consistency, without the need to design separate models for each sensor configuration. This is particularly effective for the increasingly common integration of low-cost side sensors.
To provide more fine-grained geometric consistency, we incorporate a ControlNet-style branch, which allows for precise scene manipulation through various priors, including 3D bounding boxes and BEV maps~\cite{qian2025review}.
Finally, to achieve high-fidelity temporal LiDAR generation, \lname{} introduces a \emph{LiDAR Sequence Module}.
Unlike existing methods that focus solely on single-frame generation, our model performs multi-frame generation (e.g., 5-frame sequence). 
These short-term sequences can then be iteratively extended through an autoregressive mechanism, enabling the synthesis of long-horizon~\cite{mousakhan2025orbis}, temporally stable LiDAR sequences. 
To the best of our knowledge, \lname{} serves as the first truly controllable
and multi-sensor pipeline for LiDAR simulation.

We evaluate~\name{} on two large-scale datasets: Waymo Open Dataset~\cite{sun2020scalability},  which provides the multi-sensor data from peripheral side LiDARs, and nuScenes~\cite{caesar2020nuscenes}. Compared to existing baselines, in the image-based setup, \name{} achieves an average improvement of 29.56\% in JSD and 15.36\% in MMD, with particularly superior performance in the multi-sensor configurations.

\noindent{\bf Our main contributions are as follows:}
\begin{itemize}
  \item We introduce \lname{}, a unified diffusion transformer that enables multi-sensor LiDAR generation via an \emph{Inter-sensor Module}, and supports controllable synthesis from geometric cues through a \emph{ControlNet} branch.
  \item We present a $4$D generation strategy that jointly synthesizes short multi-frame sequences with a \emph{LiDAR Sequence Module} and extends them autoregressively, reducing temporal drift and improving long-horizon consistency.
\item We show state-of-the-art (SOTA) performance on Waymo (multi-sensor) and nuScenes (single-sensor) for both single-frame generation and LiDAR sequences.
\end{itemize}

\section{Related Work and Background}

\noindent \textbf{Multisensor \lidar{} Generation.}
Generative models for autonomous driving assume a single \lidar{} sensor, whereas real vehicles often deploy a heterogeneous suite (e.g., top-mounted long-range and multiple side-mounted sensors) to capture overlapping perspectives. Multi-sensor data generation has been extensively explored in the camera domain, where recent works target the consistent synthesis of multi-camera surround-view~\cite{li2024drivingdiffusion,yan2024streetcrafter,fu2024drivegenvlm,wang2024freevs}. In contrast, to our knowledge, no prior generative model explicitly leverages multiple LiDAR sensors simultaneously.
A key technical challenge is that multi-LiDAR point clouds must be carefully aligned: sensors at different positions yield point clouds in distinct coordinate frames. Unless corrected, naively merging multi-LiDAR data leads to distributional artifacts and geometric inconsistencies~\cite{hu2024rangeldm,wu2025multi}. Existing LiDAR generators, including LiDARGen~\cite{zyrianov2022learning}, RangeLDM~\cite{hu2024rangeldm}, and LiDARCrafter~\cite{liang2026lidarcrafter}, operate on a single aggregated point cloud per scene without explicitly addressing sensor-specific geometries.
Recent range- and layout-based methods broaden LiDAR generation beyond unconditional scan synthesis. SPIRAL~\cite{zhu2025spiral} jointly generates range-view depth, reflectance, and semantic maps, while La La LiDAR~\cite{liu2026lidar} introduces scene-graph and layout conditioning for controllable object-level LiDAR scene generation. 
\red{Veila~\cite{liu2026veila} generates panoramic LiDAR from a monocular RGB image, and DynamicCity~\cite{bian2025dynamiccity} models dynamic scenes through 4D occupancy generation.}
However, these methods do not explicitly model synchronized, calibrated multi-LiDAR sensors.
Furthermore, other works using the Waymo Open Dataset~\cite{sun2020scalability}, one of the few available multi-sensor datasets, often restrict themselves to only the front-facing sensor~\cite{wu2024neural,elsayed2022savi++,xu2023discoscene,zhou2024simgen}.
As a result, there is a need for dedicated architectures employing inter-sensor modules, which maintain the unique geometric properties of individual sensors while ensuring global spatial consistency across the entire sensor suite.

\noindent \textbf{\lidar{} Generation Architectures.}
Beyond sensor configuration, the architectural foundations of LiDAR generative models are rapidly evolving. Previous latent diffusion models (LDMs), such as RangeLDM~\cite{hu2024rangeldm}, LiDM~\cite{ran2024towards}, and LiDARCrafter~\cite{liang2026lidarcrafter}, rely on conventional 2D convolutional U-Net backbones.
These architectures are often constrained by local receptive fields, which can struggle to capture the global context and long-range dependencies inherent in sparse 3D data. To address these structural bottlenecks, MM-DiT~\cite{esser2024scaling} architectures offer a more scalable solution. By leveraging self-attention, transformer-based approaches capture spatial and temporal dependencies more effectively than traditional U-Net-based frameworks.

\noindent \textbf{LiDAR Sequence Generation.}
Point cloud prediction~\cite{weng2021inverting, mersch2022self, weng2022s2net, wilson2023argoverse} has emerged as a fundamental self-supervised task in autonomous driving, utilizing historical sequences to forecast future spatial states. Early methods focused on estimating scene flow between consecutive frames to project future point cloud sequences~\cite{liu2019flownet3d, zhai2021optical}. However, these flow-based approaches incur substantial computational costs, limiting their practical applicability in real-time systems. Consequently, recent studies have shifted toward direct prediction from raw point cloud data~\cite{weng2021inverting, weng2022s2net, luo2023pcpnet, pal2024atppnet} to improve efficiency. Further advancements include 4D-Occ~\cite{khurana2023point}, which formulates the task as geometric occupancy prediction, and ViDAR~\cite{yang2024visual}, which leverages historical images to predict future LiDAR scenes.
Despite these developments, existing prediction-based frameworks generally require multiple historical frames as input, still struggle with long-horizon LiDAR forecasting, and provide limited controllability for dynamic scene generation. LiDARCrafter~\cite{liang2026lidarcrafter} addressed long-horizon generation through an autoregressive diffusion framework, but relies on ego-motion-based extrapolation, a strong prior that restricts the diversity of generated sequences. More recently, U4D~\cite{xu2026u4d} introduced uncertainty-aware 4D LiDAR generation by allocating generation capacity to challenging regions and improving spatio-temporal coherence. In contrast, \lname{} focuses on controllable multi-sensor LiDAR generation, combining long-horizon temporal consistency with calibration-aware cross-sensor alignment through a ControlNet-based conditioning branch.

\section{\lname}

\begin{figure*}[htb!]
    \centering
    \includegraphics[width=\linewidth]{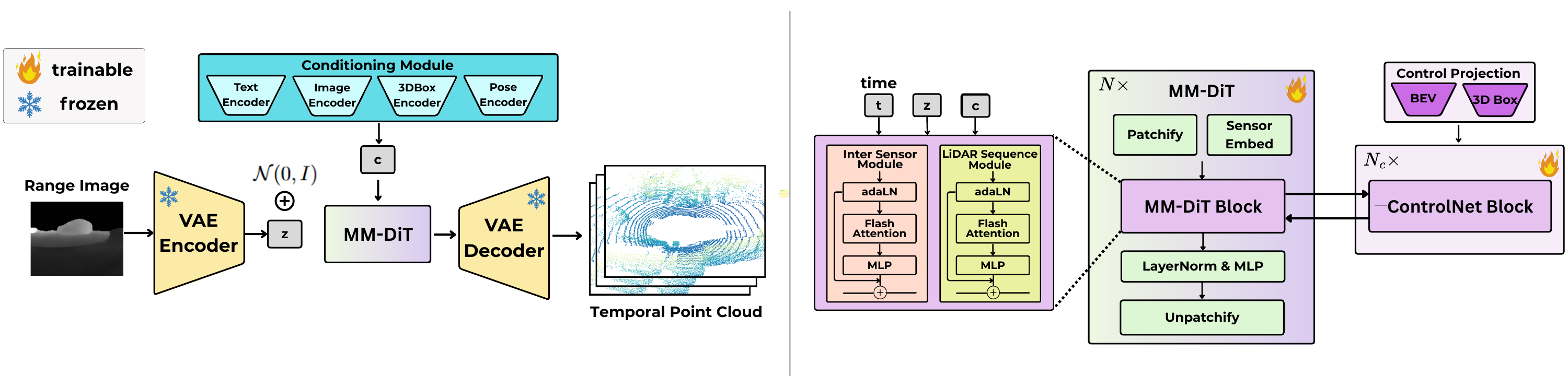}
    \caption{\textbf{Overview of \lname{} Framework}. Our framework utilizes a latent rectified-flow~\cite{liu2022flow} formulation to generate synchronized multi-sensor LiDAR sequences.\ \emph{(Left)} A frozen VAE encoder first maps range images into a compact latent space ${z}_t$.\ \emph{(Right)} These latents are processed by an MM-DiT~\cite{esser2024scaling} backbone, which incorporates two different attention mechanisms: an Inter-sensor Module for cross-sensor alignment and a LiDAR Sequence Module for temporal consistency. The generation is guided by a dedicated ControlNet~\cite{zhang2023adding} branch for BEV and 3D layout control. Finally, a frozen VAE decoder transforms the denoised latents back into high-fidelity temporal point clouds.}%
    \label{fig:architecture}
    \vspace{-.5em}
\end{figure*}

\subsection{Problem Formulation}

We consider $S$ LiDAR sensors mounted at distinct viewpoints around the vehicle. At each time step $k$, each sensor $s \in \{1, \dots, S\}$ generates a point cloud $P_k^{(s)} \subset \mathbb{R}^3$, which is projected into a range image $X_k^{(s)} \in \mathbb{R}^{H \times W \times C}$, where $H$ and $W$ denote the vertical and horizontal resolutions, respectively, and $C$ represents the feature channels (e.g., range, intensity, and elongation). To create a unified input for our model, we stack these images along the sensor dimension to form a multi-sensor tensor $X_k \in \mathbb{R}^{S \times H \times W \times C}$.

Since our objective is to generate sequences of multi-sensor LiDAR data, we denote the ground-truth and generated spatio-temporal tensors as 
$X \in \mathbb{R}^{K \times S \times H \times W \times C}$ and 
$\hat{X} \in \mathbb{R}^{K \times S \times H \times W \times C}$, 
respectively, obtained by stacking $\{X_k\}_{k=1}^K$ (and $\{\hat{X}_k\}_{k=1}^K$) along the temporal dimension, such that the $k$-th slice corresponds to the multi-sensor tensor at time step $k$.
To ensure the generated range images correspond to valid 3D structures, a pixel at $(u, v)$ with range value $r$ (depth) is mapped to Cartesian coordinates $p = [\alpha, \beta, \gamma]^\top$ using the following spherical-to-Cartesian transformation:
$\alpha = r \cdot \cos(\theta) \cdot \cos(\phi)$, $\beta = -r \cdot \cos(\theta) \cdot \sin(\phi)$, {and} $\gamma = r \cdot \sin(\theta)$,
where $\phi$ and $\theta$ denote the horizontal and vertical angles, respectively. These angles are derived from the pixel coordinates $(u, v)$ based on the field-of-view (FOV) and resolution of each sensor.

Our framework consists of two components: a LiDAR compression module and a generative backbone. For compression, we utilize a VAE that operates purely on the spatial dimensions. The encoder $E_\zeta$ processes each sensor frame independently (without temporal or cross-sensor coupling) and maps $X_k^{(s)}$ to a compact latent representation $z_k^{(s)}$. The decoder $D_\eta$ reconstructs the corresponding range image from the latent representation.
The generation process is modeled using an MM-DiT architecture~\cite{esser2024scaling}, which models joint spatio-temporal dependencies across sensors and frames. 
To maintain long-term temporal consistency and mitigate cumulative drift, we employ an autoregressive strategy by adding noise to the first $N$-conditional frame.

\subsection{The Variational Autoencoder}

To mitigate the computational cost of modeling high-dimensional range-image sequences,
we employ a VAE that compresses each input range image into a lower-dimensional latent space~\cite{rombach2022high}.
In our setting, the input tensor is $X\in\mathbb{R}^{K\times S\times H\times W\times C}$. Differently from~\cite{ran2024towards, hu2024rangeldm}, we apply a linear normalization to the raw range values, mapping the input to a symmetric interval $[-1, 1]$.  
We treat each $(k,s)$ pair (frame $k$ and sensor view $s$) as an independent sample by reshaping $(K,S)$ into the batch dimension, i.e., $B = B \cdot K\cdot S$, and encode $x\in\mathbb{R}^{H\times W\times C}$.

The encoder/decoder ($E_\zeta$ and $D_\eta$) follow the design of~\cite{hu2024rangeldm}:
the encoder maps each input range image to a latent feature map $z = E_\zeta(x)\in\mathbb{R}^{c\times h\times w}$, where the spatial dimensions are downsampled by a factor $f$, i.e., $(h,w)=(H/f,\,W/f)$.
The decoder reconstructs $\hat{x}=D_\eta(z)$.
We train our VAE by maximizing the Evidence Lower Bound (ELBO)~\cite{kingma2013auto}:
\begin{equation}\label{eq:elbo}
\begin{aligned}
\mathcal{L}_{\mathrm{ELBO}}(\zeta,\eta)
&=
\mathbb{E}_{q_\zeta(z\mid x)}
\big[\log p_\eta(x\mid z)\big] \\
&\quad
-
\lambda_1
D_{\mathrm{KL}}\!\left(
q_\zeta(z\mid x)\,\|\,p(z)
\right).
\end{aligned}
\end{equation}
where $x$ denotes a single range image drawn from $X$, and $p(z)=\mathcal{N}(0,I)$.  
The first term evaluates the reconstruction likelihood of the decoder $D_\eta$ from the latent $z$. In our implementation, this corresponds to an $L_1$ pixel-wise reconstruction loss~\cite{hu2024rangeldm}. However, as a key departure from~\cite{hu2024rangeldm}, we omit the adversarial discriminator and perceptual loss.

\subsection{Rectified Flow for LiDAR Generation}

After VAE pretraining, we freeze its encoder/decoder, and model the empirical latent distribution induced by the frozen encoder using a flow-based generative model built on a Transformer backbone, MM-DiT~\cite{esser2024scaling}. 
The Transformer processes spatio-temporal latent inputs via specialized temporal-attention and inter-sensor-attention blocks.

\noindent\textbf{Latent Rectified-Flow Formulation.} 
Given the multi-sensor range image sequence $\mathcal{X}$ and $E_\zeta$ as the pretrained VAE encoder, for the data sample we compute the latent
$
    z_1 = E_\zeta(\mathcal{X}) \sim p_1(z),
$
and draw an independent noise latent
$
    z_0 \sim p_0(z) = \mathcal{N}(0, I).
$
For $t \in [0, 1]$, we form the rectified (linear) bridge
$
    z_t = (1 - t)z_0 + t z_1.
$
Following~\cite{esser2024scaling}, the timestep $t$ is sampled according to a Logit-Normal distribution to prioritize the mid-range of the transport path.
The instantaneous ground-truth velocity is constant along the bridge:
\begin{equation*}
    \frac{d}{dt}z_t = z_1 - z_0.
\end{equation*}
\noindent\textbf{Training.} We train a time-dependent vector field $v_\theta(z_t, t, c)$ (parameterized by the Transformer) to regress this displacement by minimizing the rectified flow loss:
\begin{equation*}
    \mathcal{L}_{\mathrm{RF}} = \mathbb{E}_{t, z_0, z_1} \left\| v_\theta(z_t, t, c) - (z_1 - z_0) \right\|_2^2,
\end{equation*}
where $c$ denotes conditioning information (sensor identifiers, textual prompts, bounding-box embeddings, etc.). Under this objective, the optimal vector field is the conditional expectation:
\begin{equation*}
    v^\star(z_t, t) = \mathbb{E}\left[ z_1 - z_0 \mid z_t \right],
\end{equation*}
which provides the probabilistic interpretation of the regression target.

\noindent\textbf{Inference.} 
At inference, we generate latents by integrating the learned ODE\@:
\begin{equation*}
    \frac{dz}{dt} = v_\theta(z, t, c),
\end{equation*}
from $t=0$ to $t=1$. Because the rectified flow objective encourages near-linear transport paths, accurate samples are typically obtainable with a small number of integration steps using a numerical solver (e.g., Euler or mid-point method). Once the latent $z(1)$ is obtained, it is passed through the frozen VAE decoder $D_\eta$ to reconstruct the multi-sensor range image sequence $\hat{\mathcal{X}}$.
While the model operates on range-image latents, the generated range images are back-projected into 3D points using the corresponding LiDAR sensor geometry. The reconstructed points are then transformed into the ego-vehicle coordinate frame using calibrated sensor extrinsics and concatenated to form a unified multi-LiDAR point cloud (Appendix~\ref{sec:scene_metrics}).

\subsection{The Architecture Modules}

\noindent\textbf{MM-DiT Pre-Attention Stage.}
The generative backbone is implemented as a stack of MM-DiT blocks operating on VAE-compressed latents within the rectified-flow framework. Each block processes the latent tokens $z_t$ along with auxiliary concatenated modality tokens $\Phi = \{\phi_1(m_1) \parallel \dots \parallel \phi_D(m_D)\}$. 
Following the canonical MM-DiT design, each block begins with a pre-attention stage utilizing Adaptive Layer Normalization (AdaLN). 
The tokens are then processed through a spatio-temporal attention, where 3D rotary positional embeddings (3D RoPE)~\cite{su2024roformer} are applied to queries and keys. 
A dedicated inter-sensor attention is inserted at this stage to facilitate information exchange across multiple sensors at the same timestep. 

\noindent\textbf{The LiDAR Sequence Module.}
After reshaping the latent tokens as $\tilde{z}_t \in \mathbb{R}^{(B \cdot K)\times N \times c'}$ and the modality tokens $\tilde{\Phi}_t^{seq} \in \mathbb{R}^{(B \cdot K)\times N \times c'}$, with $N \equiv S\cdot h \cdot w$, spatio-temporal dependencies are modeled via a multi-head joint self-attention mechanism.
Let $H$ be the number of heads and $d_h$ the per-head dimension so that $c' = H d_h$. We first compute separate linear projections for $\tilde{z}_t$ and $\tilde{\Phi}_t$, using two distinct sets of learnable weights:
\begin{equation*}
\begin{aligned}
Q_{z} &= \tilde z_t W'_{Q}, \quad K_{z} = \tilde z_t W'_{K}, \quad V_{z} = \tilde z_t W'_{V} \\
Q_{\Phi} &= \tilde{\Phi}_t^{seq} W''_{Q}, \quad K_{\Phi} = \tilde{\Phi}_t^{seq} W''_{K}, \quad V_{\Phi} = \tilde{\Phi}_t^{seq} W''_{V}
\end{aligned}
\end{equation*}
where $W'_{Q,K,V} \in \mathbb{R}^{c'\times c'}$ and $W''_{Q,K,V} \in \mathbb{R}^{c'\times c'}$. 
We then apply 3D RoPE to the query and key of the latent.
Next, the tokens are concatenated to form joint representations $Q$, $K$, $V \in \mathbb{R}^{(BK) \times 2N \times c'}$:
\begin{equation*}
Q = [\tilde{Q}_{z} \parallel \tilde{Q}_{\Phi}], \quad K = [\tilde{K}_{z} \parallel \tilde{K}_{\Phi}], \quad V = [V_{z} \parallel V_{\Phi}].
\end{equation*}
To maintain temporal consistency, we enforce causal masking along the temporal axis. For the joint sequence length $2N$, we define a mapping $\sigma(i)$ that returns the temporal frame $s$ for any index $i \in \{0, \dots, 2N-1\}$ such that $\sigma(i) = s_{i \bmod N}$.
We define the causal mask $\mathcal{M}^{\mathrm{causal}}\in\mathbb{R}^{2N\times 2N}$ by $0$ if $s_j \le s_i$, while $-\infty$ if $s_j > s_i$.
This configuration allows for bidirectional interaction between the LiDAR latent and the modality token at any given time step, while preventing leakage from future frames.
Scaled dot-product attention with the causal mask (per head) $A^{(h)}$ is performed.
The resulting joint output $O = \operatorname{Concat}(A^{(1)}V^{(1)}, \dots, A^{(H)}V^{(H)}) W_O$ is split back into the latent and modality tensors
$[O_z \parallel O_\Phi] = O \in \mathbb{R}^{(BK) \times 2N \times c'}$, where $O_z, O_\Phi \in \mathbb{R}^{(BK) \times N \times c'}$ are passed to their respective residual blocks and point-wise feed-forward networks, obtaining $\tilde z'_t$ and $\tilde \Phi'_t$.

\noindent\textbf{The Inter-sensor Module.}
To resolve geometric disparities between the various LiDAR sensors positioned around the vehicle, the Inter-sensor module treats each sensor as a structured viewpoint and models their spatial interactions \red{through joint attention rather than independent per-sensor generation}. 
For this module, we employ a similar architectural configuration to the LiDAR Sequence Module, 
which consists of projected bounding box embedders to provide explicit spatial grounding for the different views.
To compute the attention between sensors, we reshape the latent output as $\tilde{z}'_t \in \mathbb{R}^{(B \cdot S)\times M \times c'}$ and the modality tokens as $\tilde{\Phi}_t^{inter} \in \mathbb{R}^{(B \cdot S)\times M \times c'}$, where $M = K \cdot h \cdot w$. 
By shifting the sensor count $K$ into the sequence dimension, the joint attention mechanism captures dependencies across different viewpoints within the same temporal frame.

The attention process follows the same mathematical formulation as described in the LiDAR Sequence Module, utilizing separate linear projections $W'_{Q,K,V}$ and $W''_{Q,K,V}$. However, the 3D RoPE is applied across the sensor index $k \in \{0, \dots, K-1\}$ and spatial coordinates $(u, v)$, rather than the temporal axis, while the causal mask is not adopted.

\noindent\textbf{LiDAR ControlNet.}
To enable precise 
\red{BEV-based structural}
control over the synthesis process, we incorporate a ControlNet-style branch~\cite{zhang2023adding}
\red{conditioned on BEV maps containing projected 3D bounding boxes (Appendix~\ref{app:implementation})}.
Following~\cite{zhang2023adding}, we create a trainable copy of the MM-DiT backbone blocks while keeping the original generative weights frozen.

The ControlNet branch adopts an internal structure identical to the primary modules, including the factorized attention schemes of the LiDAR Sequence and Inter-sensor modules. For each block in this branch, we first fuse the structural control signals (3D bounding boxes and BEV maps), with the current noisy latent $z_t$. Specifically, the control signals are processed by a lightweight encoder to produce the embedding $\Phi^{ctrl}$, which is summed with $z_t$ to form a unified input for the ControlNet blocks. 
Therefore, the control signals are directly grounded in the current denoising state, allowing for more accurate structural guidance.
The output of the $n$-th ControlNet block, $\mathrm{Block}_{\mathrm{ctrl}}^{(n)}$ integrates into the corresponding backbone, $\mathrm{Block}_{\mathrm{backbone}}^{(l)}$, 
via a zero-initialized residual connection.

\subsection{Autoregressive Generation}
Na\"{\i}ve autoregressive LiDAR generation iteratively conditions on a model's own predictions, leading to exposure bias and compounding errors over long horizons~\cite{Venkatraman_Hebert_Bagnell_2015, chen2025diffusion, huang2025selfforcing}. Small inaccuracies in early frames are amplified during rollout, causing drift and degraded generation quality. To ameliorate this, we add a small amount of noise to the ground-truth conditioning frame during training. More precisely, rather than using clean conditioning frames during training we fix the noise level of initial frames to $t \in[0.7,1]$, encouraging robustness to exposure bias during autoregression.

\section{Experimental Results}

\subsection{Experimental Settings}

\noindent\textbf{Datasets.}
The nuScenes dataset~\cite{caesar2020nuscenes} is designed for multi-modal perception in complex urban environments. We utilize 32-beam LiDAR sweeps. Each LiDAR sweep (sample) is initially represented as a range image with a resolution of $32 \times 1024$ pixels (totaling 32,768 pixels), a vertical Field of View (FOV) spanning $(+10^\circ, -30^\circ)$, and a depth range between 1 and 45 meters.
The Waymo Open Dataset~\cite{sun2020scalability} utilizes a sensor suite consisting of five LiDAR sensors. This configuration includes a primary high-performance mid-range LiDAR (Top) for long-range $360^{\circ}$ coverage, supplemented by four short-range side LiDARs (Front, Left, Right, and Rear) positioned around the vehicle's perimeter. These side sensors are specifically designed to capture the immediate surroundings with high precision, providing a unique perspective that is less susceptible to the top-down occlusions common in roof-only configurations.
Further details on both datasets are provided in Appendix~\ref{app:datasets}.

\noindent\textbf{Baselines.} 
We compare \lname{} against a comprehensive set of baselines categorized by their underlying representation. Range-based methods include LiDARGen~\cite{zyrianov2022learning}, R2DM~\cite{nakashima2024lidar}, RangeLDM~\cite{hu2024rangeldm}, LiDM~\cite{ran2024towards}, and LiDARCrafter~\cite{liang2026lidarcrafter}. Voxel-based methods include the OpenDWM models~\cite{opendwm,ni2025maskgwm}, and we additionally compare against the occupancy-centric UniScene~\cite{li2025uniscene}.
Details of these methods under our setting are discussed in Appendix~\ref{app:baselines}.

\begin{table}[htb!]
\centering
\caption{%
    \textbf{Evaluations of Scene-level Fidelity on Waymo dataset} for multi-sensor LiDAR generation.}%
\label{tab:video_multi_sensor_waymo_image_based}
\scriptsize
\setlength{\tabcolsep}{4pt}
\resizebox{\columnwidth}{!}{
\begin{tabular}{lcrrrrrr}
\toprule
\multirow{2}{*}{\textbf{Method}} & \multirow{2}{*}{\textbf{Venue}} &
\multicolumn{2}{c}{\textbf{Range}} &
\multicolumn{2}{c}{\textbf{Points}} &
\multicolumn{2}{c}{\textbf{BEV}} \\
\cmidrule(lr){3-4}\cmidrule(lr){5-6}\cmidrule(lr){7-8}
& & \textbf{FRD}$\downarrow$ & \textbf{MMD}$\downarrow$ & \textbf{FPD}$\downarrow$ & \textbf{MMD}$\downarrow$ & \textbf{JSD}$\downarrow$ & \textbf{MMD}$\downarrow$ \\
\midrule
\multicolumn{8}{l}{\textbf{Voxel}}\\
OpenDWM\cite{ni2025maskgwm} & CVPR'25 & --- & --- & 7.55 & 52.23 & 16.25 &  2.29 \\
OpenDWM-DiT\cite{opendwm} & CVPR'25 & --- & --- & 8.41 & 105.40 & 10.85 & 1.80 \\
\midrule
\multicolumn{8}{l}{\textbf{Range}}\\
LiDM\cite{ran2024towards}  & CVPR'24 & \underline{2.29} & \underline{148.83} & \underline{4.67} & \underline{65.76} & \underline{19.34} & \textbf{1.00} \\
RangeLDM\cite{hu2024rangeldm} & ECCV'24 & 10.97 & 979.33 & 122.38 & 1236.59 & 22.33 & 1.63 \\
R2DM\cite{nakashima2024lidar}             & ICRA'24 & 21.63 & 2283.90 & 40.56 & 228.04 & 27.54 & 1.97 \\
\midrule
\rowcolor[HTML]{F5D0CB}
  \textbf{\lname{}}           & --- & \textbf{0.16} & \textbf{11.89} & \textbf{1.32} & \textbf{6.90} & \textbf{8.09} & \underline{1.05} \\
\bottomrule
\end{tabular}
}
\end{table}

\begin{table}[htb!]
\caption{%
    \textbf{Evaluations of Scene-level Fidelity on nuScenes dataset} for LiDAR generation.
}%
\label{tab:video_nuscenes_image_based}
\centering
\setlength{\tabcolsep}{4pt}
\resizebox{\columnwidth}{!}{
    \begin{tabular}{lcrrrrrr}
    \toprule
    \multirow{2}{*}{\textbf{Method}} & \multirow{2}{*}{\textbf{Venue}} &
    \multicolumn{2}{c}{\textbf{Range}} &
    \multicolumn{2}{c}{\textbf{Points}} &
    \multicolumn{2}{c}{\textbf{BEV}} \\
    \cmidrule(lr){3-4}\cmidrule(lr){5-6}\cmidrule(lr){7-8}
    & & \textbf{FRD}$\downarrow$ & \textbf{MMD}$\downarrow$ & \textbf{FPD}$\downarrow$ & \textbf{MMD}$\downarrow$ & \textbf{JSD}$\downarrow$ & \textbf{MMD}$\downarrow$ \\
    \midrule
    \multicolumn{8}{l}{\textbf{Voxel}}\\
    UniScene\cite{li2025uniscene}      & CVPR'25 & ---     & ---    & --- & --- & 31.55 & 13.61 \\
    OpenDWM\cite{ni2025maskgwm}        & CVPR'25 & ---     & ---    & 0.95 & 5.43 & 20.17 & 5.61 \\
    OpenDWM-DiT\cite{opendwm}          & CVPR'25 & ---     & ---    & \underline{0.49} & \textbf{0.50} & 19.90 & 5.73 \\
    \midrule
    \multicolumn{8}{l}{\textbf{Range}}\\
    LiDM\cite{ran2024towards}              & CVPR'24 & 495.54 & 0.18  & 2.06 & 1.83 & 5.86 & 0.73 \\
    RangeLDM\cite{hu2024rangeldm}          & ECCV'24 & ---     & ---    & 14.22 & 184.74 & 5.47 & 1.92 \\
    R2DM\cite{nakashima2024lidar}          & ICRA'24 & 243.35 & 1.40  & --- & --- & 3.51 & 0.71 \\
    LiDARCrafter\cite{liang2026lidarcrafter}  & AAAI'26 & \underline{194.30} & \textbf{0.08}  & --- & --- & \underline{3.11} & \underline{0.42} \\
    \midrule
    \rowcolor[HTML]{F5D0CB}
      \lname{}           & ---     & \textbf{181.41}  &  \underline{0.09}  & \textbf{0.21} & \underline{0.93} & \textbf{3.08} & \textbf{0.27} \\ 
    \bottomrule
    \end{tabular}
}
\end{table}

\noindent\textbf{Evaluation Metrics.}
Following~\cite{liang2026lidarcrafter}, we evaluate the quality of the generated LiDAR data across three complementary representations:
\textit{(1) Range-image fidelity} using Fréchet Range Distance (FRD) and Maximum Mean Discrepancy (MMD);
\textit{(2) Point-cloud geometry} using Fréchet Point Distance (FPD) and MMD\@; and
\textit{(3) BEV distribution} using Jensen--Shannon Divergence (JSD) and MMD\@.
To evaluate temporal dynamics, we follow~\cite{liang2026lidarcrafter} by measuring frame-to-frame transformation accuracy and sequence smoothness using Temporal Transform Consistency Error (TTCE) and Chamfer Temporal Consistency (CTC). We additionally adapt three metrics from the video synthesis literature to the LiDAR domain to provide a more comprehensive assessment of \textit{(4) Temporal coherence}: Blind Video Temporal Consistency (BVC)~\cite{10.1145/2816795.2818107}, Fréchet Video Distance (FVD)~\cite{unterthiner2018towards}, and Fréchet Video Motion Distance (FVMD)~\cite{liu2024fr}. Finally, we evaluate rollout quality by analyzing the accumulated point-cloud sequences.
MMD values are reported in $10^{-4}$ and JSD values in $10^{-2}$. In all tables, the best result is highlighted in \textbf{bold} and the second-best in \underline{underline}.
\red{We evaluate a fixed, calibrated four-LiDAR configuration, using sensor-specific range-image geometry and extrinsic calibration for ego-frame alignment.} 
More details are provided in Appendix~\ref{app:metrics}. 


\begin{figure*}[htb!]
    \centering

    \textbf{nuScenes}\par\vspace{0.15cm}

    \noindent
    \begin{subfigure}[t]{0.21\textwidth}
        \centering
        \includegraphics[width=\linewidth,keepaspectratio]{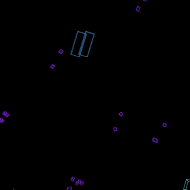}
        \caption*{\textbf{BEV/3D Box}}
    \end{subfigure}\hspace{0.05\textwidth}%
    \begin{subfigure}[t]{0.21\textwidth}
        \centering
        \includegraphics[width=\linewidth,keepaspectratio]{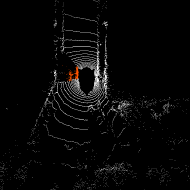}
        \caption*{$t=0$}
    \end{subfigure}\hspace{0.015\textwidth}%
    \begin{subfigure}[t]{0.21\textwidth}
        \centering
        \includegraphics[width=\linewidth,keepaspectratio]{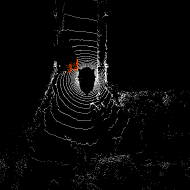}
        \caption*{$t=1$}
    \end{subfigure}\hspace{0.015\textwidth}%
    \begin{subfigure}[t]{0.21\textwidth}
        \centering
        \includegraphics[width=\linewidth,keepaspectratio]{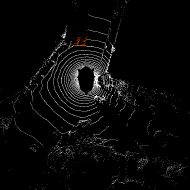}
        \caption*{$t=2$}
    \end{subfigure}

    \vspace{0.15cm}
    \rule{\linewidth}{0.5pt}
    \vspace{-0.5cm}

    \textbf{Waymo}\par\vspace{-0.3cm}

    \includegraphics[width=0.935\textwidth,keepaspectratio]{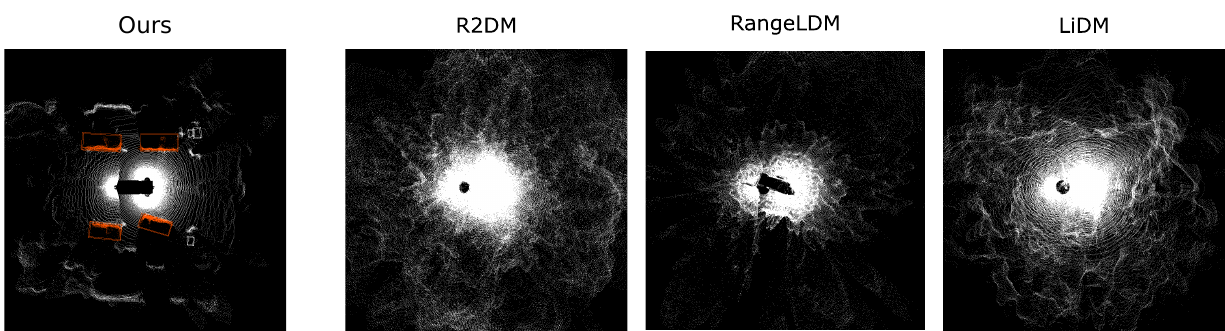}

    \caption{
    \textbf{Qualitative Results}.
    The \emph{top} row illustrates temporal sequence generation on nuScenes, where \lname{} maintains strict temporal consistency and 3D bounding box alignment.
    The \emph{bottom} row demonstrates multi-sensor Waymo generation, where a single unified scene is reconstructed by synchronizing outputs from four distinct generated sensors.
    Compared to baselines, \lname{} produces scan patterns with improved scene-level geometric fidelity, effectively handling the complexities of multi-sensor generation.
    }%
    \label{fig:qualitative_results}
    \vspace{-.5em}
\end{figure*}

\subsection{Image Generation} 
We evaluate \lname{} on single-frame LiDAR generation across multi- and single-sensor benchmarks.
Additional details and ablation studies are in Appendix~\ref{app:implementation} and~\ref{app:ablation}.

\begin{table}[htb!]
\centering
\caption{\textbf{4D LiDAR generation on Waymo} using video metrics including: \emph{Temporal Consistency} and 
\emph{Rollout-level} metrics. 
}%
\label{tab:video_multi_sensor_waymo_video_based}
\resizebox{\columnwidth}{!}{
\begin{tabular}{lccccccccccc}
\toprule
\multirow{2}{*}{\textbf{Method}} & \multirow{2}{*}{\textbf{Venue}} & \multicolumn{2}{c}{\textbf{TTCE} $\downarrow$} & \multicolumn{4}{c}{\textbf{CTC} $\downarrow$} & \multicolumn{3}{c}{\textbf{Rollout} $\downarrow$}\\
\cmidrule(lr){3-4} \cmidrule(l){5-8} \cmidrule(l){9-11}
 &  & \textbf{1} & \textbf{2} & \textbf{1} & \textbf{2} & \textbf{3} & \textbf{4} & \textbf{CD} & \textbf{JSD} & \textbf{MMD} \\
 \midrule
\multicolumn{8}{l}{\textbf{Voxel}}\\
OpenDWM\cite{ni2025maskgwm} (t=5) & CVPR’25 & 990.17 & 990.74 & 0.37 & 0.52 & 0.65 & 0.68 & 2.81 & 0.53 & 0.57 \\
OpenDWM-DiT\cite{opendwm} (t=5) & CVPR’25 & 1025.78 & 1026.43 & 0.39 & 0.59 & 0.76 & 0.81 & 0.78 & 0.13 & 0.34 \\
\midrule
\multicolumn{11}{l}{\textbf{Range-Temporal}}\\
LiDM\cite{ran2024towards}(t=3) & CVPR’24 & 1444.74 & 1444.48 & 0.20 & 0.24 & --- & --- & 0.53 & 0.52 & 1.81 \\
RangeLDM\cite{hu2024rangeldm}(t=3) & ECCV’24 & 1686.14 & 1686.89 & 0.24 & 0.27 & --- & --- & 0.84 & 0.64 & 8.75 \\
\midrule
\rowcolor[HTML]{F5D0CB}
\textbf{\lname} (t=5) & --- & 1016.74 & 1017.38 & \textbf{0.06} & \textbf{0.10} & \textbf{0.12} & \textbf{0.14} & \textbf{0.07} & \textbf{0.10} & \textbf{0.03} \\
\bottomrule
\end{tabular}
}
\end{table}

\noindent\textbf{Waymo.}
Table~\ref{tab:video_multi_sensor_waymo_image_based} reports scene-level fidelity on Waymo, our primary benchmark for the multi-sensor design, where \lname{} outperforms all baselines.
In the range-image domain, it improves FRD by 93.0\% over the second-best method.
Point-cloud fidelity follows the same trend, with a 71.7\% FPD reduction relative to LiDM\@.
On layout, \lname{} reduces JSD by 25.4\% over the next-best baseline.
These results indicate better scene-geometry preservation,
further supported by our Ablation Study (Appendix~\ref{sec:ablate_diffusion}) showing that the combined LiDAR Sequence and Inter-sensor Modules reduce FRD by 88.8\% over naive stacking. 

\begin{table}[htb!]
\centering
\caption{\textbf{4D LiDAR generation on nuScenes} using video metrics including: \emph{Temporal Consistency} and \emph{Rollout-level} metrics. The temporal horizon is fixed to $t=5$ for all models.
}%
\label{tab:video_nuscenes_video_based}
\resizebox{\columnwidth}{!}{
\begin{tabular}{lccccccccccc}
\toprule
\multirow{2}{*}{\textbf{Method}} & \multirow{2}{*}{\textbf{Venue}} & \multicolumn{2}{c}{\textbf{TTCE} $\downarrow$} & \multicolumn{4}{c}{\textbf{CTC} $\downarrow$} & \multicolumn{3}{c}{\textbf{Rollout} $\downarrow$}\\
\cmidrule(lr){3-4} \cmidrule(l){5-8} \cmidrule(l){9-11}
 &  & \textbf{3} & \textbf{4} & \textbf{1} & \textbf{2} & \textbf{3} & \textbf{4} & \textbf{CD} & \textbf{JSD} & \textbf{MMD} \\
 \midrule
\multicolumn{11}{l}{\textbf{Voxel}}\\
UniScene\cite{li2025uniscene} & CVPR’25 & 2.74 & 3.69 & \underline{0.90} & 1.84 & 3.64 & \underline{3.90} & --- & --- & --- \\
OpenDWM\cite{ni2025maskgwm} & CVPR’25 & 2.68 & 3.65 & 1.02 & 2.02 & 3.37 & 5.05 & 1.20 & \underline{0.17} & 9.99\\
OpenDWM-DiT\cite{opendwm} & CVPR’25 & 2.71 & 3.66 & \textbf{0.89}& \underline{1.79} & 3.06 & 4.64 & \underline{0.68} & \underline{0.17} & \underline{8.80} \\
\midrule
\multicolumn{11}{l}{\textbf{Range-Temporal}}\\
LiDARCrafter\cite{liang2026lidarcrafter} & AAAI'26 & \underline{2.65} & \underline{3.56} & 1.12 & 2.38 & \underline{3.02} & 4.81 & --- & --- & --- \\ 
\midrule
\rowcolor[HTML]{F5D0CB}

\textbf{\lname} & --- & \textbf{2.40} & \textbf{3.01} & 1.42 & \textbf{1.71} & \textbf{2.10} & \textbf{2.94} & \textbf{0.41} & \textbf{0.14} & \textbf{0.03}\\
\bottomrule
\end{tabular}
\vspace{-.5em}
}
\end{table}

\begin{table*}[htb!]
\centering
\caption{\textbf{Auto-regressive generation on Waymo}, including \emph{Temporal Consistency}, 
\emph{Rollout-level} 
and \emph{Motion-level} metrics. 
}%
\label{tab:video_waymo_ar}
\resizebox{0.95\textwidth}{!}{
    \begin{tabular}{@{}lcccccccccccc@{}}
    \toprule
    \multirow{2}{*}{\textbf{Method}} & \multirow{2}{*}{Venue} & \multicolumn{2}{c}{\textbf{TTCE} $\downarrow$} & \multicolumn{4}{c}{\textbf{CTC} $\downarrow$} & \multicolumn{3}{c}{\textbf{Rollout} $\downarrow$} & \multicolumn{2}{c}{\textbf{Motion}}\\
    \cmidrule(lr){3-4} \cmidrule(l){5-8} \cmidrule(l){9-11} \cmidrule(l){12-13} 
     &  & \textbf{3} & \textbf{4} & \textbf{1} & \textbf{2} & \textbf{3} & \textbf{4} & \textbf{CD} & \textbf{JSD} & \textbf{MMD} & \textbf{BVC}$\uparrow$ & \textbf{FVD}$\downarrow$\\
     \midrule
OpenDWM~\cite{ni2025maskgwm} (t=12) & CVPR'25 & 471.34 & 471.40 & 0.42 & 0.55 & 0.66 & 0.73 & 0.11 & 0.38 & 0.17 & --- & --- \\
OpenDWM-DiT~\cite{opendwm} (t=12) & CVPR'25 & 424.65 & 424.86 & 2.07 & 2.63 & 2.81 & 3.17 & \textbf{0.07} & 0.37 & 0.14 & --- & --- \\
\midrule
\rowcolor[HTML]{F5D0CB}
\textbf{\lname} & --- & \textbf{369.63} & \textbf{369.74} & \textbf{0.05} & \textbf{0.07} & \textbf{0.08} & \textbf{0.09} & \textbf{0.07} & \textbf{0.12} & \textbf{0.02} & 0.79 & 57.42 \\
\bottomrule
\end{tabular}%
}
\end{table*}

\begin{table*}[htb!]
\centering
\caption{\textbf{Autoregressive generation on nuScenes}, including \emph{Temporal Consistency},
\emph{Rollout-level} and
\emph{Motion-level} metrics.
}%
\label{tab:video_nuscenes_ar}
\resizebox{0.95\textwidth}{!}{
    \begin{tabular}{@{}lccccccccccccc@{}}
    \toprule
    \multirow{2}{*}{\textbf{Method}} & \multirow{2}{*}{Venue} & \multicolumn{2}{c}{\textbf{TTCE} $\downarrow$} & \multicolumn{4}{c}{\textbf{CTC} $\downarrow$} & \multicolumn{3}{c}{\textbf{Rollout} $\downarrow$} & \multicolumn{3}{c}{\textbf{Motion}}\\
    \cmidrule(lr){3-4} \cmidrule(l){5-8} \cmidrule(l){9-11} \cmidrule(l){12-14} 
     &  & \textbf{3} & \textbf{4} & \textbf{1} & \textbf{2} & \textbf{3} & \textbf{4} & \textbf{CD} & \textbf{JSD} & \textbf{MMD} & \textbf{BVC}$\uparrow$ & \textbf{FVD}$\downarrow$ & \textbf{FVMD}$\downarrow$\\
     \midrule
    UniScene~\cite{li2025uniscene} (t=5) & CVPR'25 & 2.74 & 3.69 & 0.90 & 1.84 & 3.64 & 3.90 & --- & --- & --- & --- & --- & --- \\
    OpenDWM~\cite{ni2025maskgwm} (t=16) & CVPR'25 & 2.64 & 3.16 & \underline{0.82} & 1.62 & 2.83 & 4.67 & 0.97 & 0.46 & 0.27 & 0.22 & 3923.11 & 4147.61 \\
    OpenDWM-DiT~\cite{opendwm} (t=16) & CVPR'25 & \underline{1.23} & \underline{1.39} & \textbf{0.68} & \underline{1.42} & \underline{2.42} & \underline{3.83} & \underline{0.83} & \underline{0.24} & \textbf{0.13} & \underline{0.43} & \underline{2835.68} & \underline{2261.98} \\
    LiDARCrafter~\cite{liang2026lidarcrafter} (t=5) & AAAI'26 & 2.65 & 3.56 & 1.12 & 2.38 & 3.02 & 4.81 & --- & --- & --- & --- & --- & --- \\ 
    \midrule
    \rowcolor[HTML]{F5D0CB}
    \textbf{\lname} (t=16) & --- & \textbf{0.41} & \textbf{0.50} & 1.09 & \textbf{1.22} & \textbf{1.36} & \textbf{1.47} & \textbf{0.47} & \textbf{0.22} & \underline{0.26} & \textbf{0.88} & \textbf{307.07} & \textbf{1672.90} \\
    \bottomrule
    \end{tabular}
    \vspace{-1em}
}
\end{table*}

\noindent\textbf{nuScenes.}
Table~\ref{tab:video_nuscenes_image_based} reports results on the single-sensor nuScenes benchmark, where \lname{} achieves SOTA across all three representation domains.
The largest improvement is in the point-cloud domain, where our model achieves a $57.1\%$ reduction in FPD compared with OpenDWM-DiT.
BEV alignment is also improved, with \lname{} BEV MMD representing a $35.7\%$ reduction over LiDARCrafter.
Our method also improves range-image fidelity, reducing FRD by $6.6\%$ compared with LiDARCrafter. These results show the proposed framework remains effective even without the Inter-sensor Module.

\subsection{4D LiDAR Generation}

We evaluate the temporal coherence of \lname{} on multi- and single-sensor LiDAR sequence generation.

\noindent\textbf{Waymo.}
Table~\ref{tab:video_multi_sensor_waymo_video_based} reports 4D generation metrics on Waymo.
Since no prior method supports multi-sensor 4D LiDAR generation, we evaluate \lname{}
on Waymo's fixed four-SIDE-LiDAR configuration
at a rollout horizon of $t{=}5$ to assess temporal coherence and cross-sensor consistency. The model remains highly stable across all four CTC steps and achieves the best performance across all rollout-level metrics. Qualitative results in Figure~\ref{fig:qualitative_results} further show that \lname{} better preserves foreground structure, temporal alignment, and cross-sensor geometric consistency than range-based baselines.
Additional multi-sensor rollouts, temporal consistency, and controllability examples are provided in Appendix Figures~\ref{fig:waymo_cross_sensor_alignment} and~\ref{fig:waymo_temporal_consistency}.

\noindent\textbf{nuScenes.}
Table~\ref{tab:video_nuscenes_video_based} reports temporal generation results on the single-sensor nuScenes benchmark.\
\lname{} achieves the best TTCE scores and outperforms prior methods on longer-horizon CTC metrics, with the performance gap increasing as sequence length increases.
\lname{} achieves a $15.4\%$ reduction in TTCE at step 4 compared with LiDARCrafter, indicating better adherence to physically plausible motion trajectories. Although OpenDWM-DiT achieves a lower CTC-1 error, its error accumulates rapidly over longer horizons. In contrast, \lname{} reduces CTC-4 by $36.6\%$ relative to OpenDWM-DiT.
Rollout quality also improves, with a $39.7\%$ and $65.8\%$ reduction in CD compared with OpenDWM-DiT and OpenDWM, respectively. These results demonstrate our proposed framework effectively mitigates the geometric drift commonly associated with long-horizon 4D LiDAR generation.

\subsection{Autoregressive 4D LiDAR Generation}
We assess long-horizon stability by extending \lname{} autoregressively beyond its training window, in both multi- and single-sensor settings.

\noindent\textbf{Waymo.} Table~\ref{tab:video_waymo_ar} reports autoregressive generation on Waymo, where four side LiDARs make cross-sensor consistency especially hard to maintain as the rollout horizon grows. Extending the generated window to $t{=}12$, \lname{} achieves the best temporal consistency, reducing TTCE by $13.0\%$ over the strongest baseline and reducing CTC by an average of $87.8\%$ across all four evaluation steps. At the rollout level, \lname{} further reduces JSD by $67.6\%$ and MMD by $85.7\%$ compared with the strongest baseline. These results show the proposed framework effectively preserves cross-sensor geometric consistency and temporal coherence during long autoregressive rollouts.

\noindent\textbf{nuScenes.} Table~\ref{tab:video_nuscenes_ar} evaluates 16-frame rollouts on nuScenes, where \lname{} shows substantially improved long-horizon stability over previous methods. While LiDARCrafter's CTC increases by $329.5\%$ across the evaluated steps, \lname{} grows by only $34.9\%$, consistent with our noise-augmented training strategy for imperfect autoregressive conditioning.\
\lname{} further reduces TTCE by $66.7\%$ and $64.0\%$ at steps 3 and 4, respectively, over the second-best method, suggesting the proposed flow-matching transformer better preserves physically plausible ego-motion.
The gap between TTCE and CTC suggests dynamic objects remain harder to model than ego-motion.
At the distributional level, \lname{} reduces FVD by $89.2\%$ over OpenDWM-DiT, bringing the generated sequences substantially closer to the real LiDAR distribution in appearance and dynamics.\
\lname{} reduces FVMD by $26.0\%$ while improving BVC by $104.7\%$, demonstrating more realistic motion and stronger temporal stability.
These results show \lname{} maintains high quality over long horizons, a key capability for scalable autonomy simulation.

\section{Conclusions and Limitations}
We present \lname{}, a unified rectified-flow transformer for controllable multi-sensor LiDAR generation in a compact range-image latent space.
It comprises three key components: an Inter-sensor Module that models geometric dependencies across heterogeneous LiDAR suites, a ControlNet-style conditioning branch for fine-grained structural control via BEV priors and 3D bounding boxes, and a noise-augmented autoregressive strategy that mitigates cumulative drift over long horizons.
On Waymo and nuScenes, \lname{} achieves strong performance in single-frame fidelity, temporal consistency, and long-horizon rollout quality, with particularly strong gains in the multi-sensor setting unsupported by previous methods.

Current limitations include the lack of quantitative evaluation of foreground fidelity and bounding-box adherence. 
Our evaluation focuses on Waymo's fixed, calibrated four-LiDAR configuration, motivating further study of sensor conditioning, cross-sensor overlap, and adaptability to different sensor counts, types, layouts, and calibrations.
The LiDAR Sequence and Inter-sensor Modules are evaluated jointly, leaving their individual contributions unquantified. 
Evaluation against additional recent 4D LiDAR generation methods remains limited. 
Downstream benefits for 3D detection, semantic segmentation, occupancy prediction, and planning remain unevaluated.\ 
Future work will address richer dynamic object modeling and integration with downstream planning and validation pipelines, while further reducing the sim-to-real gap.




\newpage

 {
     \small
     \bibliographystyle{ieeenat_fullname}
     \bibliography{main}
 }

\clearpage

\section*{Appendix}
\section{Datasets}%
\label{app:datasets}
We evaluate our approach using two large-scale autonomous driving datasets: nuScenes~\cite{caesar2020nuscenes} and the Waymo Open Dataset~\cite{sun2020scalability}. Each provides multi-modal sensor data together with annotations. In this section, we summarize their overall statistics, sensor configurations, image resolutions, and available labels.

\subsection{nuScenes}%
\label{sec:dataset_nuscenes}
The nuScenes dataset~\cite{caesar2020nuscenes} provides diverse multimodal sensor data collected using two identically instrumented vehicles for autonomous driving research.
The data were recorded with a surround-view sensor suite comprising one spinning 32-beam \lidar{} sensor, six RGB cameras, and five automotive radar sensors.
Together, these sensors provide complementary observations with full $360^\circ$ coverage around the ego vehicle (specifications in Table~\ref{tab:nuscenes_lidar_specs}).
The \lidar{} sensor captures point clouds at $20$ Hz, with a $360^\circ$ horizontal field of view and a vertical field of view from $-30^\circ$ to $10^\circ$, producing up to $1.4$ million points per second with an effective range of up to $70$ m.
In addition, the cameras operate at $12$ Hz and the radar sensors at $13$ Hz.
\begin{table}[htb!]
\begin{center}
\caption{\textbf{LiDAR Sensor Specifications for the nuScenes Dataset}. We report the vertical field of view (\emph{VFOV}), horizontal coverage (\emph{HFOV}), range-image resolution
(rows$\times$columns), valid range (m), and intensity range.}%
\label{tab:nuscenes_lidar_specs}
\begin{small}
\resizebox{\columnwidth}{!}{
\begin{tabular}{lccccc}
\toprule
Sensor & VFOV & HFOV & Resolution & Range (m) & Intensity\\
\midrule
Velodyne HDL-32E & $[+10^{\circ};-30^{\circ}]$ & $360^{\circ}$ & (32, 1080) & $[2.0; 100.0]$ & $[0; 225]$ \\
\bottomrule
\end{tabular}
}
\end{small}
\end{center}
\end{table}

nuScenes provides a large corpus of annotated 3D bounding boxes for urban driving scenes captured under diverse traffic, lighting, and weather conditions.
The full dataset contains 1,000 scenes, each 20s long, selected from 84 driving logs collected in Boston and Singapore.
Although the sensor streams are temporally dense, 3D box annotations are provided only for keyframes sampled at 2 Hz.
In total, the dataset includes approximately 40,000 annotated keyframes, about 400,000 LiDAR sweeps, 1.4M camera images, 1.4M radar sweeps, and 1.4M annotated 3D bounding boxes across 23 classes.
Annotations are defined in the global coordinate frame and are distributed with calibrated sensor poses, enabling projection into the ego or sensor coordinate frames when needed for model training and evaluation.
Beyond object boxes, nuScenes also provides rich scene-level metadata, including location, weather, time of day, and object attributes.

\subsection{Waymo}%
\label{sec:dataset_waymo}
The Waymo Open Dataset~\cite{sun2020scalability} provides diverse multi-modal sensor data collected from a fleet of self-driving vehicles. Each vehicle is equipped with five high-resolution \lidar{} sensors: one mounted on the roof (TOP) and four located on the SIDEs: front (F), rear (R), side-left (SL), and side-right (SR). All \lidar{} data is provided as range images; detailed specifications\footnote{\url{https://waymo.com/open/data/perception/\#lidar-data}} are listed in Table~\ref{tab:waymo_stats_sensor_configs}.

\begin{table}[H]
\centering
\caption{\textbf{LiDAR Sensor Specifications on Waymo Open Dataset}. 
We report vertical field of view (VFOV), horizontal coverage (HFOV), range-image resolution, valid range (m), and intensity range for TOP and SIDE sensors.}%
\label{tab:waymo_stats_sensor_configs}
\setlength{\tabcolsep}{3pt}      
\renewcommand{\arraystretch}{0.95}
\resizebox{\columnwidth}{!}{%
\begin{tabular}{lccccc}
\toprule
Sensor & VFOV & HFOV & Resolution & Range (m) & Intensity \\
\midrule
TOP & $[-17.6^\circ, +2.4^\circ]$ & $360^\circ$ & $64\times2650$ & $[1.2, 75.0]$ & $[7\times10^{-6}, 4\times10^{6}]$ \\
F/SL/SR/R & $[-90^\circ, +30^\circ]$ & --- & $200\times600$ & $[5\times10^{-3}, 20.0]$ & $[5\times10^{-6}, 40.5]$ \\
\bottomrule
\end{tabular}%
}
\vspace{-2mm}
\end{table}

\begin{figure*}[t]
    \centering
    \begin{subfigure}[b]{0.995\textwidth}
        \centering
        \includegraphics[width=\textwidth,height=2.8cm,keepaspectratio]{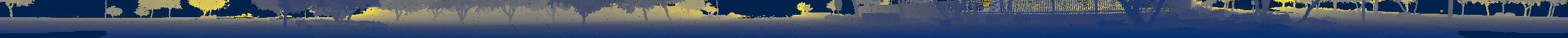}
        \caption*{Top}
    \end{subfigure}

    \vspace{0.4em}

    \begin{subfigure}[b]{0.245\textwidth}
        \centering
        \includegraphics[width=\textwidth,height=2.8cm,keepaspectratio]{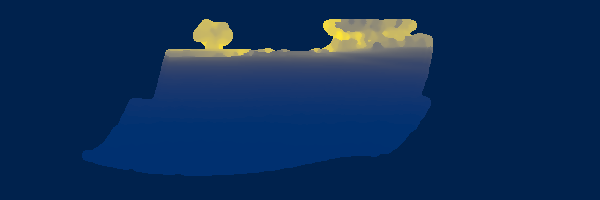}
        \caption*{Front}
    \end{subfigure}\hfill
    \begin{subfigure}[b]{0.245\textwidth}
        \centering
        \includegraphics[width=\textwidth,height=2.8cm,keepaspectratio]{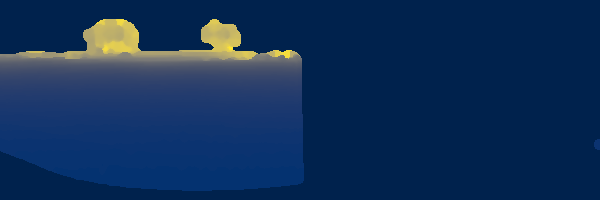}
        \caption*{Side-Left}
    \end{subfigure}\hfill
    \begin{subfigure}[b]{0.245\textwidth}
        \centering
        \includegraphics[width=\textwidth,height=2.8cm,keepaspectratio]{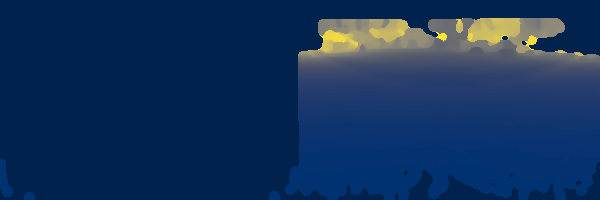}
        \caption*{Side-Right}
    \end{subfigure}\hfill
    \begin{subfigure}[b]{0.245\textwidth}
        \centering
        \includegraphics[width=\textwidth,height=2.8cm,keepaspectratio]{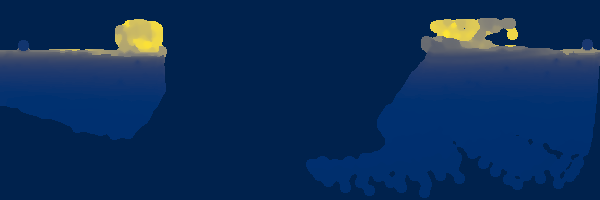}
        \caption*{Rear}
    \end{subfigure}

    \caption{\textbf{Waymo range-image examples.} \textit{Top:} TOP \lidar{} range image ($64\times2650$), with $360^\circ$ horizontal coverage. \textit{Bottom:} four SIDE \lidar{} range images ($200\times600$ each), with higher vertical resolution and partial-view coverage.}%
    \label{fig:waymo_example_range}
\end{figure*}

\begin{figure*}[htbp]
    \centering
    \begin{subfigure}[t]{0.46\textwidth}
        \centering
        \includegraphics[width=\textwidth]{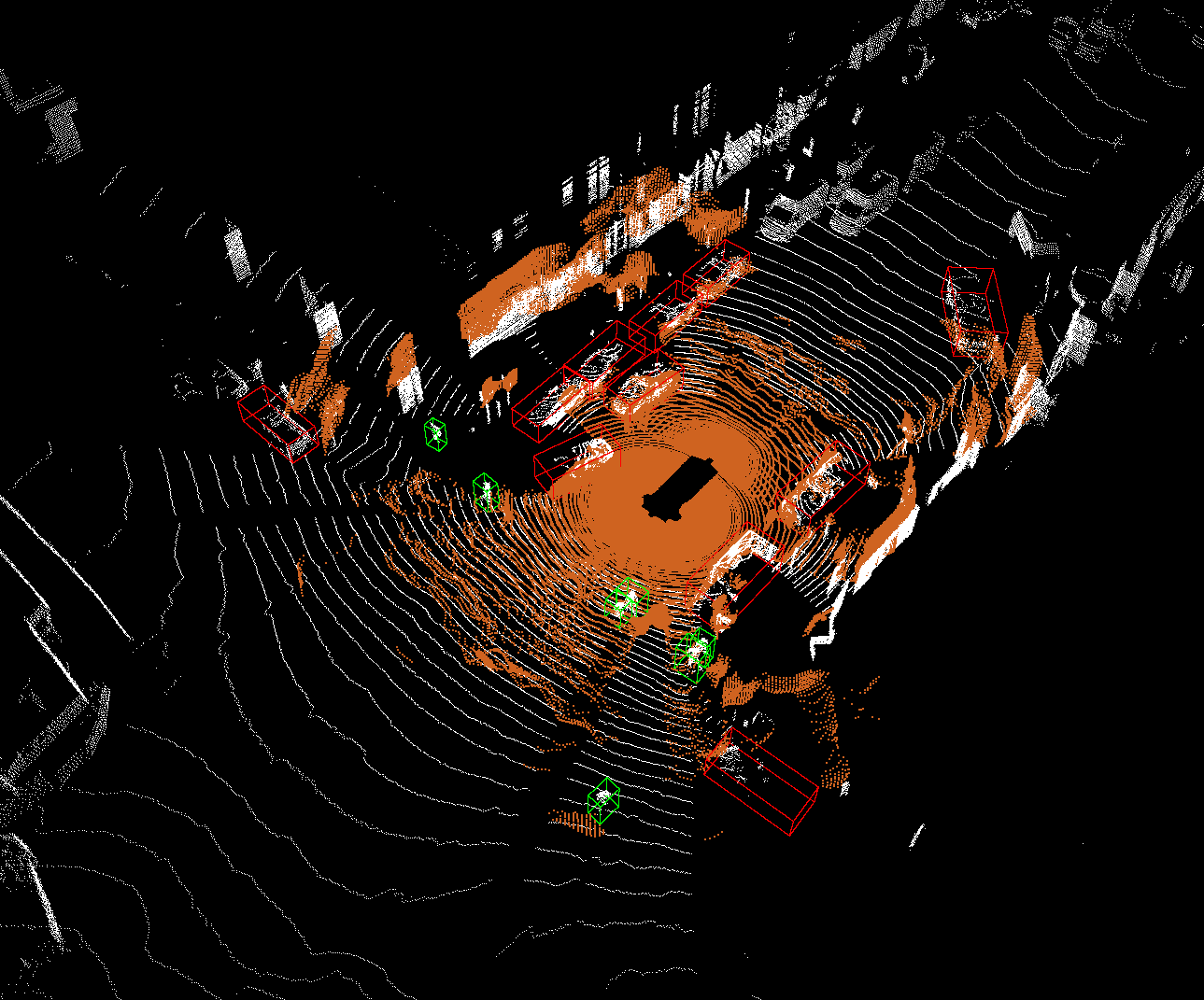}
    \end{subfigure}
    \begin{subfigure}[t]{0.48\textwidth}
        \centering
        \includegraphics[width=\textwidth]{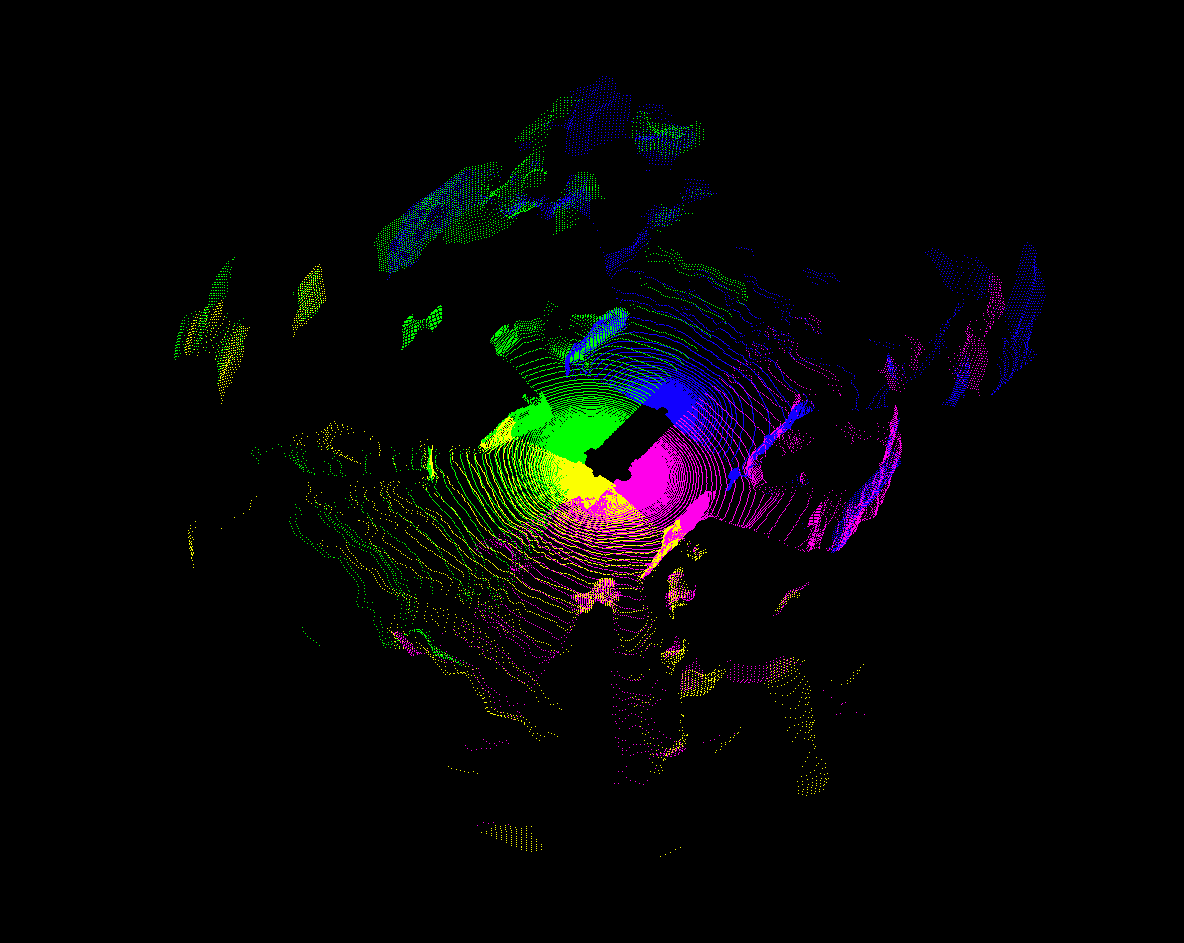}
    \end{subfigure}
    \caption{%
    \textbf{Examples of Waymo point-cloud coverage.}
    \textit{Left}: Point cloud from each SIDE LiDAR sensor, illustrating the partial horizontal coverage provided by each individual sensor.
    \textit{Right}: Point cloud obtained by fusing the four SIDE LiDAR sensors, shown in comparison with the TOP LiDAR\@.
    The SIDE sensors are color-coded by position: front (\textcolor{yellow}{yellow}), left (\textcolor{pink}{pink}), rear (\textcolor{blue}{blue}), and right (\textcolor{green}{green}).
    }%
    \label{fig:waymo_example_scene}
    \vspace{-0.5cm}
\end{figure*}

Figures~\ref{fig:waymo_example_range} and~\ref{fig:waymo_example_scene} show typical range images and corresponding point clouds from the TOP sensor, covering $360^\circ$ horizontally, up to 75m range, and 64 vertical beams; and from the four SIDE sensors, each with a larger vertical FOV, 200 beams, and a maximum range of 20m. The precise horizontal field of view (FOV) for each SIDE sensor is not specified in the official documentation. We therefore estimate each sensor's effective FOV as the proportion of valid columns (i.e., columns with at least one valid measurement) in the range image, out of 600 columns. Assuming a total sensor FOV of $X^\circ$, the effective FOV is computed as $\frac{\#\text{valid columns}}{600} \times X^\circ$.

In all experiments, we use only the four \emph{side} \lidar{} sensors to emulate a multi-sensor input scenario.

Waymo provides a large corpus of high-quality annotated 3D LiDAR bounding boxes. We use annotations for three semantic classes: \texttt{VEHICLE}, \texttt{PEDESTRIAN}, and \texttt{CYCLIST}. Since boxes are labeled in the vehicle (ego) frame, we project them into the \lidar{} frame by inverting the extrinsic calibration matrix. As SIDE \lidar{} sensors capture only short-range points, we retain boxes fully contained within each sensor's valid range.

The resulting dataset contains approximately 1.45M \lidar{} box annotations across over 1000 scenes from geographically distinct regions. Each scene is 20 seconds at 10Hz, totaling 200 frames per scene. For input construction, each scene is divided into overlapping clips of 17 frames with a stride of 5 frames, so consecutive clips share up to 12 frames.

\begin{figure}[htb!]
\centering
\includegraphics[width=0.8\columnwidth]{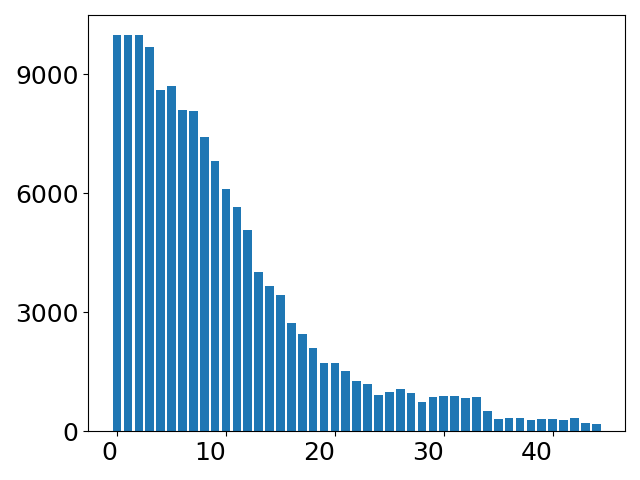}
\caption{\textbf{Per-frame} annotations.}%
\label{fig:waymo_per_frame}
\end{figure}

Statistics for our training set using only SIDE LiDAR sensors are provided in Table~\ref{tab:waymo_anno_stats},~\ref{tab:waymo_split} and Figure~\ref{fig:waymo_per_frame}.

\begin{table}[htb!]
\centering
\caption{\textbf{Statistics} on annotations.}%
\label{tab:waymo_anno_stats}

\begin{tabular}{lcc}
\toprule
 & TrainVal & Test\\
\midrule
\texttt{Vehicle} & 941.6K & 227K \\
\texttt{Pedestrian} & 499.1K & 121.4K \\
\texttt{Cyclist} & 11.8K & 3.4K \\
\midrule
Total & 1451.6K & 352K \\
\bottomrule
\end{tabular}

\end{table}

\begin{table}[htb!]
\centering
\caption{\textbf{Size} of the data splits.}%
\label{tab:waymo_split}

\begin{tabular}{lccc}
\toprule
 & Train & Val & Test\\
\midrule
Scenes & 798 & 202 & 150 \\
Frames & 158K & 40K & 30K \\
Clips & 21K & 5K & 5K \\
\bottomrule
\end{tabular}

\end{table}

\begin{figure*}[t]
    \centering

    \begin{subfigure}[b]{0.24\textwidth}
        \centering
        \includegraphics[width=\textwidth,height=2.5cm,keepaspectratio]{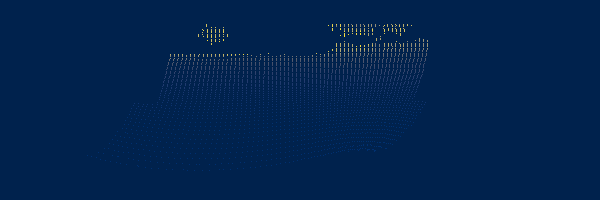}
    \end{subfigure}
    \begin{subfigure}[b]{0.24\textwidth}
        \centering
        \includegraphics[width=\textwidth,height=2.5cm,keepaspectratio]{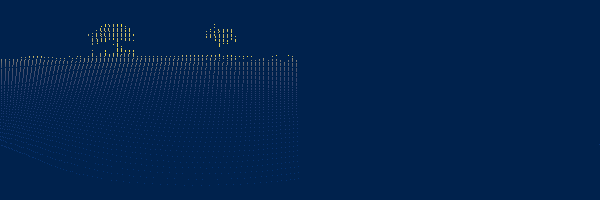}
    \end{subfigure}
    \begin{subfigure}[b]{0.24\textwidth}
        \centering
        \includegraphics[width=\textwidth,height=2.5cm,keepaspectratio]{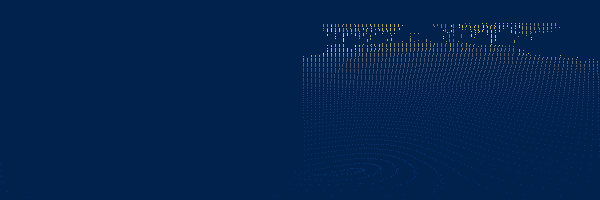}
    \end{subfigure}
    \begin{subfigure}[b]{0.24\textwidth}
        \centering
        \includegraphics[width=\textwidth,height=2.5cm,keepaspectratio]{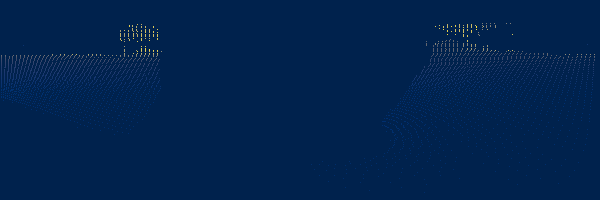}
    \end{subfigure}

    \vspace{0.25em} 

    \begin{subfigure}[b]{0.24\textwidth}
        \centering
        \includegraphics[width=\textwidth,height=2.5cm,keepaspectratio]{assets/graphics/waymo_example_front_dense.png}
    \end{subfigure}
    \begin{subfigure}[b]{0.24\textwidth}
        \centering
        \includegraphics[width=\textwidth,height=2.5cm,keepaspectratio]{assets/graphics/waymo_example_left_dense.png}
    \end{subfigure}
    \begin{subfigure}[b]{0.24\textwidth}
        \centering
        \includegraphics[width=\textwidth,height=2.5cm,keepaspectratio]{assets/graphics/waymo_example_right_dense.png}
    \end{subfigure}
    \begin{subfigure}[b]{0.24\textwidth}
        \centering
        \includegraphics[width=\textwidth,height=2.5cm,keepaspectratio]{assets/graphics/waymo_example_rear_dense.png}
    \end{subfigure}
    \caption{\textbf{Densification on the Waymo SIDE \lidar{} range images}. The top row shows the original sparse range images from Waymo, while the bottom row shows the densified results. The densification algorithm~\cite{ku2018defense} fills in missing pixels and significantly enhances the visual quality and completeness of the range images.}%
    \label{fig:waymo_example_densify}
\end{figure*}

\noindent \textbf{Point Cloud Densification}. The SIDE \lidar{} range images in Waymo are sparse and contain many invalid pixels. Following~\cite{ku2018defense}, we apply a fast CPU-based depth-completion algorithm to fill invalid pixels with estimated depths. The method first performs nearest-neighbor dilation using a diamond-shaped kernel, and then refines the result via smoothing steps that encourage locally planar surfaces and suppress outliers with bilateral filtering. As shown in Figure~\ref{fig:waymo_example_densify}, densification substantially improves the visual quality of the resulting depth maps and point clouds.

\section{Implementation Details}%
\label{app:implementation}

\paragraph{Data Pre-processing \& Post-processing.} For point cloud generation, each LiDAR scan is optionally projected to a dense 2D range image using spherical projection:
Given a 3D point $(x, y, z)$, the corresponding spherical coordinates are computed as:
\begin{equation}
\label{eq:xyz_to_range}
\begin{aligned}
    d = \sqrt{x^2 + y^2 + z^2}, \\
\theta = \arcsin\left(\frac{z}{d}\right), \\
\phi = \arctan2(-y, x).
\end{aligned}
\end{equation}
where $d$ is the measured range, $\theta$ is the elevation angle, and $\phi$ is the azimuth angle. For the Waymo dataset, range images are provided directly, and this step is skipped. To address sparsity in the raw range images, we use the \texttt{ip\_basic} implementation\footnote{\url{https://github.com/kujason/ip_basic}} for efficient depth completion.

Each valid range measurement $d$ is normalized to $[-1, 1]$ via linear scaling:
\begin{equation}
d^{\mathrm{norm}} = \frac{d}{d_{\max}} \times 2 - 1,
\label{eq:linear_norm}
\end{equation}
where $d_{\max}$ is the maximum valid range for the sensor. For generation, we apply the inverse mapping.

Predicted range images are projected back to 3D space using the inverse of the spherical mapping:
\begin{equation}
\label{eq:range_to_xyz}
x = d \cdot \cos\theta \cdot \cos\phi, \quad
y = -d \cdot \cos\theta \cdot \sin\phi, \quad
z = d \cdot \sin\theta,
\end{equation}
where $d$ is the depth value at the pixel, and $\theta$, $\phi$ are the vertical and horizontal angles for that pixel.
 Points with range $d$ outside the post-processing interval are filtered out:
$[0.5, 20.0]$ for Waymo SIDE LiDAR, and $[2.0, 95.0]$ for nuScenes.

For BEV controllable generation, we project 3D bounding boxes and other semantic elements from the ego-vehicle frame onto the BEV plane. Given a bounding box with corners $\{\mathbf{b}_i^{\text{ego}}\}$ in the ego frame, we apply a fixed ego-to-BEV transformation.
Boxes and other features are then rasterized onto a discrete 2D grid, where the grid bounds and cell size are chosen according to the dataset. We also incorporate BEV roadmap layers (e.g., road, lanes), which are available from both Waymo and nuScenes.

\paragraph{Training Procedures.}

\lname{} follows a three-stage latent generative training protocol.

\noindent\textbf{VAE Pretraining}.
We first pre-train the VAE using a reconstruction objective on range images, encoding each LiDAR sensor independently into a compact latent representation. After convergence, the VAE weights are frozen.

\noindent \textbf{Latent Rectified Flow Training}.
We then train a Rectified Flow model to learn the dynamics in the latent space. For sequence modeling, latent representations from all time steps ($T$) and sensors ($S$) are stacked along the batch dimension ($B$), enabling joint spatio-temporal learning directly in latent space.

\noindent \textbf{ControlNet Training}.
Finally, we freeze the base Rectified Flow model and train a ControlNet branch~\cite{zhang2023adding} for conditional generation. Control blocks are inserted into the backbone at a fixed interval (one every three backbone blocks). The model is conditioned on structural inputs, including BEV maps with projected 3D bounding boxes, enabling controllable scene generation while preserving the pretrained generative prior.

\paragraph{Training Hyper-parameters.}
The VAE is trained for 50 epochs using the AdamW optimizer with a base learning rate of $4.5\times10^{-6}$ and a batch size of 16. We employ an $\ell_1$ reconstruction objective with a KL regularization weight of $1\times10^{-6}$. The encoder---decoder architecture uses a latent dimensionality of 4 channels, a base channel width of 64 with multipliers $[1,2,2,4]$, and two residual blocks per resolution level with SiLU activations.
Circular padding is adopted to respect the periodic structure of range images, and no attention layers are used in the VAE\@.
Mixed-precision (FP16) training is used throughout.

The rectified flow model is trained with an $\ell_2$ loss and a log-normal noise schedule for 500000 iterations using a batch size of $16$. We use AdamW optimizer with learning rate $1{\times}10^{-4}$ for nuScenes, $8{\times}10^{-5}$ for Waymo, $\beta_1=0.9$, $\beta_2=0.999$, $\epsilon=10^{-8}$. The learning rate is linearly warmed up over the first 0.01\% of steps and then held constant. Gradients are clipped to a maximum norm of $0.1$. Mixed-precision (FP16) training is used throughout, with FSDP and zero3 sharding for memory efficiency.

During diffusion and ControlNet training, we set the MM-DiT patch size to 1.
Multiple conditioning modalities are used, including text, images, frame pose, and BEV projections of 3D bounding boxes for explicit scene-level guidance.
To enhance robustness and generalization, condition dropout is applied independently to each condition with probability $0.15$; for short clips (fewer than 5 frames), all conditions are jointly dropped with probability $0.15$. BEV maps are embedded into the VAE latent space using a spatial encoder consisting of 5 layers of 2D convolutional blocks with residual connections.

For ControlNet, 30\% of the dataset is set aside for training, as reusing the diffusion model data for this stage leads to minimal improvements.
\paragraph{Sampling and Inference.}
For generation, we employ rectified flow with a linear transport map for the diffusion process and use the Euler–Maruyama solver for sampling. We run 30 sampling steps with a classifier-free guidance scale of $3.0$. Additional sampling hyperparameters are set as needed.

\section{Evaluation}%
\label{app:metrics}

We evaluate LiDAR sequence generation using complementary metrics that capture: \textit{(1) Per-scene fidelity}, \textit{(2) Temporal rollout realism}, \textit{(3) Temporal consistency} across frames, and \textit{(4) Video-level quality} aggregated over sequences. Unless stated otherwise, all metrics are computed on Waymo using the four SIDE \lidar{} sensors.

\subsection{Scene-level Metrics}%
\label{sec:scene_metrics}
\noindent \textbf{Scene Definition.} We define a \emph{scene} at time $t$ as the fused multi-sensor point cloud obtained by combining the \lidar{} point clouds. Specifically, for each sensor $s\in\{1,\dots,S\}$ (with $S{=}4$ in Waymo and with $S{=}1$ in nuScenes), we project points from the sensor frame into the ego-vehicle frame using the provided extrinsics:
\begin{equation}
{p}_{s}^{\text{ego}} = T_{\text{ego}\leftarrow s}\,{p}_{s}, \qquad s\in\{1,\dots,S\},
\end{equation}
Furthermore, we concatenate the transformed points to form the fused scene point cloud $\mathcal{P}_t^{\text{ego}}=\bigcup_{s=1}^S {p}_{t,s}^{\text{ego}}$.

\noindent \textbf{Sampling Protocol.} 
For statistical reliability, we evaluate a total of 10000 sensor frames (i.e., per-sensor range images) in all scene-level metrics.
Concretely, for models that generate \emph{single frames}, we generate 10000 scenes; each scene comprises independently generated range images (one per sensor) that are re-projected and fused in the ego frame. For models that generate \emph{sequences}, we generate 2000 sequences of length $T=5$ (10000 scenes total), and evaluate each time step as an individual scene.

\noindent \textbf{Distributional Fidelity Metrics.} To quantify scene-level fidelity, we adopt standard distributional similarity metrics widely used in LiDAR generative modeling~\cite{zyrianov2022learning}. For each metric, we compute features for real samples and generated samples, fit Gaussians to their distributions, and measure the divergence between the two distributions.

\noindent \textbf{Fréchet Range Distance (FRD).}
FRD~\cite{zyrianov2022learning} measures the discrepancy between feature distributions of real and generated range images for each sensor. Following prior work, we extract features using a pretrained RangeNet-53~\cite{milioto2019rangenet++} segmentation network\footnote{The pre-trained RangeNet++ is downloaded from \url{https://github.com/PRBonn/lidar-bonnetal}}. Let $\{\mathbf{f}_i^{r}\}_{i=1}^{N}$ and $\{\mathbf{f}_i^{g}\}_{i=1}^{N}$ denote features from real and generated samples, with empirical means $(\boldsymbol{\mu}_r,\boldsymbol{\mu}_g)$ and covariances $(\boldsymbol{\Sigma}_r,\boldsymbol{\Sigma}_g)$. FRD is computed as the Fréchet distance between two Gaussians:
\begin{equation}
\mathrm{FRD} = \|{\mu}_r-{\mu}_g\|_2^2
+\mathrm{Tr}\!\left({\Sigma}_r+{\Sigma}_g
-2\left({\Sigma}_r{\Sigma}_g\right)^{\frac{1}{2}}\right).
\end{equation}

\noindent \textbf{Fréchet Point Volume Distance (FPD).}
To assess realism in 3D, we re-project generated range images into point clouds, and extract their features using a pretrained Sparse Point-Voxel Convolution Network~\cite{qi2017pointnet}\footnote{The pre-trained SPVCNN is downloaded from \url{https://github.com/hancyran/LiDAR-Diffusion/tree/main/lidm/eval}}.
We then compute the Fréchet distance in the same form as above, using point-cloud feature statistics.

\noindent \textbf{Jensen-Shannon Divergence (JSD).}
To compare spatial occupancy patterns, we voxelize each fused ego-frame scene into a BEV occupancy histogram over an $H{\times}W$ grid. Let $P$ and $Q$ denote the \emph{marginal} BEV occupancy distributions obtained by averaging normalized histograms over generated and real scenes. We compute:
\begin{equation}
\begin{aligned}
\mathrm{JSD}(P\|Q)=\frac{1}{2}\mathrm{KL}(P\|M)+\frac{1}{2}\mathrm{KL}(Q\|M),
\\ \text{where } M=\frac{1}{2}(P+Q).
\end{aligned}
\end{equation}

\noindent \textbf{Maximum Mean Discrepancy (MMD).}
We additionally compute the maximum mean discrepancy between the BEV occupancy distributions. Given samples $\{x_i\}_{i=1}^{N}\sim P$ (real) and $\{y_j\}_{j=1}^{N}\sim Q$ (generated), MMD with kernel $k(\cdot,\cdot)$ is:
\begin{equation}
\begin{aligned}
\mathrm{MMD}^2(P,Q)={}&\frac{1}{N(N-1)}\sum_{i\neq i'} k(x_i,x_{i'}) \\
&+\frac{1}{N(N-1)}\sum_{j\neq j'} k(y_j,y_{j'}) \\
&-\frac{2}{N^{2}}\sum_{i,j} k(x_i,y_j).
\end{aligned}
\end{equation}
Unless noted otherwise, we use an RBF kernel on the flattened BEV histograms.

For each fused scene $\mathcal{P}_t^{\text{ego}}$, we compute a BEV occupancy grid by discretizing the plane into uniform cells and marking a cell occupied if it contains at least one point. The resulting binary occupancy map is normalized into a histogram and used to form the marginal distributions for JSD and MMD\@.

\subsection{Temporal Consistency Metrics}%
\label{sec:temporal_metrics}
\textbf{Temporal Transformation Consistency Error (TTCE)} To measure video generation performance, we follow~\cite{liang2026lidarcrafter} and inherit the temporal consistency metrics TTCE and CTC, which quantify how consistently generated sequences preserve temporal motion by comparing frame-to-frame rigid transformations with ground truth. Using Iterative Closest Point (ICP), we compute the estimated transformations $T_t^\text{pred} = [R_t^\text{pred} \mid t_t^\text{pred}]$ between consecutive frames and compare them to the true transformations $T_t^\text{gt} = [R_t^\text{gt} \mid t_t^\text{gt}]$:

\begin{align}
\text{TTCE}_\text{rot} &= \frac{1}{T-1} \sum_{t=1}^{T-1} \left\| R_t^\text{pred} (R_t^\text{gt})^\top - I \right\|_F, \\
\text{TTCE}_\text{trans} &= \frac{1}{T-1} \sum_{t=1}^{T-1} \left\| t_t^\text{pred} - t_t^\text{gt} \right\|_2^2
\end{align}

Smaller TTCE values indicate that the predicted motion closely matches the ground truth, reflecting smoother and more realistic temporal dynamics.

\textbf{Chamfer Temporal Consistency (CTC)}
measures the frame-to-frame geometric smoothness of generated point clouds. Consecutive frames are aligned using the ground-truth transformations:

\begin{equation}
\tilde{P}_{t+1} = (R_t^\text{gt})^{-1} \left( P_{t+1} - t_t^\text{gt} \right)
\end{equation}

The Chamfer Distance between $P_t$ and the aligned frame $\tilde{P}_{t+1}$ is:

\begin{equation}
\begin{aligned}
\text{CD}(P_t, \tilde{P}_{t+1}) ={}&
\frac{1}{|P_t|} \sum_{x \in P_t} \min_{y \in \tilde{P}_{t+1}} \|x - y\|_2^2 \\
& + \frac{1}{|\tilde{P}_{t+1}|} \sum_{y \in \tilde{P}_{t+1}} \min_{x \in P_t} \|y - x\|_2^2
\end{aligned}
\end{equation}

The overall CTC score is the average Chamfer Distance over all consecutive frame pairs:

\begin{equation}
\text{CTC} = \frac{1}{T-1} \sum_{t=1}^{T-1} \text{CD}(P_t, \tilde{P}_{t+1})
\end{equation}

Lower CTC values correspond to smoother temporal transitions and higher frame-to-frame coherence in the generated sequence.

\subsection{Rollout-level Metrics}%
\label{sec:rollout_metrics}
To assess temporal consistency and scene fidelity across a sequence altogether, we evaluate the quality of LiDAR rollouts, i.e.\ the accumulation of generated LiDAR frames over time.

\begin{figure*}[thb!]
    \centering
    \includegraphics[width=\linewidth]{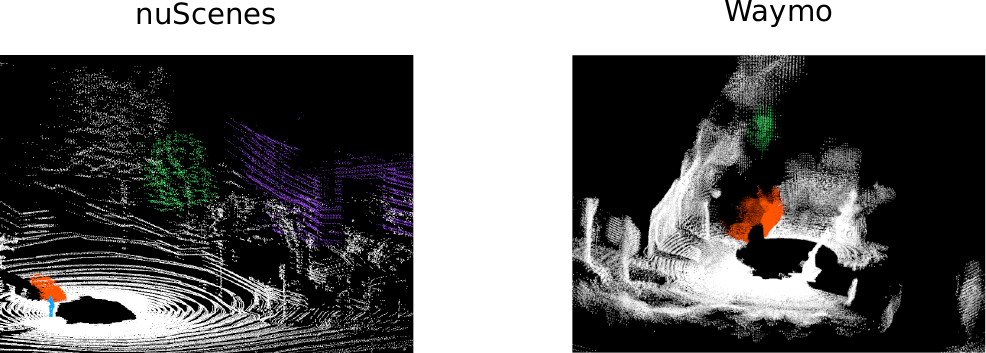}
    \caption{
        \textbf{Rollout examples of temporal consistency on nuScenes and Waymo.}
         For each dataset, we visualize a scene composed of multiple point clouds acquired at different timestamps. The consistent alignment of the colored regions across time indicates that the reconstruction preserves temporal consistency. The highlighted objects include a vehicle (\textcolor{orange}{orange}), a pedestrian (\textcolor{blue}{blue}), vegetation (\textcolor{green}{green}), and background structures (\textcolor{magenta}{purple}).
    }%
    \label{fig:rollout_example}
    \vspace{-1.0em}
\end{figure*}

\noindent \textbf{Global Alignment and Accumulation.}
Following the KITTI-360 protocol~\cite{liao2022kitti}, at each time step $t$ we transform each frame's generated point cloud $\mathcal{P}^{\text{ego}}_t$ to the global coordinate frame using the ground-truth ego-pose $T_t$:
\begin{equation}
\widehat{\mathcal{P}}_t^{\text{global}} = \left\{ T_t {p}_{t,s}^{\text{ego}} : {p}_{t,s}^{\text{ego}} \in \mathcal{P}^{\text{ego}}_t \right\}.
\end{equation}
A rollout for a sequence of length $T$ is then defined as the set union of the aligned point clouds:
\begin{equation}
\mathcal{P}^{\text{rollout}}_{\text{gen}} = \bigcup_{t=1}^T \widehat{\mathcal{P}}_t^{\text{global}}.
\end{equation}
We compare rollout-level statistics using both generated and ground-truth sequences $\mathcal{P}^{\text{rollout}}_{\text{gen}}$ and $\mathcal{P}^{\text{rollout}}_{\text{real}}$. This procedure captures static scene structure (e.g., buildings, road geometry) as well as the spatiotemporal evolution of dynamic objects (e.g., vehicles, pedestrians) (Figure~\ref{fig:rollout_example}).

We compute the Jensen-Shannon divergence (JSD) and maximum mean discrepancy (MMD) as described in Sec.~\ref{sec:scene_metrics}, but using BEV histograms aggregated over the combined rollout point cloud.  

We further quantify geometric fidelity between the ground-truth and generated rollout point clouds using the symmetric Chamfer Distance (CD):
\begin{align}
\mathrm{CD}\big(\mathcal{P}_{\mathrm{real}}, \mathcal{P}_{\mathrm{gen}}\big) = 
& \; \frac{1}{|\mathcal{P}_{\mathrm{real}}|} \sum_{{p} \in \mathcal{P}_{\mathrm{real}}} \min_{{\hat{p}} \in \mathcal{P}_{\mathrm{gen}}} \|{p} - {\hat{p}}\|_2^2 \nonumber \\
& + \frac{1}{|\mathcal{P}_{\mathrm{gen}}|} \sum_{\mathbf{\hat{p}} \in \mathcal{P}_{\mathrm{gen}}} \min_{{p} \in \mathcal{P}_{\mathrm{real}}} \|{\hat{p}} - {p}\|_2^2
\end{align}
where $\mathcal{P}^{\text{rollout}}_{\text{real}}$ and $\mathcal{P}^{\text{rollout}}_{\text{gen}}$ denote the accumulated ground-truth and model-generated rollout point clouds, respectively. 

These rollout-level metrics provide a holistic measure of both structural realism and temporal coherence in generative LiDAR modeling.

\subsection{Video Evaluation Metrics}%
\label{sec:video_metrics}
The temporal consistency metrics (TTCE and CTC) discussed above primarily assess rigid geometric alignment between consecutive frames. However, these metrics do not adequately capture the quality of dynamic scenes in which objects undergo non-rigid motion or deformation. To address this limitation, we adapt established video quality metrics from the computer vision literature to the \lidar{} domain.

We evaluate video generation quality through three metrics organized across two complementary dimensions:

\noindent \textbf{Temporal Consistency.} 
To quantify temporal stability without relying on rigid transformations, we employ an optical-flow warping error $E_w$ and a relative Blind Video temporal Consistency index (BVC)~\cite{yin2025asurvey}. We estimate dense optical flow~\cite{fang2024single} between consecutive frames, warp the previous frame toward the current frame using the estimated flow field, and compute a masked photometric discrepancy on non-occluded pixels. We aggregate both short-range errors (between consecutive frames) and long-range errors (relative to the first frame) to obtain $E_w$. The BVC index is defined as $\mathrm{BVC} = E_w^{\mathrm{gt}} / E_w^{\mathrm{pred}}$, providing a normalized measure where $\mathrm{BVC} > 1$ indicates superior temporal smoothness compared to ground truth, $\mathrm{BVC} = 1$ indicates matching stability, and $\mathrm{BVC} < 1$ indicates increased flicker or temporal instability.

\noindent \textbf{Video-level Realism and Motion.}
At the distributional level, Fr\'echet Video Distance (FVD)~\cite{unterthiner2018towards} compares real and generated videos in the feature space of an I3D network~\cite{carreira2017quo} using 16-frame clips at $224\times224$ resolution. Complementing FVD's appearance-focused evaluation, Fr\'echet Video Motion Distance (FVMD)~\cite{liu2024fr} compares distributions of tracked keypoint trajectories, providing a motion-centric assessment of temporal dynamics.

\section{Ablation Studies}%
\label{app:ablation}
We conduct ablations on the VAE design choice, the diffusion backbone, and other key designs for their validation. Unless specified, all ablation experiments are conducted on the Waymo Open Dataset~\cite{sun2020scalability}, for \emph{multi-sensor \lidar{} scene generation ($t=1$)} setting.

\subsection{VAE}%
\label{sec:ablate_vae}
Our framework adopts the Latent Diffusion~\cite{rombach2022high} paradigm, employing a VAE for compressing high-dimensional range images into a compact latent space. We base our VAE design on recent state-of-the-art architectures, LiDM~\cite{ran2024towards} and RangeLDM~\cite{hu2024rangeldm}, and systematically study the impact of key modeling choices.

All ablations are performed with a fixed compression ratio $f=64$, a batch size of 16, and 50 training epochs, following the default settings for each method's official implementation. 
Following~\cite{li2025uniscene,ni2025maskgwm}, we adopt an image-based VAE (2D) instead of a temporal 3D architecture, as the goal here is efficient per-frame compression of range images rather than joint spatiotemporal modeling.
This architecture also 
enables more efficient training and inference, similar to designs in~\cite{li2025uniscene,ni2025maskgwm}. 

\begin{figure*}[htbp]
    \centering
    \begin{subfigure}[t]{\textwidth}
        \centering
        \includegraphics[width=\textwidth]{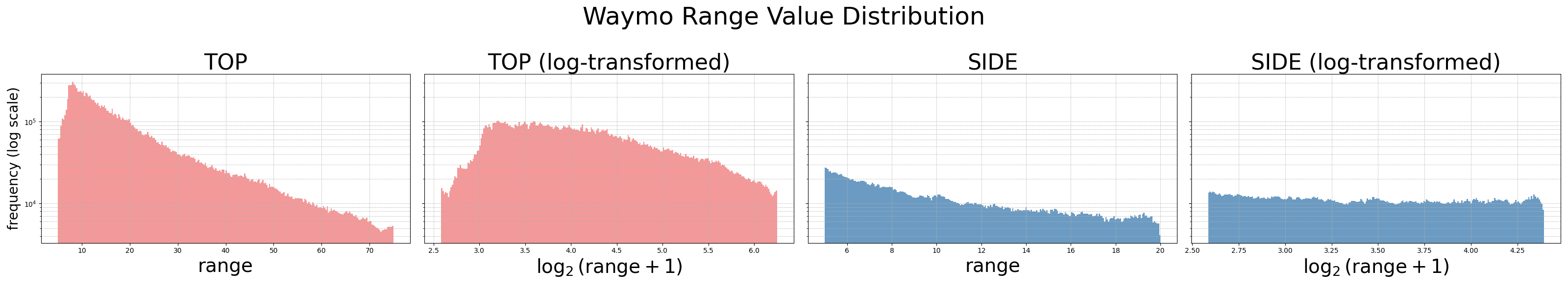}
    \end{subfigure}
    \caption{\textbf{Impact of Logarithmic Transformation on LiDAR range distributions.} \textit{Left:} For TOP \lidar{}, the log transform significantly reduces the skewness of high dynamic range values. \textit{Right:} For SIDE \lidar{} on Waymo, where the range values are less skewed and cover a shorter distance, the effect of the log transform is minimal.}%
    \label{fig:waymo_range_dist}
\end{figure*}

Table~\ref{tab:ablate_vae_multi_sensor_waymo} presents quantitative VAE reconstruction results for different configurations on Waymo SIDE LiDAR\@.
We observe the following: (1) The use of \emph{Circular Convolution}~\cite{schubert2019circular} in RangeLDM leads to improved latent representations relative to LiDM across all metrics (cf.\ rows 3--6 vs 1--2), confirming the benefit of circularity for sensors with $360^\circ$ coverage.
(2) Incorporating adversarial loss into VAE training is \emph{challenging}, with performance highly sensitive to the activation schedule; improper tuning can significantly degrade results (cf.\ rows 1--2 and 3--4).
(3) VAE performance deteriorates with aggressive input resizing, likely due to increased information loss (cf.\ rows 3--4 vs rows 5--6).
Nevertheless, this resizing may be necessary for \emph{compatibility with U-Net-based diffusion} backbones, which constrain input resolution and highlight a limitation of the U-Net design.
(4) For SIDE LiDAR on Waymo (range 0--20 m), log compression (Equation~\eqref{eq:log_norm}) provides little benefit, since the dynamic range is limited and linear normalization (Equation~\eqref{eq:linear_norm}) already preserves the range scale adequately without introducing an additional nonlinear transformation (cf.\ rows 6--7 and Figure~\ref{fig:waymo_range_dist}).

\begin{table*}[t]
\begin{center}
\caption{\textbf{VAE ablation results on Waymo SIDE LiDAR.} Reconstruction quality is compared across VAE backbones and normalization strategies. \textit{Configuration $\dagger$}: compression rate $f=64$, range densification, and filling invalid points. $\mathcal{L}_{\text{Desc}}$ denotes the discriminator (adversarial) loss. Performance is evaluated using perceptual metrics: R-FSVD, R-FPVD and statistical metrics: Chamfer Distance (CD), Earth Mover’s Distance (EMD), following~\cite{ran2024towards}. The best result is highlighted in \textbf{bold} and the second-best in \underline{underline}.}%
\label{tab:ablate_vae_multi_sensor_waymo}
\resizebox{\textwidth}{!}{
    \begin{tabular}{lccccHcccc}
    \toprule
     \textbf{Method} & \textbf{Compression} & \textbf{Input Size} & \textbf{Transform} & \textbf{Variant} & \textbf{R-FRID} $(10^{-3})\downarrow$ & \textbf{R-FSVD} $\downarrow$ & \textbf{R-FPVD} $\downarrow$ & \textbf{CD} $(10^{-3})\downarrow$ & \textbf{EMD} $(10^{-3})\downarrow$ \\
    \midrule
    \multirow{2}{*}{LiDM\cite{ran2024towards}} & $f_c=4,f_p=4 ^{\dagger}$ & 208$\times$640 & LogNorm & --- & 46.91 & 426.71 & 191.74 & 505.98 & 102.68 \\
     & $f_c=4,f_p=4 ^{\dagger}$ & 208$\times$640  & LogNorm & \textcolor{blue}{w/o $\mathcal{L}_{Desc}$} & 1.02 & 1.83 & 0.32 & 2.81 & 16.25 \\
    \midrule
    \multirow{4}{*}{RangeLDM\cite{hu2024rangeldm}} & $f_p=8^{\dagger}$ & 224$\times$608 & LogNorm & --- & 0.14 & 3.94 & 0.31 & 3.43 & 23.96 \\
     & $f_p=8^{\dagger}$ & 224$\times$608 & LogNorm & \textcolor{blue}{w/o $\mathcal{L}_{Desc}$} & 0.04 & \underline{1.26} & 0.19 & 0.98 & 7.39 \\
     & $f_p=8^{\dagger}$ & 200$\times$600 & LogNorm & --- & --- & 3.55 & 0.27 & 1.27 & 13.11 \\
     & $f_p=8^{\dagger}$ & 200$\times$600 & LogNorm & \textcolor{blue}{w/o $\mathcal{L}_{Desc}$} & --- & 1.44 & \underline{0.15}& \underline{{0.49}} & \underline{4.40} \\
    \midrule
    \rowcolor[HTML]{F5D0CB}
    \lname{} VAE &$f_p=8^{\dagger}$ & 200$\times$600 & \textcolor{blue}{Linear} & \textcolor{blue}{w/o $\mathcal{L}_{Desc}$} & --- & \textbf{{0.41}} & \textbf{{0.13}} & \textbf{0.48} & \textbf{{4.37}} \\
    \bottomrule
    \end{tabular}
}
\end{center}
\end{table*}

\subsection{Diffusion Backbone: DiT vs U-Net}%
\label{sec:ablate_diffusion}
For LiDAR generation in the Latent Diffusion framework~\cite{rombach2022high}, the choice of diffusion backbone is critical. Range images, representing $360^{\circ}$ LiDAR sweeps as anisotropic 2D panoramas, exhibit a pronounced width–height asymmetry: the width spans the full angular resolution, whereas the limited number of sensor beams constrains the height. This results in unique long-range horizontal dependencies that differ fundamentally from those in natural images. In this section, we ablate U-Net and Diffusion Transformer (DiT)~\cite{peebles2023scalable} backbones, comparing their effectiveness in modeling these dependencies and their parameter efficiency.

\begin{figure*}[t]
    \begin{subfigure}[b]{0.48\textwidth}
        \centering
        {\small Original $(200,600)$}
        \includegraphics[width=\textwidth,height=2.5cm,keepaspectratio]{assets/graphics/waymo_example_front_dense.png}
    \end{subfigure}
    \begin{subfigure}[b]{0.48\textwidth}
        \centering
        {\small Resized $(224,608)$}
        \includegraphics[width=\textwidth,height=2.5cm,keepaspectratio]{assets/graphics/waymo_example_front_dense.png}
    \end{subfigure}

    \begin{subfigure}[b]{0.48\textwidth}
        \centering
        \includegraphics[width=\textwidth,height=6.cm]{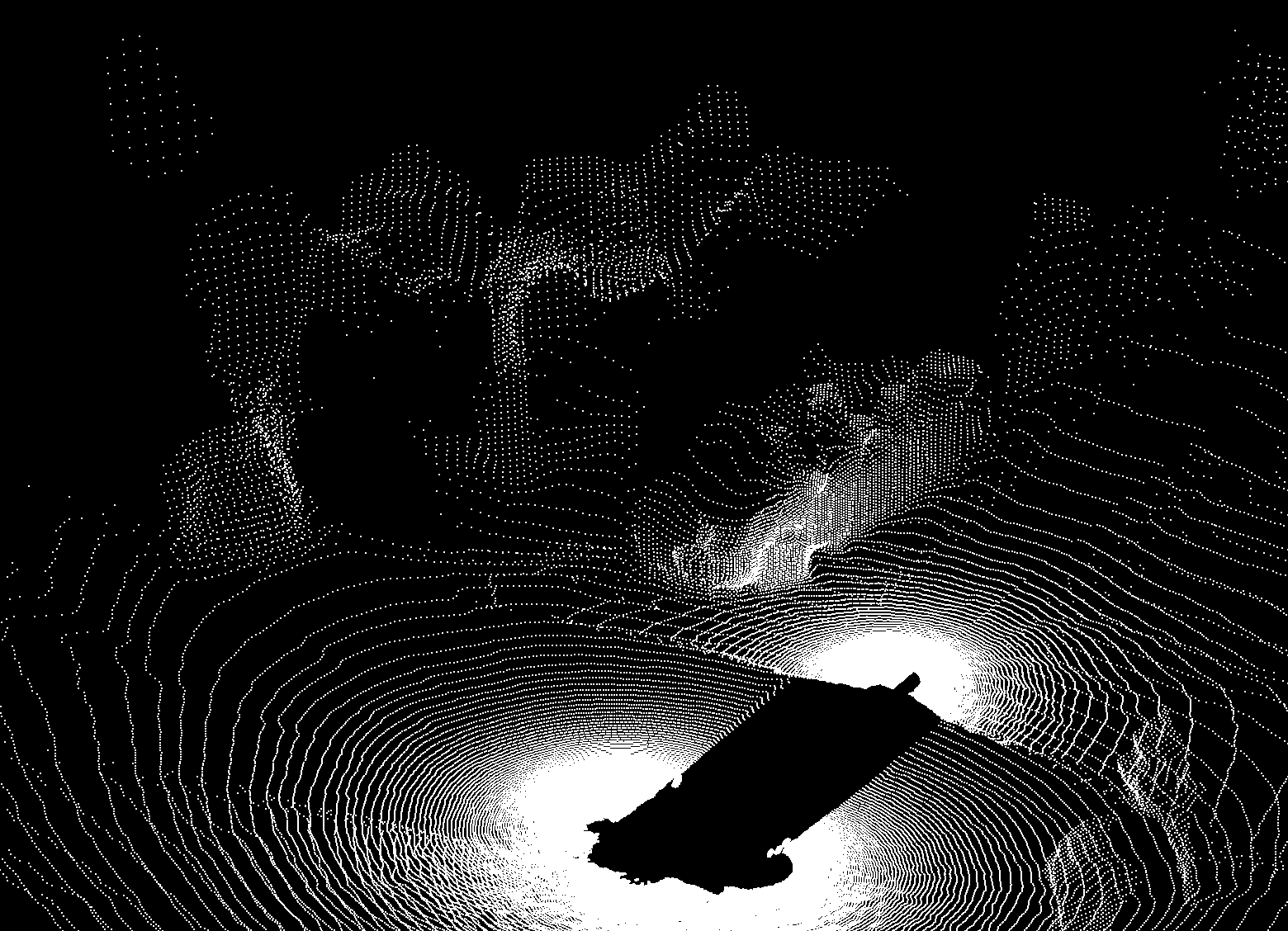}
    \end{subfigure}
    \begin{subfigure}[b]{0.48\textwidth}
        \centering
        \includegraphics[width=\textwidth,height=6.cm]{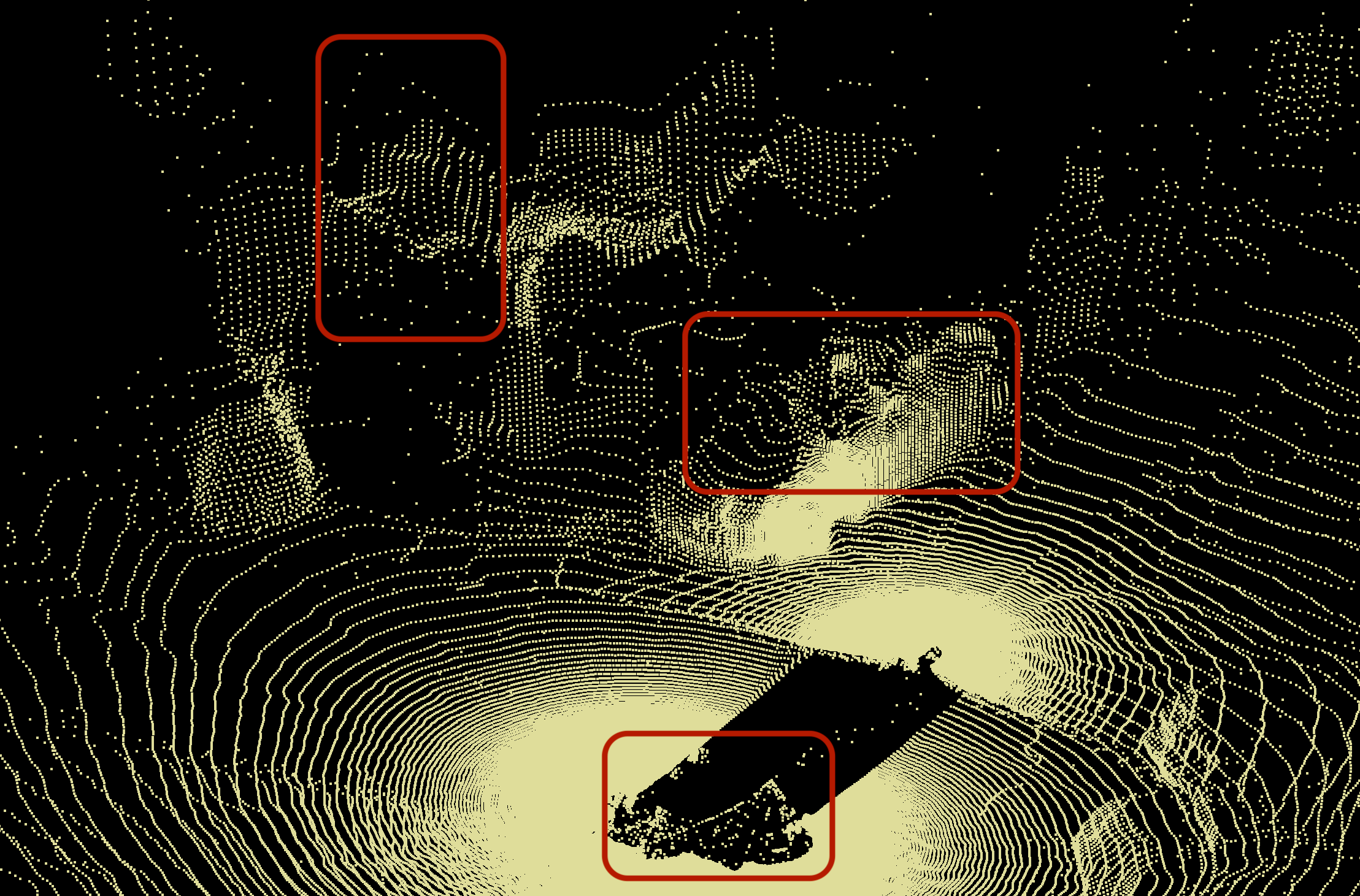}
    \end{subfigure}
    \caption{\textbf{Effects of range image resizing on generated point clouds.} Resizing the range image from $(200,600)$ to $(224,608)$ introduces noise and geometric distortions in the resulting point cloud, noticeably affecting object shapes (e.g., trees, cars).\ \emph{Best viewed when zoomed in}.}%
    \label{fig:waymo_example_distortion_by_resize}
\end{figure*}

While U-Net architectures remain a popular backbone that preserves local structures, they are inherently limited in modeling the extreme aspect ratios and horizontal context of range images. In contrast, DiT leverages global self-attention to capture long-range dependencies directly and offers flexible input resolution via dynamic padding, therefore avoiding the destructive resizing often required by the U-Net. As observed in Section~\ref{sec:ablate_vae} and Figure~\ref{fig:waymo_example_distortion_by_resize}, \emph{maintaining native resolution is crucial}, as resizing artifacts can significantly degrade downstream point cloud reconstruction.

Table~\ref{tab:ablate_dm_multi_sensor_waymo} reports single-frame LiDAR generation results for state-of-the-art methods LiDM~\cite{ran2024towards} and RangeLDM~\cite{hu2024rangeldm} across multiple backbones. LiDM demonstrates strong performance on Waymo SIDE LiDAR generation, notably outperforming RangeLDM (cf.\ rows 1 vs. 3), with further gains when employing DiT. We find that the default hyperparameters of RangeLDM require more tuning (e.g., learning rates), particularly for side LiDARs with smaller dynamic ranges and higher cross-sensor correlation demands.
DiT-based models outperform U-Nets across most settings (cf.\ rows 1 vs. 2 and 3 vs. 4), while using approximately 50\% fewer parameters and reducing GPU memory consumption to about two-thirds of that required by U-Net.
These results highlight the parameter-efficiency advantages of DiT over U-Nets, validating the benefits of transformer-based backbones for LiDAR range image generation.

\begin{table*}[htb!]
\begin{center}
\caption{\textbf{Ablations on Diffusion Backbone on Waymo SIDE LiDAR}. \textit{Configuration $\dagger$}: compression rate $f=64$, range densification, filling invalid points and \emph{without} discriminator. The best result is highlighted in \textbf{bold} and the second-best in \underline{underline}.}%
\label{tab:ablate_dm_multi_sensor_waymo}
\resizebox{\textwidth}{!}{
    \begin{tabular}{lcccccccc}
    \toprule
     \multirow{2}{*}{\textbf{Method}} & \multirow{2}{*}{\makecell{\textbf{Input}\\\textbf{Size}}} & \multirow{2}{*}{\textbf{Backbone}} & \multicolumn{2}{c}{\textbf{Range}} & \multicolumn{2}{c}{\textbf{Points}} & \multicolumn{2}{c}{\textbf{BEV}} \\
      \cmidrule(l){4-5}
      \cmidrule(l){6-7}
      \cmidrule(l){8-9}
     & & & \textbf{FRD} $\downarrow$ & \textbf{MMD} $\downarrow$ & \textbf{FPVD} $\downarrow$ & \textbf{MMD} $\downarrow$ & \textbf{JSD} $\downarrow$ & \textbf{MMD} $\downarrow$ \\
    \midrule
     \multirow{2}{*}{LiDM$^{\dagger}$~\cite{ran2024towards}} & 208$\times$640 & U-Net & 2.29 & 148.83 & 4.67 & \underline{65.76} & 0.19 & 1.00 \\
                                                                            & 200$\times$600 & \textcolor{blue}{DiT} & \underline{0.45} & 77.91 & \underline{2.52} & 113.33 & \underline{0.15} & \textbf{{1.00}} \\
    \cmidrule{1-9}
     \multirow{2}{*}{RangeLDM$^{\dagger}$~\cite{hu2024rangeldm}} & 224$\times$608 & U-Net & 10.97 & 979.33 & 122.38 & 1236.59 & 0.22 & 1.63 \\
                                                                                & 200$\times$600 & \textcolor{blue}{DiT} & 9.07 & 938.83 & 61.47 & 374.81 & 0.19 & 1.16 \\    
    \cmidrule{1-9}
    \multirow{5}{*}{\textbf{\lname{}}$^{\dagger}$ (t=5)}
& \multirow{5}{*}{200$\times$600}
& \makecell[l]{\textcolor{blue}{DiT}\\{+ Naive-Stacking}}
& 1.66 & {55.67} & 14.82 & 156.55 & 0.20 & 1.62 \\

& & \makecell[l]{\textcolor{blue}{MM-DiT}\\{+ Naive-Stacking}}
&  1.43 & \underline{51.94} & 14.48 & 139.31 & 0.16 & 1.51  \\

& &
\makecell[l]{
\textcolor{blue}{MM-DiT}\\
\textcolor{blue}{+ \lidar{}-Sequence}\\
\textcolor{blue}{+ Inter-Sensor}
}
& \textbf{0.16} & \textbf{11.89} & \textbf{1.32} & \textbf{6.90} & \textbf{0.08} & \underline{1.05} \\
    \bottomrule
    \end{tabular} 
}
\end{center}
\end{table*}

Finally, we perform two ablation studies on the proposed \lname{} architecture. First, we compare DiT and MM-DiT under the same \emph{Naive Stacking} strategy, where temporal and multi-sensor latent representations are simply concatenated before being processed by the backbone. This isolates the contribution of the backbone architecture itself. 
Second, we compare this Naive Stacking baseline against the proposed \emph{\lidar{} Sequence} and \emph{Inter-sensor} modules while keeping the MM-DiT backbone fixed, therefore isolating the contribution of explicit temporal and cross-sensor modeling.
Table~\ref{tab:ablate_dm_multi_sensor_waymo} shows that replacing DiT with MM-DiT under the same \emph{Naive Stacking} setting improves the results by an average of $10.1\%$ across all evaluation metrics.
In contrast, replacing Naive Stacking with the proposed \emph{\lidar{} Sequence} and \emph{Inter-sensor} modules yields substantially larger gains, reducing FRD by $88.8\%$, point-cloud FPD by $90.9\%$, and BEV JSD by $50.0\%$ relative to the MM-DiT baseline.
These results demonstrate that the improvements reported in the main paper are largely driven by the combined proposed temporal and cross-sensor modules rather than by the MM-DiT backbone alone.

\section{Baselines}%
\label{app:baselines}
We compare \lname{} with a broad suite of recent methods for LiDAR generation. Table~\ref{tab:competitors} summarizes the key characteristics of each baseline, including problem setting, network architecture, generative paradigm, datasets used for training, and the availability of code and pretrained checkpoints \emph{at the time of submission}. In contrast to prior work, \lname{} is specifically designed for the challenging \emph{multi-sensor \lidar{} sequence} generation setting and supports a wide range of input conditions.

\begin{table*}[htb!]
\caption{\textbf{Summary of existing methods for LiDAR generation.} Overview of compared methods, indicating sensor setup, generative model, available modalities, supported input conditions, training datasets (\textbf{K}: KITTI-360~\cite{liao2022kitti}, \textbf{N}: nuScenes~\cite{caesar2020nuscenes}, \textbf{W}: Waymo~\cite{sun2020scalability}), and code/checkpoint availability.}%
\label{tab:competitors}
\resizebox{\textwidth}{!}{
\begin{tabular}{lccccccccc}
\toprule 
 & \textbf{Venue} & \textbf{Modality} & \makecell{\textbf{\lidar{}} \\ \textbf{Sensors}}& \makecell{\textbf{\lidar{}} \\ \textbf{Frames}} & \makecell{\textbf{Supported} \\ \textbf{Conditions}} & \makecell{\textbf{Diffusion} \\ \textbf{Model}}  & \textbf{Dataset} & \textbf{Code} & \textbf{Checkpoint}\\
\cmidrule{1-10}
LiDAR Diffusion\cite{ran2024towards} & CVPR’24 & \lidar{} & 1 & 1 & image,semantic map & U-Net & K & \href{https://github.com/hancyran/LiDAR-Diffusion}{Yes} & K \\
RangeLDM\cite{hu2024rangeldm} & ECCV’24 & \lidar{} & 1 & 1 & inpaint,downsample & U-Net & K,N & \href{https://github.com/WoodwindHu/RangeLDM}{Yes} & K,N\\
R2DM\cite{nakashima2024lidar} & ICRA’24 & \lidar{} & 1 & 1 & --- & U-Net & K & \href{https://github.com/kazuto1011/r2dm}{Yes} & K \\
\cmidrule{1-10}
LiDARDM\cite{zyrianov2025lidardm} & ICRA’25 & \lidar{} & 1 & multi & map & U-Net & K,W & \href{https://github.com/vzyrianov/LidarDM}{Yes} & K,W \\
UniScene\cite{li2025uniscene} & CVPR’25 & \makecell{\lidar{}\\\&Camera} & 1 & multi & BEV$\rightarrow$semantic occupancy & Sparse U-Net & N & \href{https://github.com/Arlo0o/UniScene-Unified-Occupancy-centric-Driving-Scene-Generation}{Yes} & N \\
OpenDWM\cite{ni2025maskgwm} & CVPR’25 & \lidar{} & 1 & multi & BEV,map,text & U-Net/DiT& N & \href{https://github.com/SenseTime-FVG/OpenDWM}{Yes} & N \\
LiDARCrafter\cite{liang2026lidarcrafter} & AAAI’26 & \lidar{} & 1 & multi & text$\rightarrow$layout$\rightarrow$object$\rightarrow$trajectory & U-Net & N & \href{https://github.com/worldbench/LiDARCrafter}{Yes} & --- \\
\cmidrule{1-10}
\rowcolor[HTML]{F5D0CB}
\textbf{\lname} & --- & \textbf{\lidar{}} & \textbf{multi} & \textbf{multi} & \textbf{BEV,object,image,text} & \textbf{MM-DiT} & \textbf{N,W} & \multicolumn{2}{c}{---} \\ 
\bottomrule
\end{tabular}
}
\end{table*}

Many existing methods~\cite{zyrianov2022learning,ran2024towards,hu2024rangeldm,nakashima2024lidar} are designed for single-sensor LiDAR generation. To ensure fair comparison, we adapt these methods for multi-sensor LiDAR generation, as detailed in the following sections, which also specify the training hyperparameters we used on Waymo. Certain methods could not be included in our Waymo experiments due to design complexity or prohibitive data requirements. For example, UniScene~\cite{li2025uniscene} requires semantic occupancy annotations, which are not readily available at scale for Waymo. Similarly, LiDARDM~\cite{zyrianov2025lidardm}, a rendering-based technique that relies on Neural Kernel Surface Reconstruction~\cite{huang2023nksr} and ray-casting, is not directly comparable to range-image or voxel-based LiDAR generation.

\subsection{R2DM}
\noindent \textbf{Adaptation for Waymo}.
We adapt R2DM~\cite{nakashima2024lidar} for the Waymo Dataset with four SIDE LiDAR sensors: front, left, rear, and right.
The range image resolution is set to $200 \times 600$ with a single depth channel, and  depth values span between $[1, 20]$ meters.
Ray angles are initialized for a vertical FOV from $30^\circ$ (top) to $-90^\circ$ (bottom). 
For the SIDE LiDAR sensors, we use the effective horizontal FOV estimated from valid range-image columns, following Sec. 6.2.
To handle multiple sensors, we stack their views into a $[B, S, C, H, W]$ tensor ($S=4$ sensors) and reshape to $[B \times S, C, H, W]$ for efficient batch processing as in Figure~\ref{fig:waymo_naive_sensor_stack}.

\noindent \textbf{Training Configuration}. Following the upstream implementation\footnote{\url{https://github.com/kazuto1011/r2dm}}, we train with AdamW (betas 0.9/0.99, weight decay 0.0, $\epsilon{=}10^{-8}$), a base learning rate of $4.5\times10^{-6}$ scaled to $3.6\times10^{-5}$. The learning rate follows a cosine annealing schedule with a 10k-step warmup, and training proceeds for 300k optimization steps. 
Training uses gradient clipping with a maximum norm of 1.0 and an exponential moving average (EMA, $\beta=0.995$) with updates every 10 steps.
The code is run with Python 3.12 and PyTorch version 2.9.0.

\subsection{LiDM}
\noindent \textbf{Adaptation for Waymo}. We adapt LiDM~\cite{ran2024towards} to Waymo SIDE \lidar{} range images by using depth-only inputs. The range images are resized to $208\times640$ to match the downsampling constraints of its U-Net backbone.
Depth values are clipped to $[1, 20]$ for compatibility with the LiDM baseline
and normalized using the logarithmic mapping from the original work:
\begin{equation}
d^{\mathrm{norm}}=\frac{\log_2(d+1)}{\log_2(d_{\max}+1)}\times2-1.
\label{eq:log_norm}
\end{equation}

\begin{figure*}[htb!]
    \centering
    \begin{subfigure}[t]{\textwidth}
        \centering
        \includegraphics[width=\textwidth]{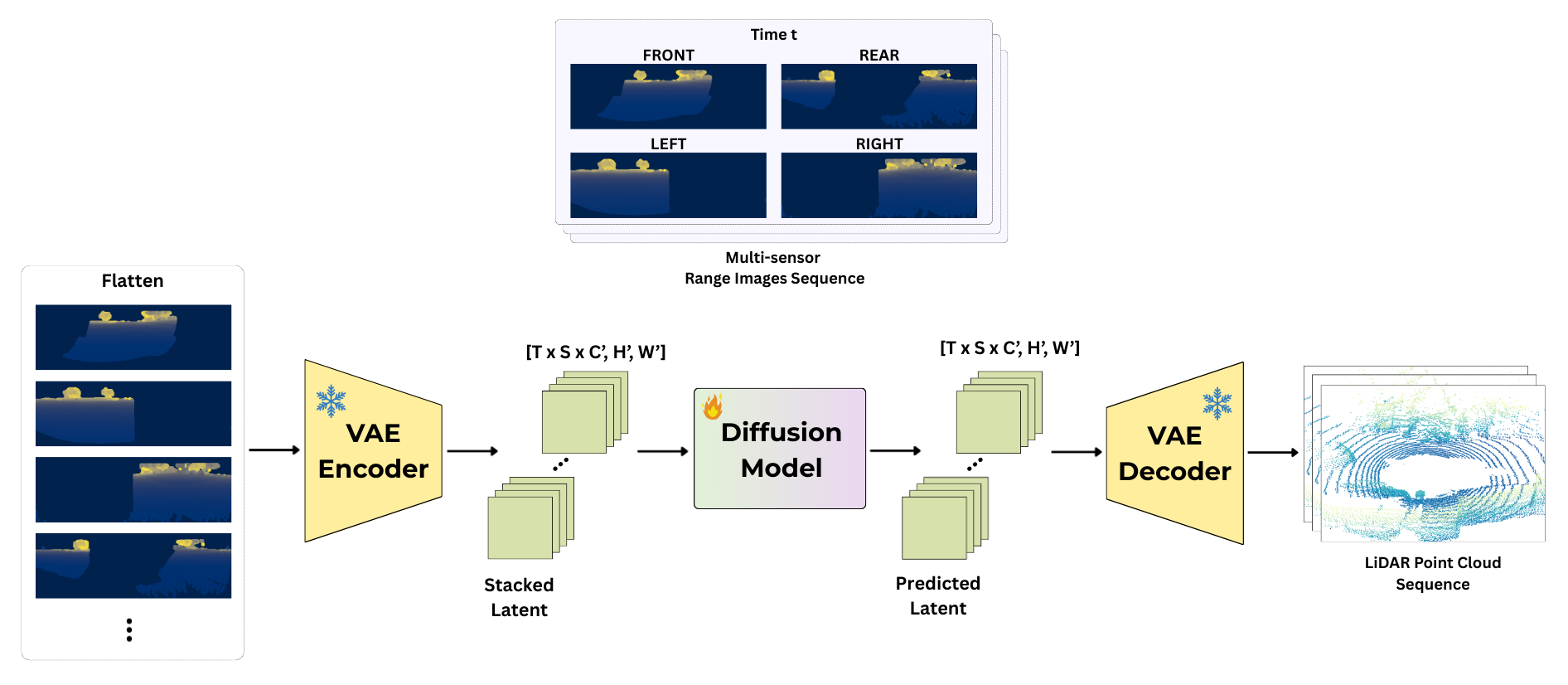}
    \end{subfigure}
    \caption{\textbf{Naive sensor stacking for multi-sensor \lidar{} generation.} Each sensor range image at time step $t$ is independently encoded by the VAE\@. The diffusion model operates on the stacked multi-sensor latents, enabling generation (or prediction) of joint sensor, and can be extended to sequences by stacking along the temporal dimension.}%
    \label{fig:waymo_naive_sensor_stack}
\end{figure*}

To accommodate multiple sensors, we train a single 2D VAE shared across sensors while encoding each sensor independently (i.e., without cross-sensor interaction). We use a compression setting ($f{=}64$) with curve-wise factor $f_c{=}4$ and patch-wise factor $f_p{=}4$, producing latents of size $16\times52\times40$ per sensor. For diffusion training, we stack latents across sensors and time: per-sensor latents are arranged as $[B,S,16,52,40]$ and reshaped to $[B,S{\times}16,52,40]$; for sequences, we additionally fold the temporal dimension $T$ into the channel axis, yielding $[B,T{\times}S{\times}16,52,40]$, as shown in Figure~\ref{fig:waymo_naive_sensor_stack}.

\noindent \textbf{Training configuration}. Unless stated otherwise, we follow the official implementation\footnote{\url{https://github.com/hancyran/LiDAR-Diffusion}}. We retain the adversarial loss in the VAE objective. The VAE (encoder/decoder and discriminator) is trained with Adam (betas 0.5/0.9) and learning rate $4.5\times10^{-6}$; we train the diffusion model with learning rate $1\times10^{-6}$. We train for 300k optimization steps 
with a batch size of 8, and use EMA with decay $\beta{=}0.995$ updated every 10 steps. Experiments are run with Python 3.12 and PyTorch version 2.9.0.

\subsection{RangeLDM}
\noindent \textbf{Adaptation for Waymo}. We adapt RangeLDM~\cite{hu2024rangeldm} to Waymo SIDE \lidar{} range images following the same multi-sensor stacking strategy as for LiDM (Figure~\ref{fig:waymo_naive_sensor_stack}). We use the default RangeLDM compression setting with patch factor $f_p{=}8$, corresponding to an overall compression ratio $f{=}64$.

\noindent \textbf{Training configuration}. We follow the official implementation\footnote{\url{https://github.com/WoodwindHu/RangeLDM}} and train for 500k optimization steps 
(batch size 8 per GPU\@; Python 3.12; PyTorch 2.9.0). In preliminary experiments, we observed training instability when generating Waymo SIDE \lidar{} data: although the VAE achieves strong reconstruction performance (Tables~\ref{tab:ablate_vae_multi_sensor_waymo} and~\ref{tab:ablate_dm_multi_sensor_waymo}), diffusion training may collapse mid-training, leading to a sharp degradation in sample quality without obvious numerical failures. We attribute this sensitivity primarily to the learning rate. While the original work uses $10^{-4}$, we reduce the learning rate to $10^{-6}$ for Waymo; with the original setting, training does not reliably progress.

\subsection{OpenDWM}

\noindent \textbf{Adaptation for Waymo}.  
We adapt the LiDAR generation components provided in the OpenDWM codebase\footnote{\url{https://github.com/SenseTime-FVG/OpenDWM}} to the Waymo SIDE \lidar{} setting.  
Specifically, we use its LiDAR VAE tokenizer together with the temporal DiT diffusion model.  
Because the released configuration is designed for sequence generation from a single \texttt{LIDAR\_TOP} stream, we replace the original single-sensor setup with the four SIDE sensors (front, left, rear, right) and apply the same naive multi-sensor stacking strategy used for LiDM and RangeLDM (Figure~\ref{fig:waymo_naive_sensor_stack}).  
Each sensor is encoded independently using a shared VAE\@.  
The resulting latent tensors are then stacked along the sensor and temporal dimensions before being processed by the temporal diffusion transformer.  
We otherwise retain the point-cloud voxelization and latent parameterization from the official release, which voxelizes the scene within a $[-50,50]\times[-50,50]\times[-3,5]$\,m volume with voxel sizes $(0.15625, 0.15625, 0.125)$ and uses a 32-channel latent representation.  
\begin{figure*}[htb!]
\centering
\begin{subfigure}[b]{0.24\textwidth}
    \centering
    \includegraphics[width=\textwidth,height=2.5cm]{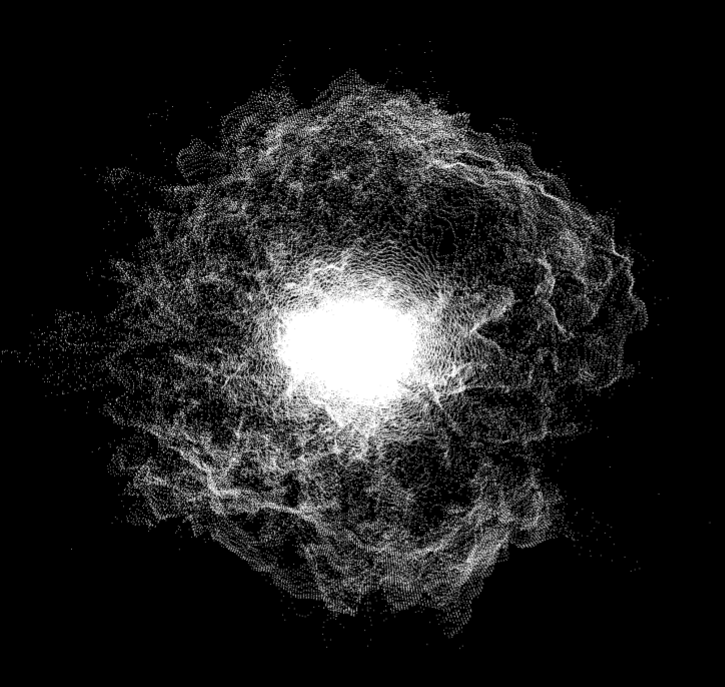}
    \caption*{R2DM}
\end{subfigure}
\begin{subfigure}[b]{0.24\textwidth}
    \centering
    \includegraphics[width=\textwidth,height=2.5cm]{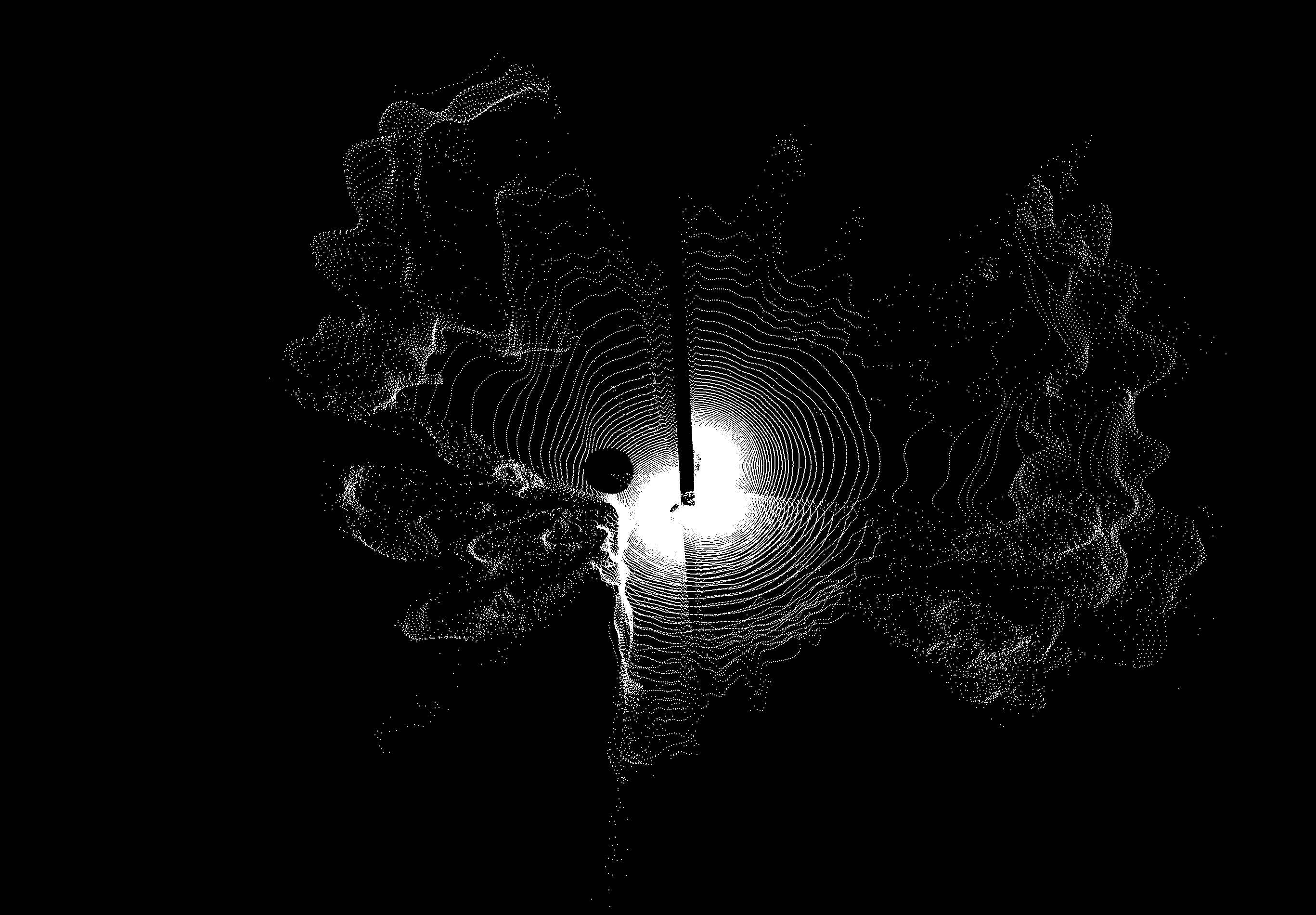}
    \caption*{LiDM}
\end{subfigure}
\begin{subfigure}[b]{0.24\textwidth}
    \centering
    \includegraphics[width=\textwidth,height=2.5cm]{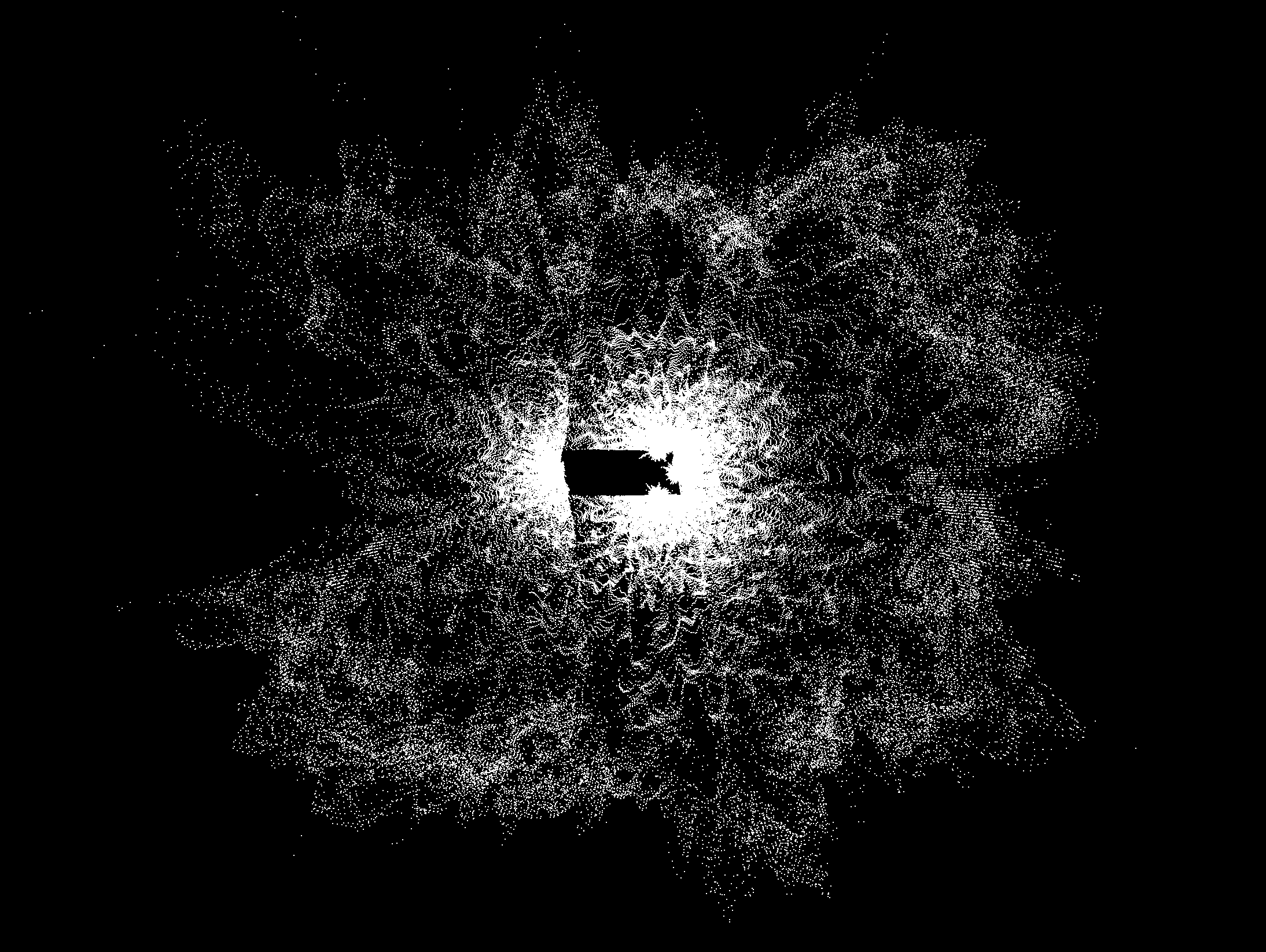}
    \caption*{RangeLDM}
\end{subfigure}
\begin{subfigure}[b]{0.24\textwidth}
    \centering
    \includegraphics[width=\textwidth,height=2.5cm]{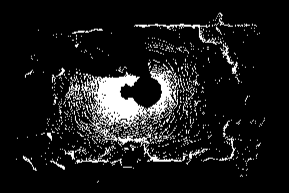}
    \caption*{OpenDWM}
\end{subfigure}

\vspace{0.01cm}
\rule{\linewidth}{0.5pt}
\vspace{0.01cm}

\begin{subfigure}[b]{0.24\textwidth}
    \centering
    \includegraphics[width=\textwidth,height=2.5cm,keepaspectratio]{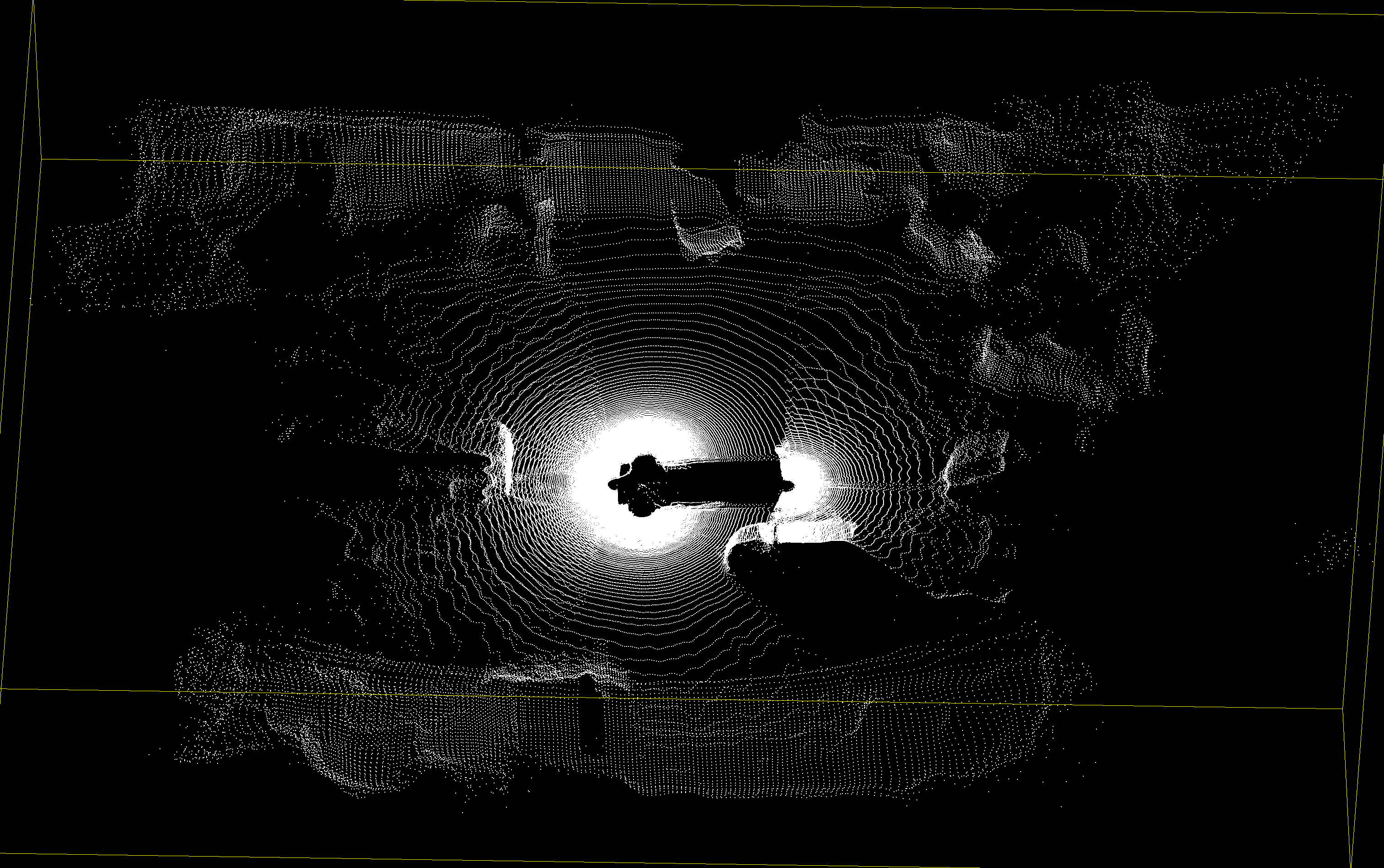}
\end{subfigure}
\begin{subfigure}[b]{0.24\textwidth}
    \centering
    \includegraphics[width=\textwidth,height=2.5cm,keepaspectratio]{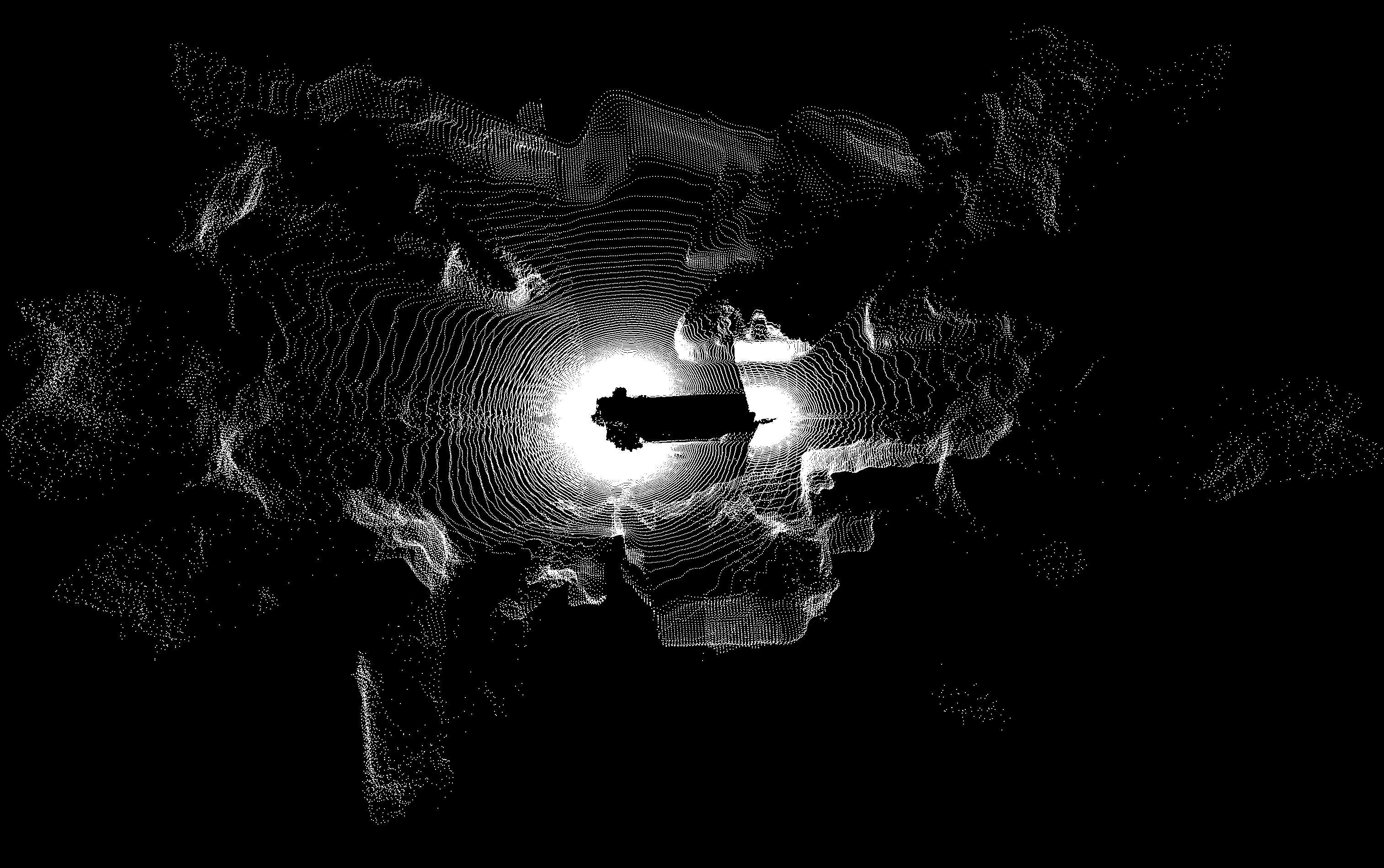}
\end{subfigure}
\begin{subfigure}[b]{0.24\textwidth}
    \centering
    \includegraphics[width=\textwidth,height=2.5cm,keepaspectratio]{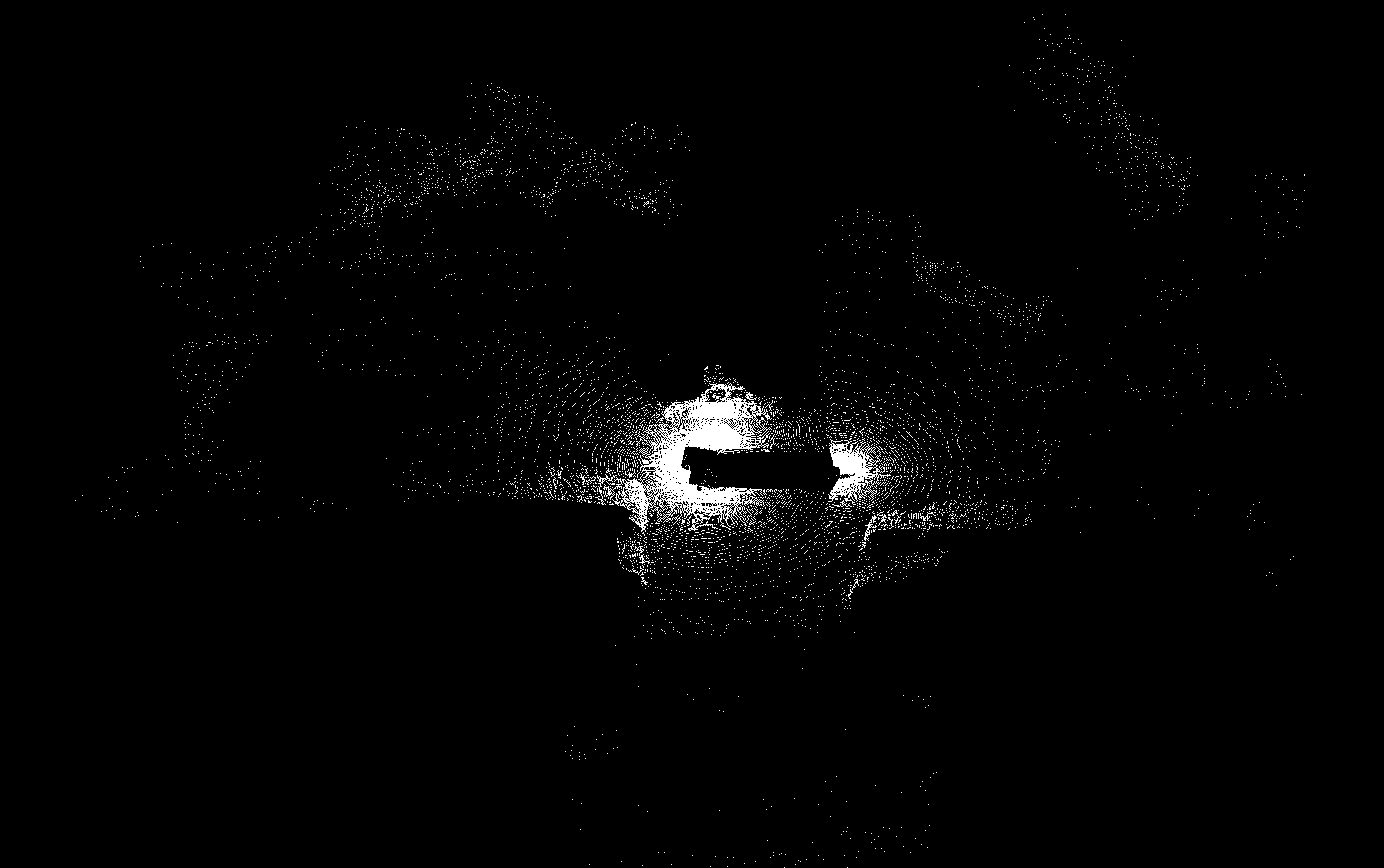}
\end{subfigure}
\begin{subfigure}[b]{0.24\textwidth}
    \centering
    \includegraphics[width=\textwidth,height=2.5cm,keepaspectratio]{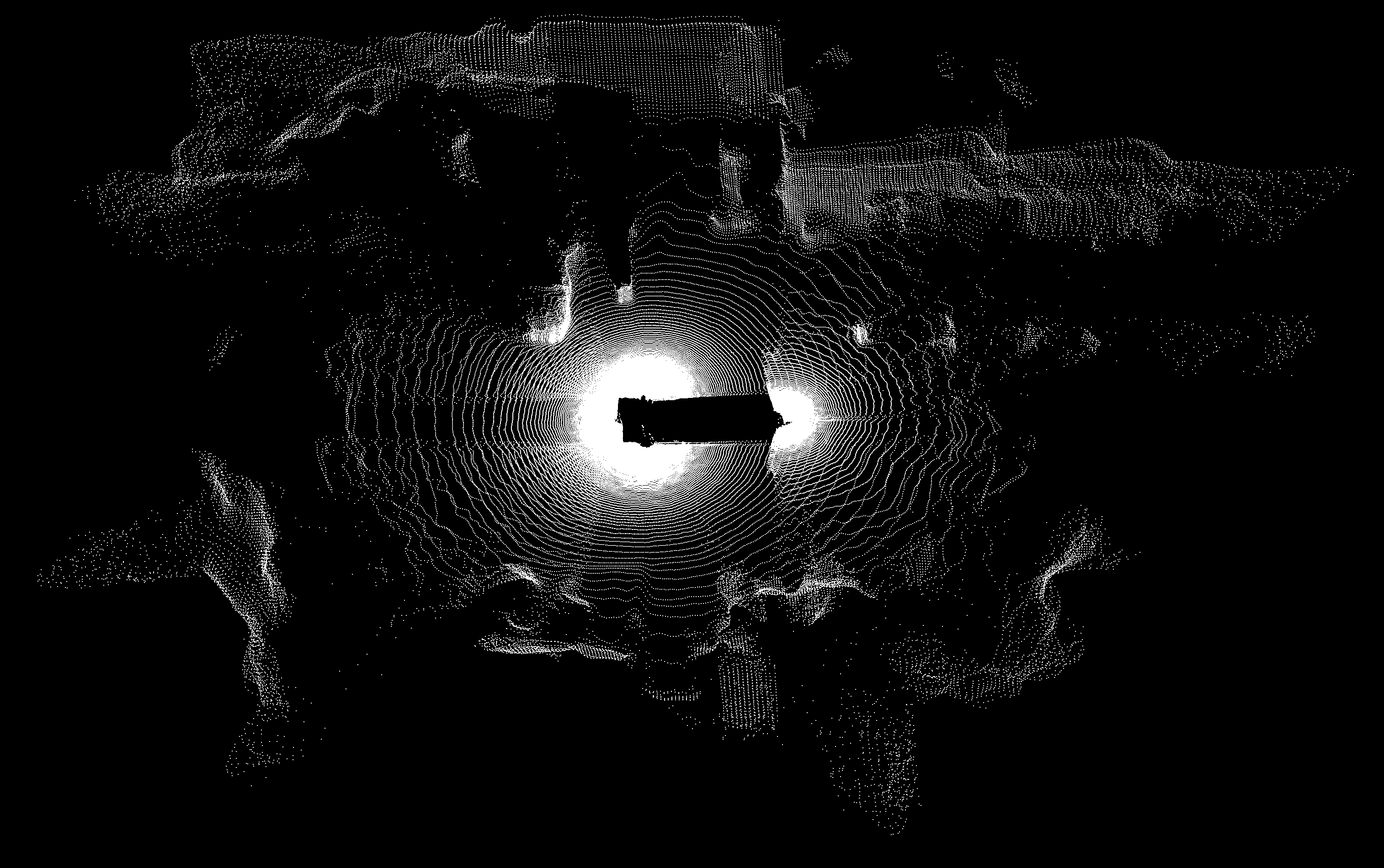}
\end{subfigure}

\begin{subfigure}[b]{0.24\textwidth}
    \centering
    \includegraphics[width=\textwidth,height=2.5cm,keepaspectratio]{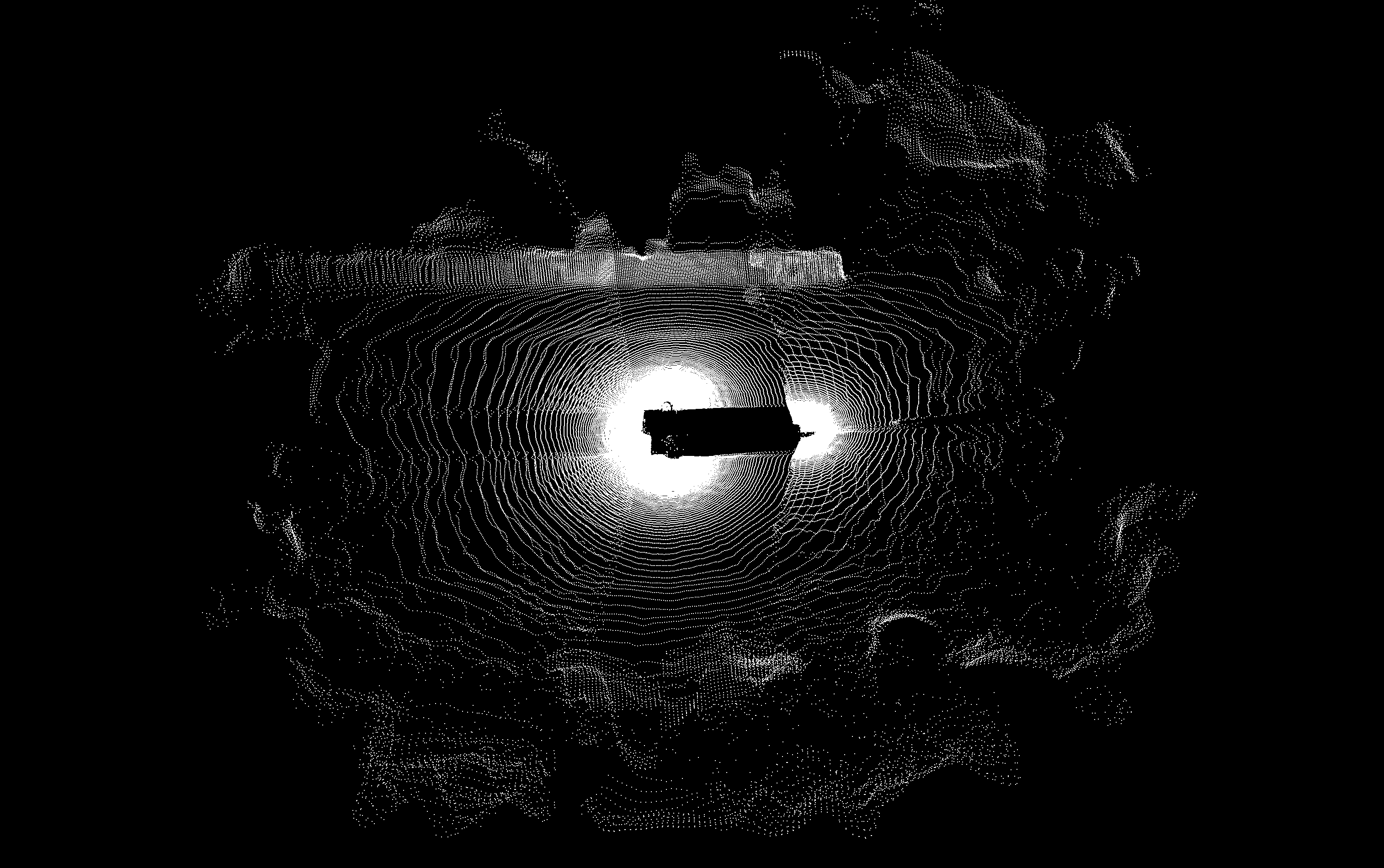}
\end{subfigure}
\begin{subfigure}[b]{0.24\textwidth}
    \centering
    \includegraphics[width=\textwidth,height=2.5cm,keepaspectratio]{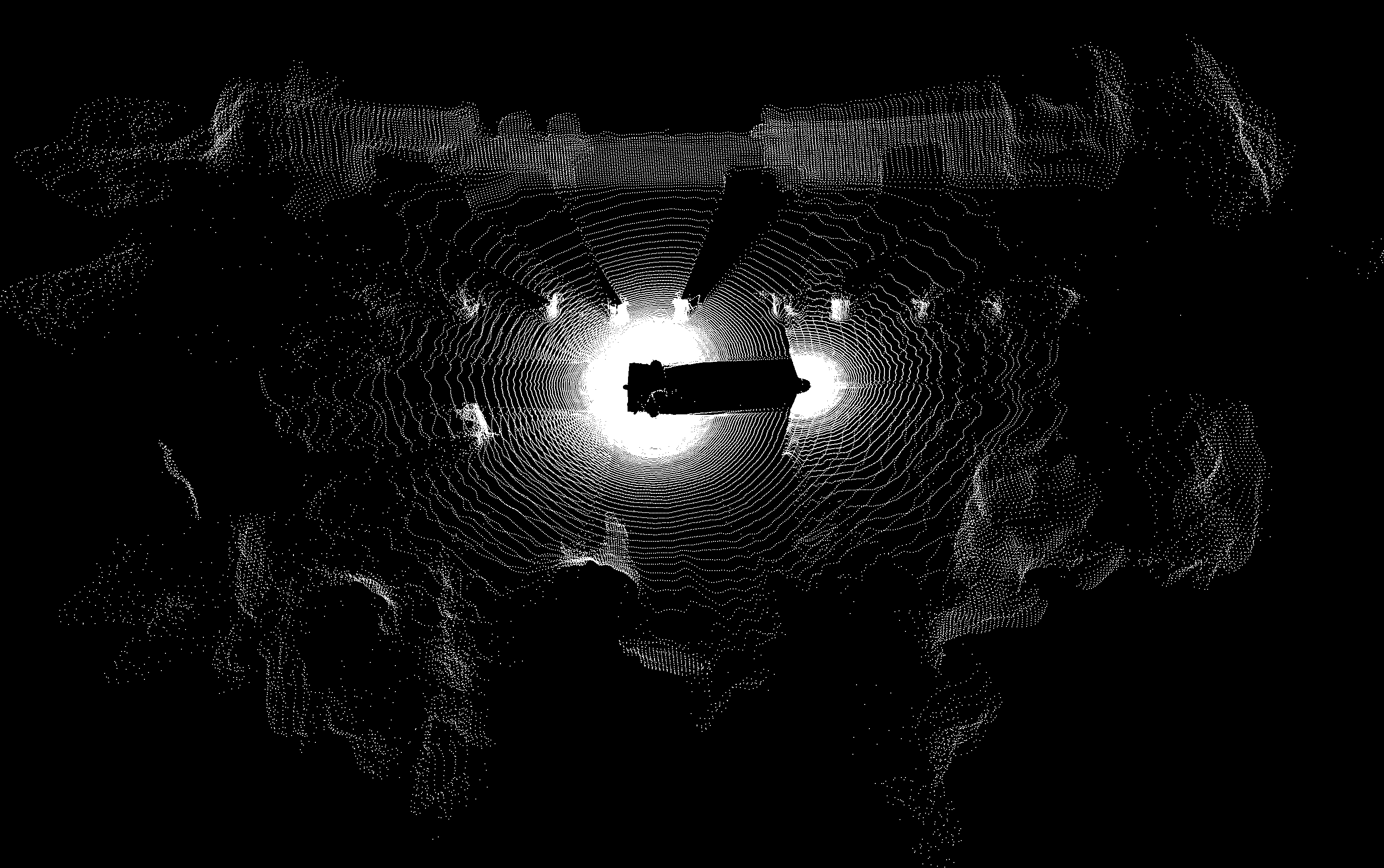}
\end{subfigure}
\begin{subfigure}[b]{0.24\textwidth}
    \centering
    \includegraphics[width=\textwidth,height=2.5cm,keepaspectratio]{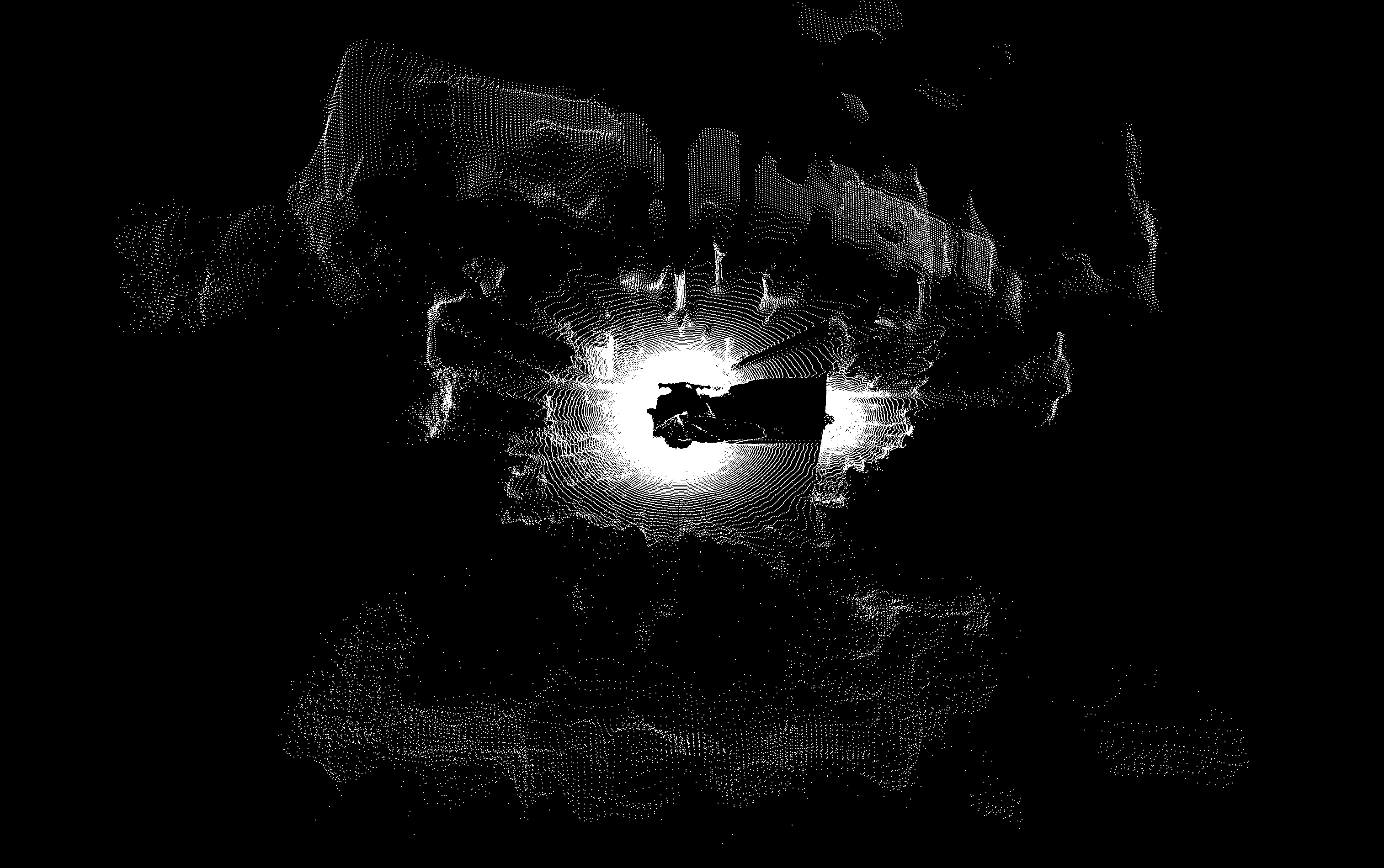}
\end{subfigure}
\begin{subfigure}[b]{0.24\textwidth}
    \centering
    \includegraphics[width=\textwidth,height=2.5cm,keepaspectratio]{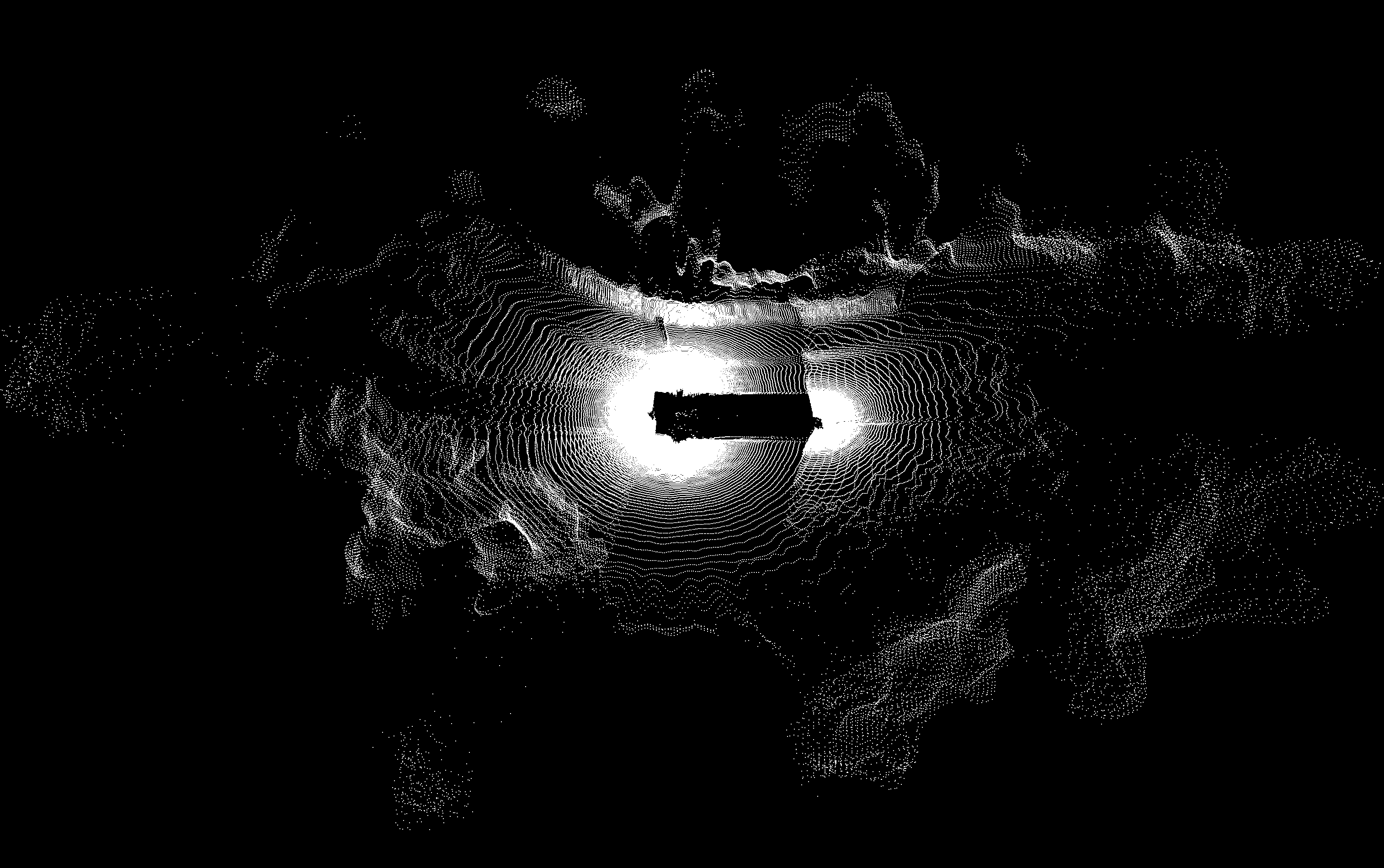}
\end{subfigure}
\centering
{\small \textbf{\lname{}}}

\caption{\textbf{Cross-Sensor Alignment on Waymo.} Qualitative comparison between our method (the last two rows) and competing approaches (the first row). While competitors generate each sensor stream independently, leading to misalignment and geometric inconsistencies across viewpoints, our {\bf Inter-Sensor Module} enables coherent multi-view generation with consistent object geometry and spatial alignment across all LiDAR sensors.}\label{fig:waymo_cross_sensor_alignment}
\end{figure*}

\noindent \textbf{Training configuration}.  
We follow the public OpenDWM LiDAR configurations as closely as possible.  
The released VAE configuration uses AdamW with learning rate $10^{-4}$ and betas $(0.9, 0.95)$.  
The temporal DiT configuration uses AdamW with learning rate $5\times10^{-5}$, betas $(0.9, 0.99)$, and weight decay $10^{-4}$, together with cosine annealing ($T_{\max}=100000$, $\eta_{\min}=10^{-5}$).  
The official temporal configuration enables gradient checkpointing, uses batch size 1, a FlowMatch Euler scheduler with 1000 training timesteps, and conditions on three reference frames.  
We note that the released OpenDWM LiDAR models are primarily configured for \texttt{LIDAR\_TOP} data and layout-conditioned generation.  
As a result, the Waymo SIDE LiDAR setting requires non-trivial adaptation and is not a native training configuration in the original release.

\section{Waymo Qualitative Results}

In this section, we present qualitative results on the Waymo dataset. We highlight the impact of our architectural modules on the generation quality, demonstrating improvements in cross-sensor alignment, temporal consistency, and controllable scene synthesis.

\paragraph{Inter-sensor Alignment.}
The {\bf Inter-sensor Module} provides clear benefits by enabling joint modeling of all LiDAR sensors within the same temporal frame. Instead of processing each sensor independently, the module allows the network to learn direct interactions across viewpoints through attention.
As shown in Figure~\ref{fig:waymo_cross_sensor_alignment}, the module enables coherent multi-view integration, leading to improved alignment and more accurate scene understanding compared to the independent sensor generation of the competitors.

\begin{figure*}[t]
\centering
\setlength{\tabcolsep}{2pt}
\renewcommand{\arraystretch}{0.85}


\newlength{\tcwidth}
\newlength{\tcheight}
\setlength{\tcwidth}{0.2\textwidth}
\setlength{\tcheight}{0.115\textheight}

\newcommand{\tcimg}[1]{%
    \includegraphics[width=\tcwidth,height=\tcheight]{#1}%
}

{\small \textbf{\lname{}}}

\vspace{0.3em}

\begin{tabular}{cccc}
\textbf{BEV/3D Box} & \textbf{$t=1$} & \textbf{$t=3$} & \textbf{$t=5$} \\

\tcimg{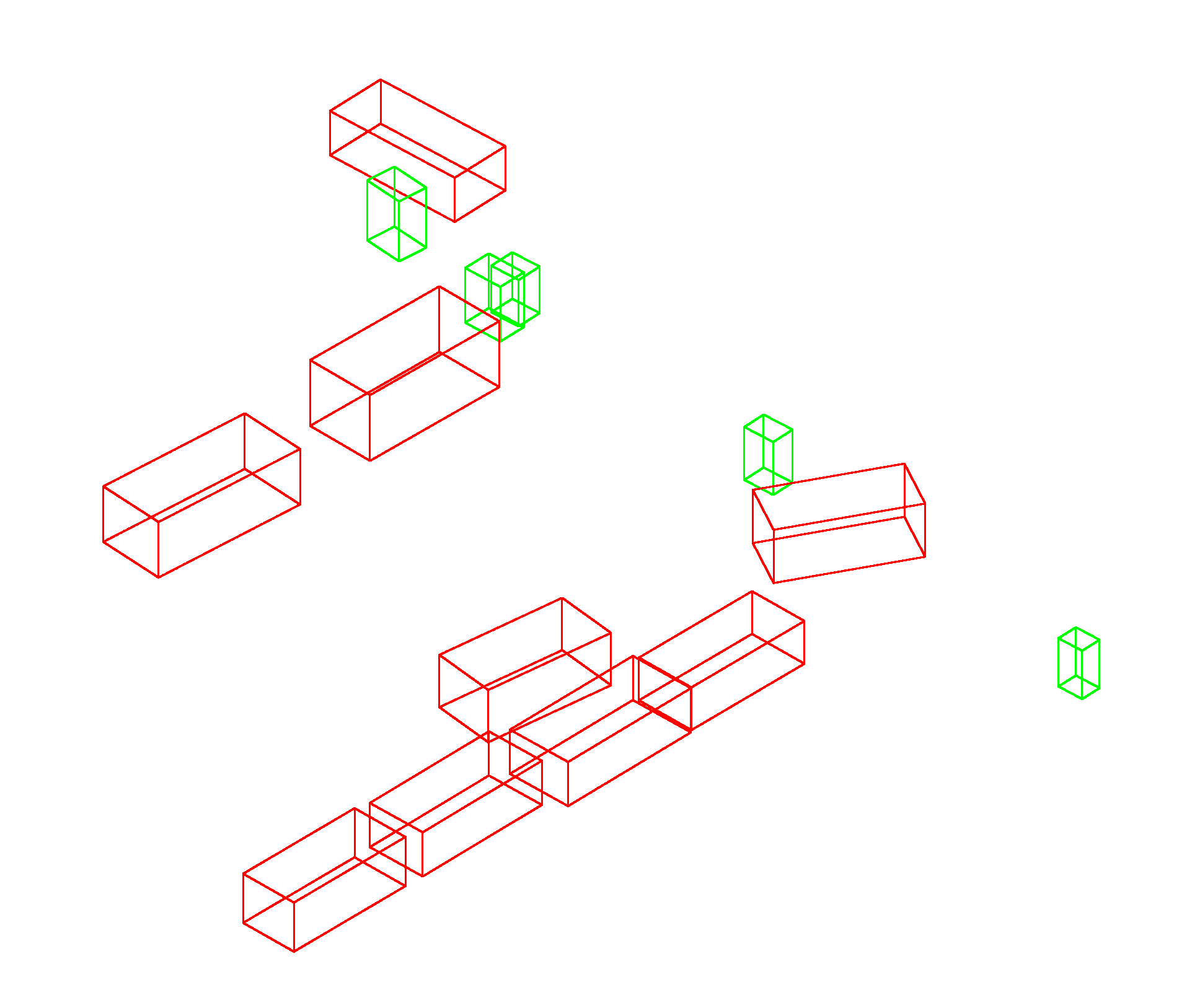} &
\tcimg{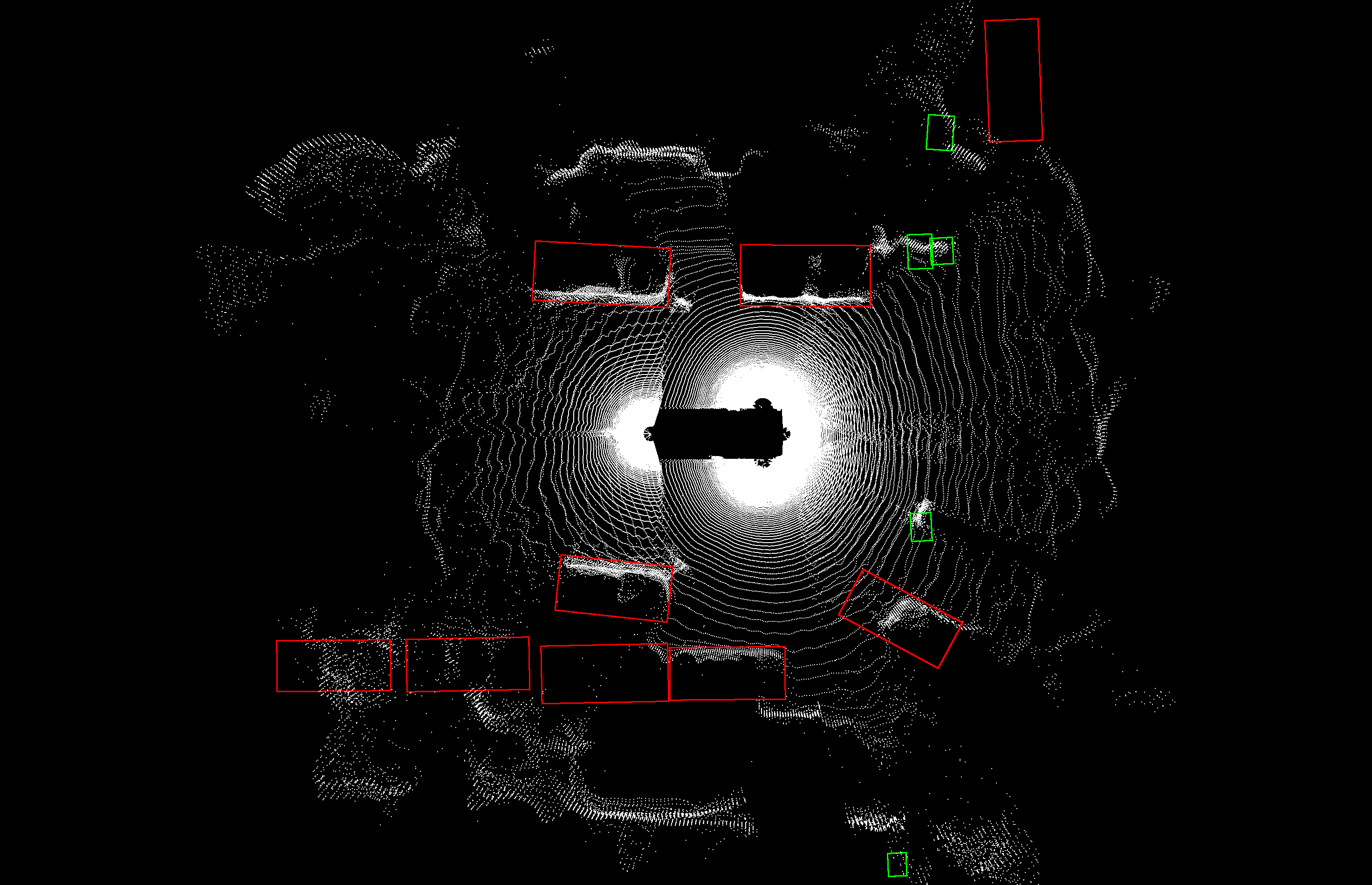} &
\tcimg{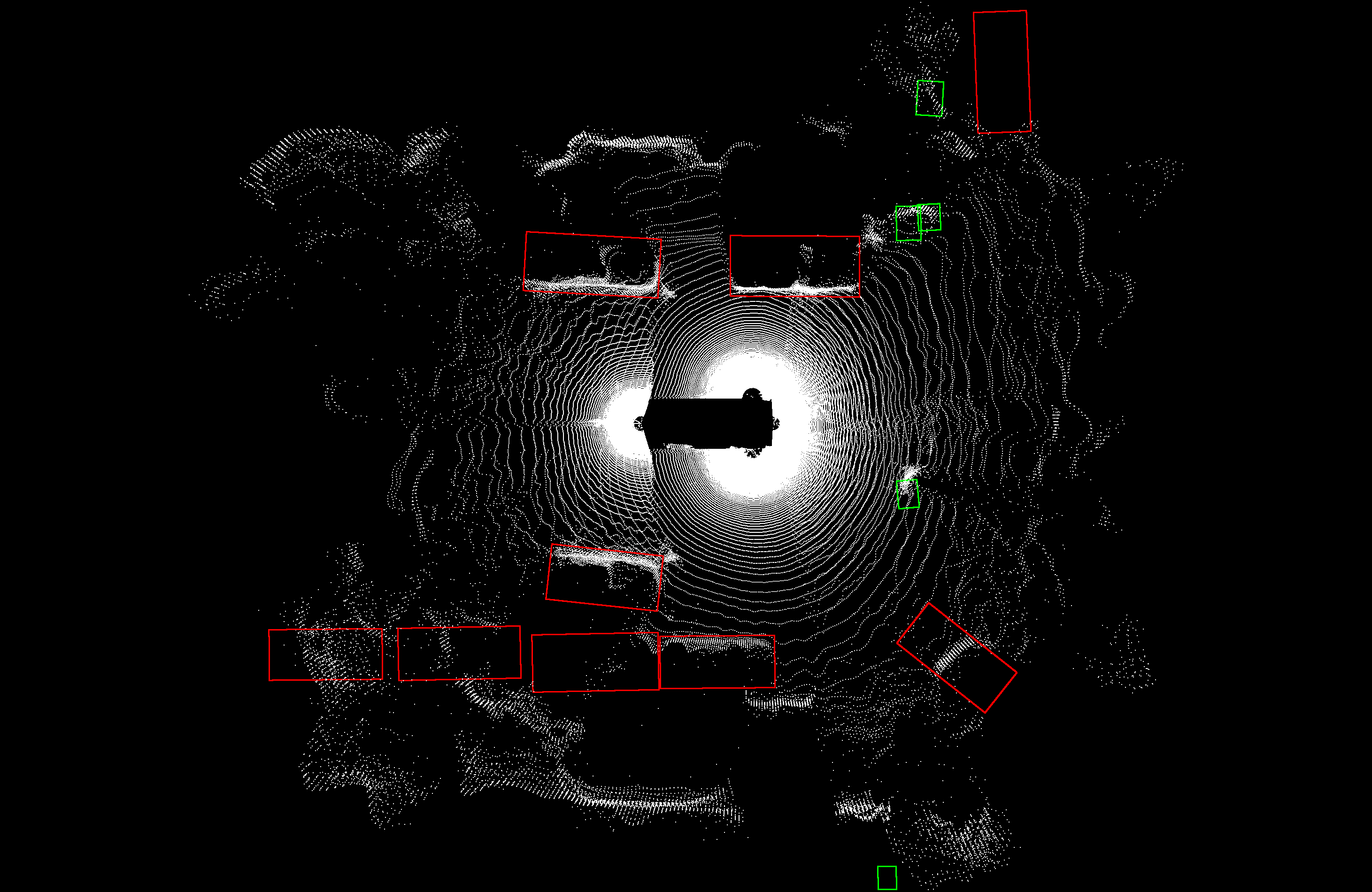} &
\tcimg{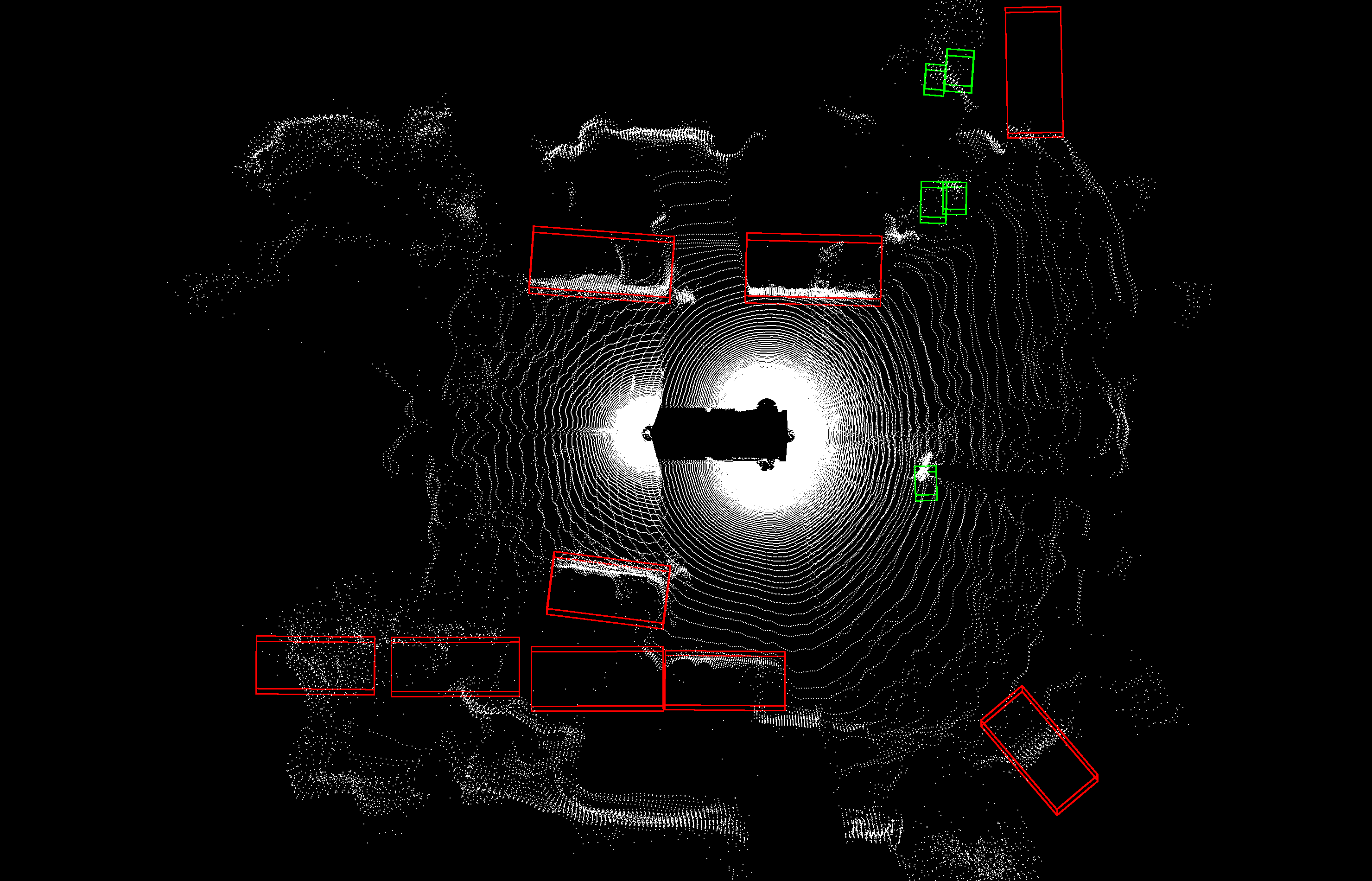} \\[-0.2em]

\tcimg{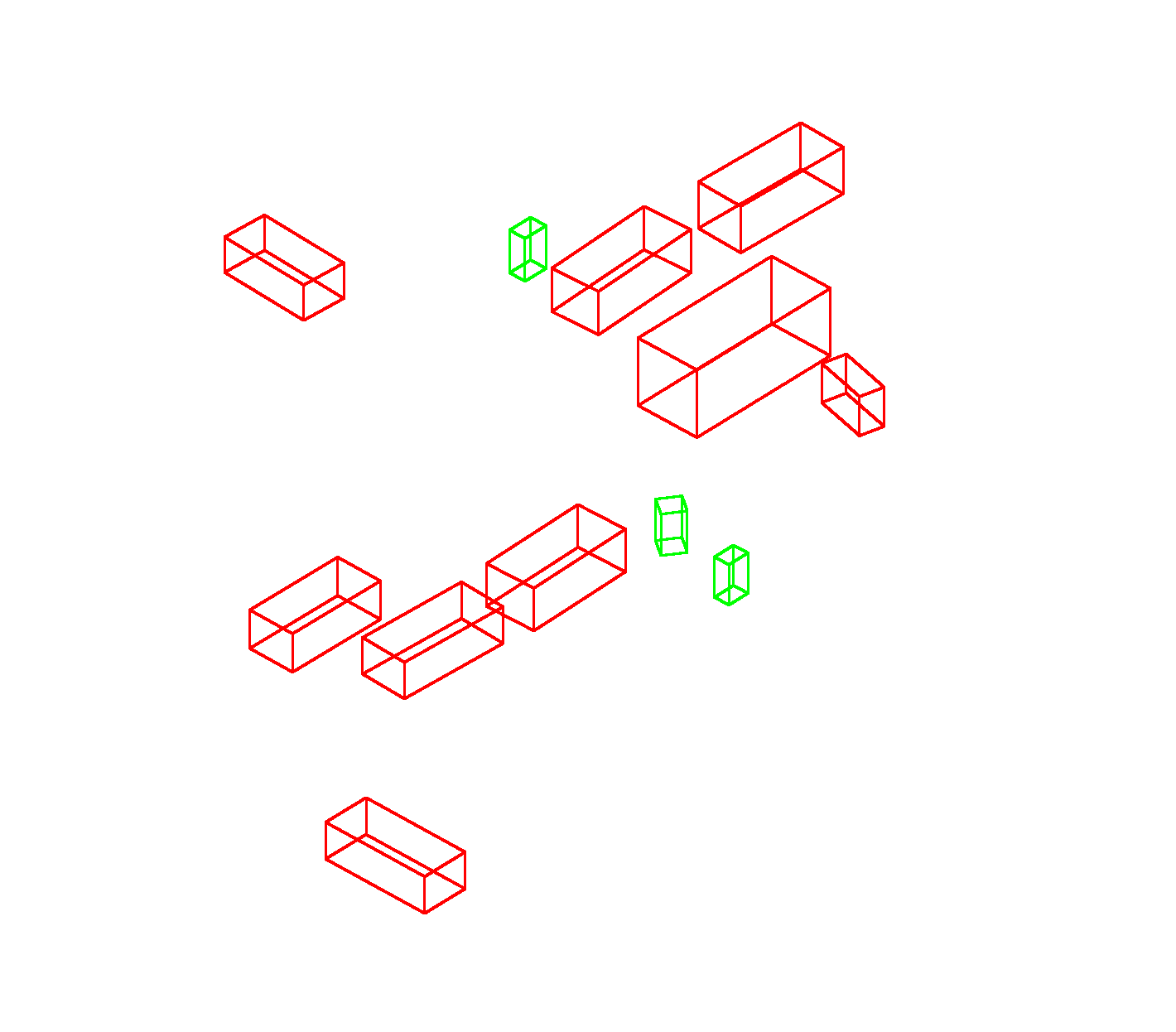} &
\tcimg{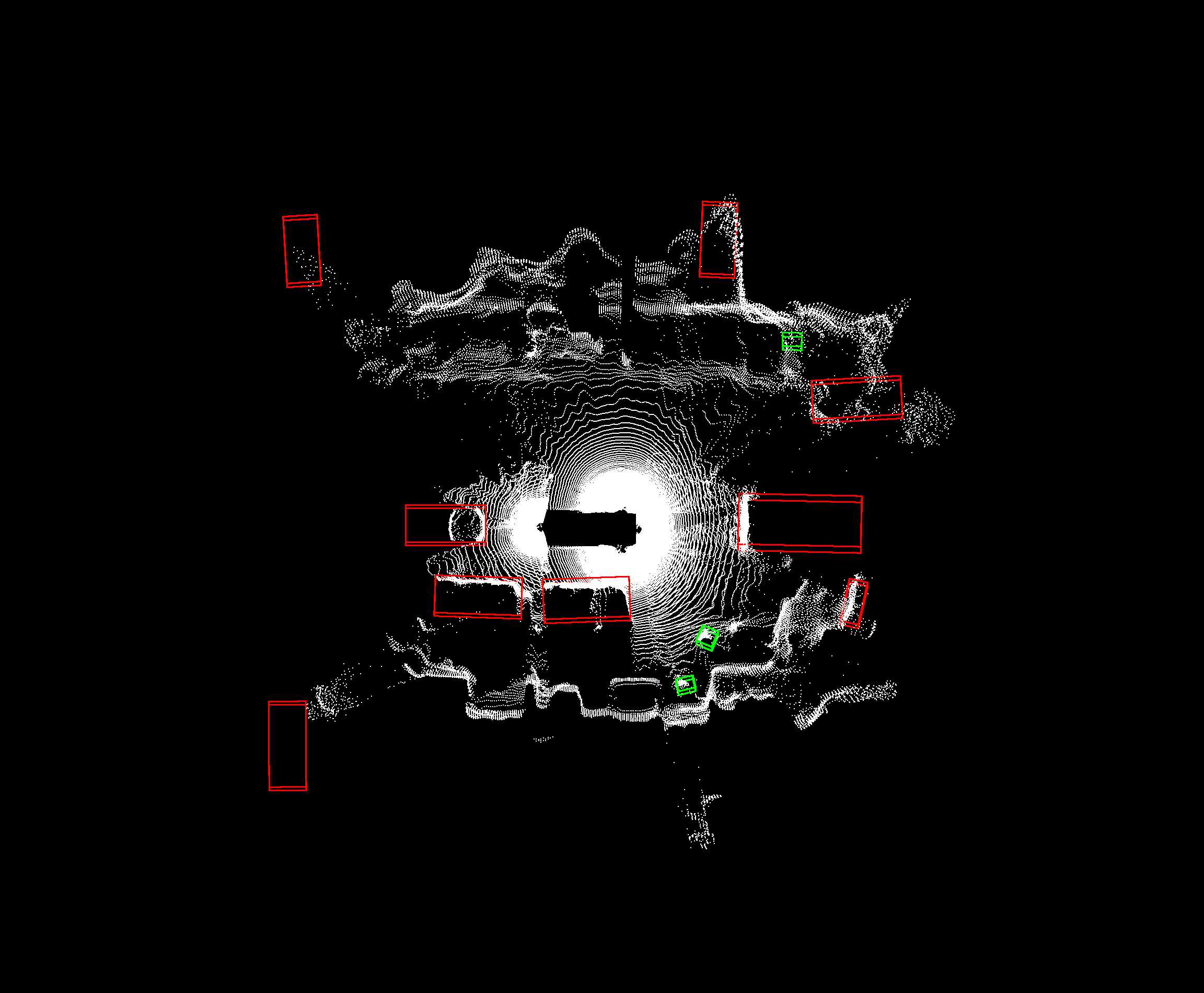} &
\tcimg{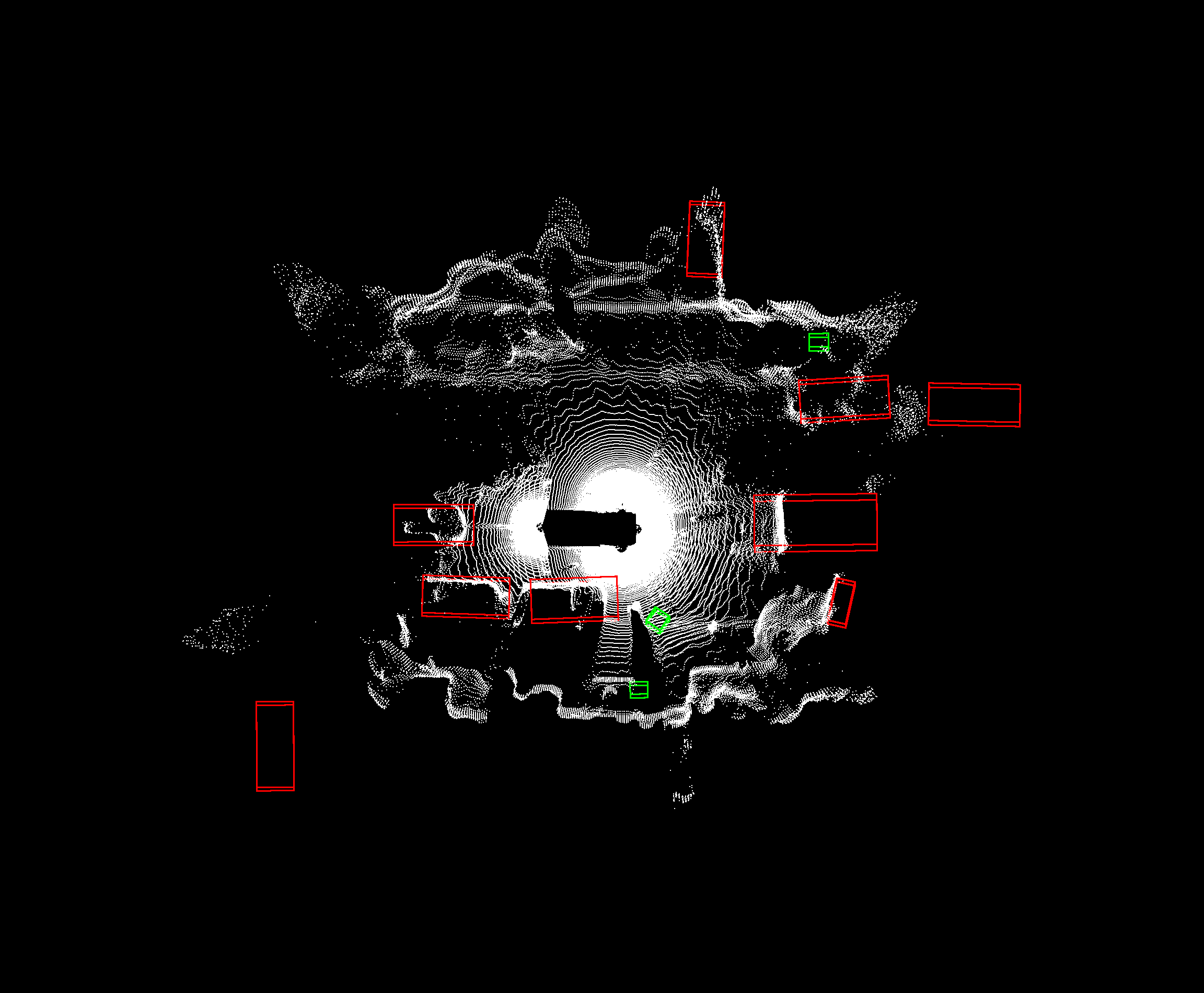} &
\tcimg{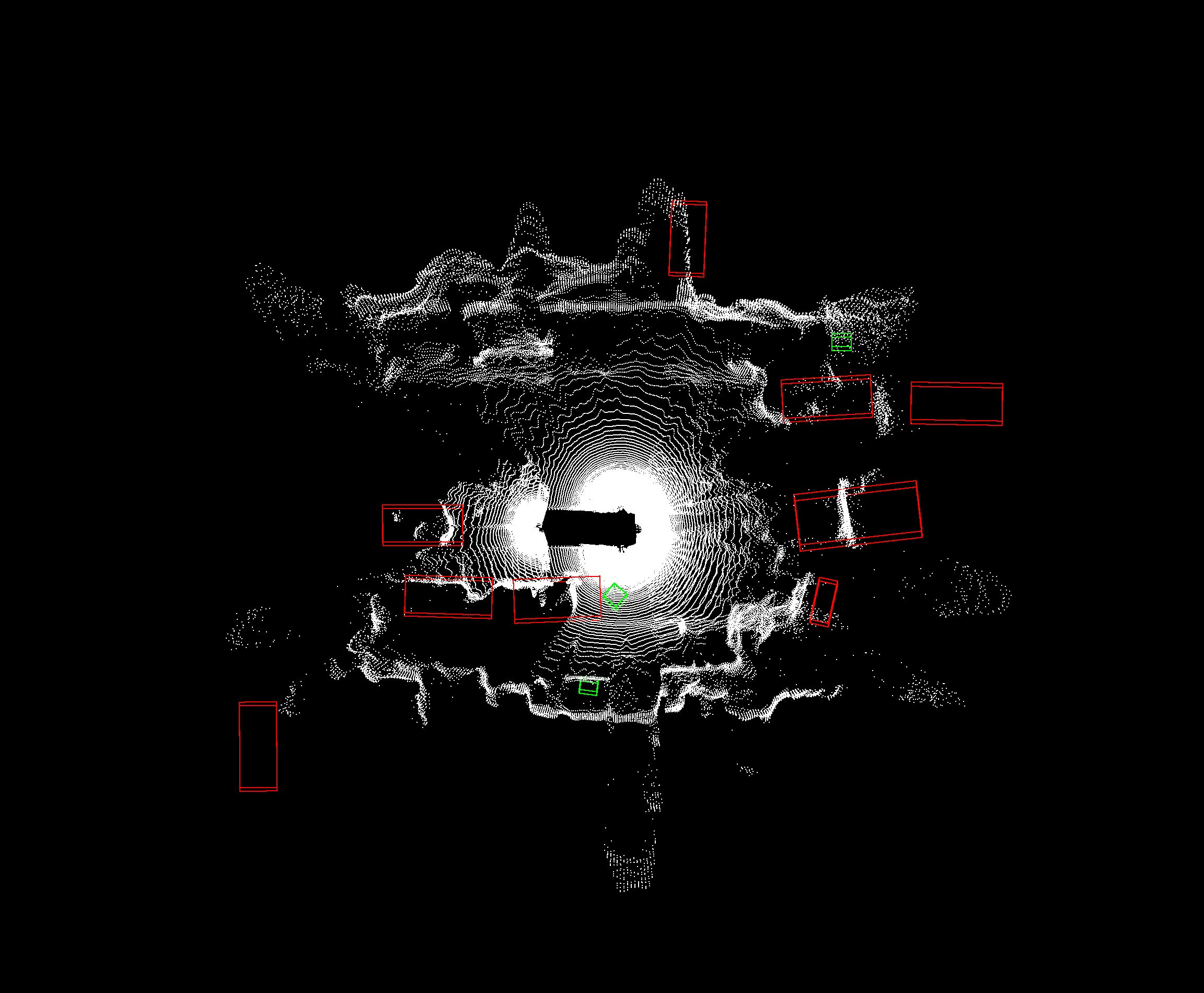} \\[-0.2em]

\tcimg{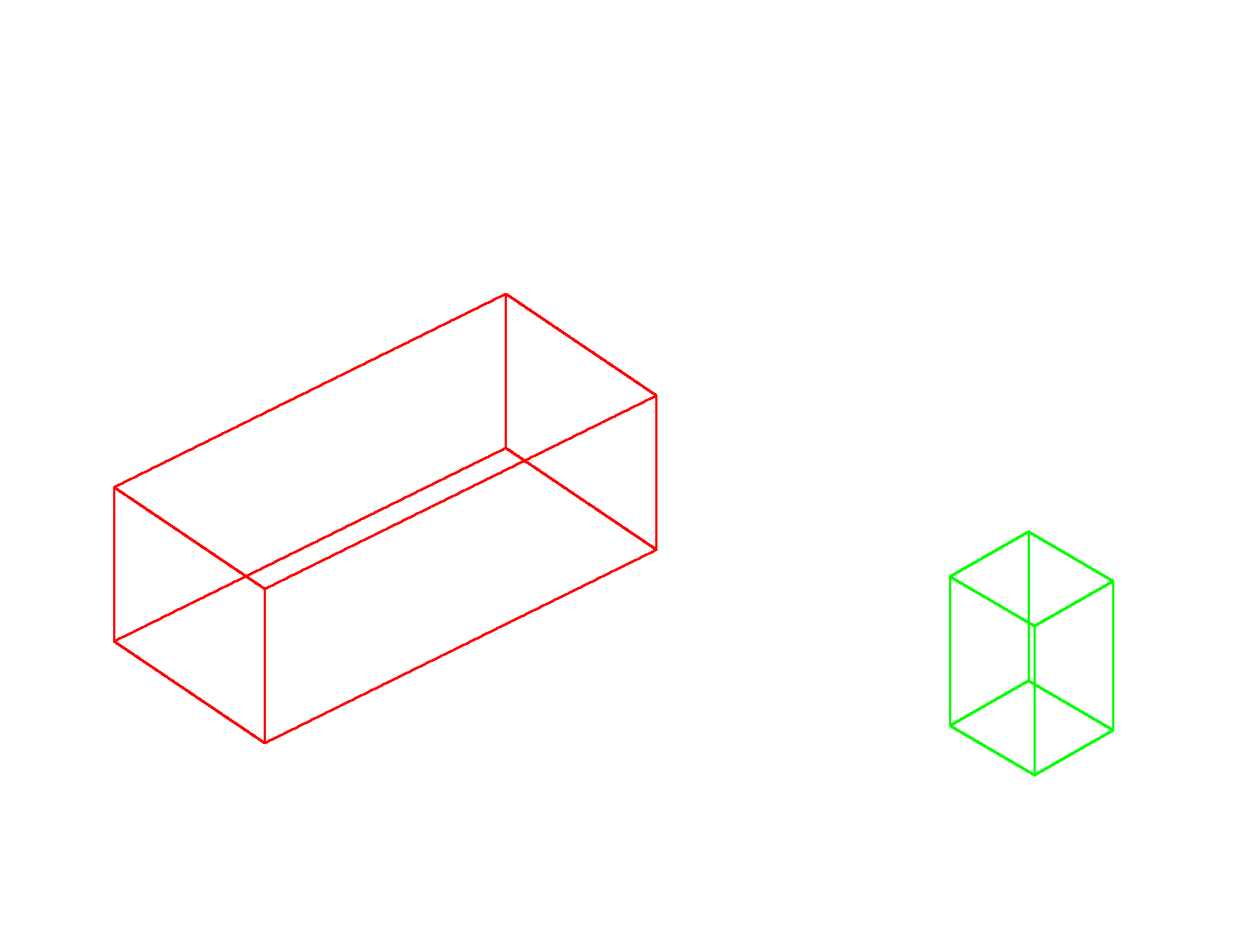} &
\tcimg{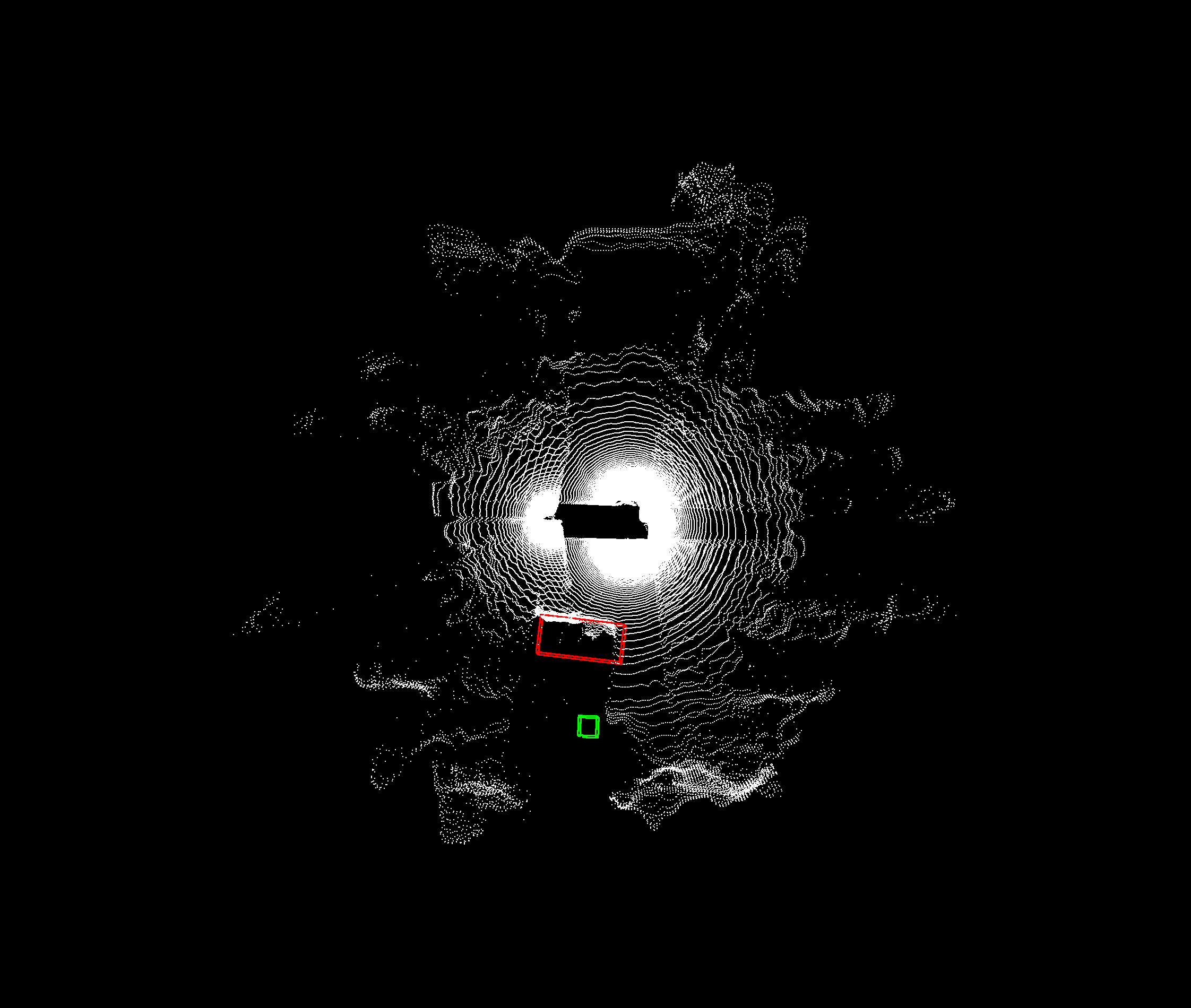} &
\tcimg{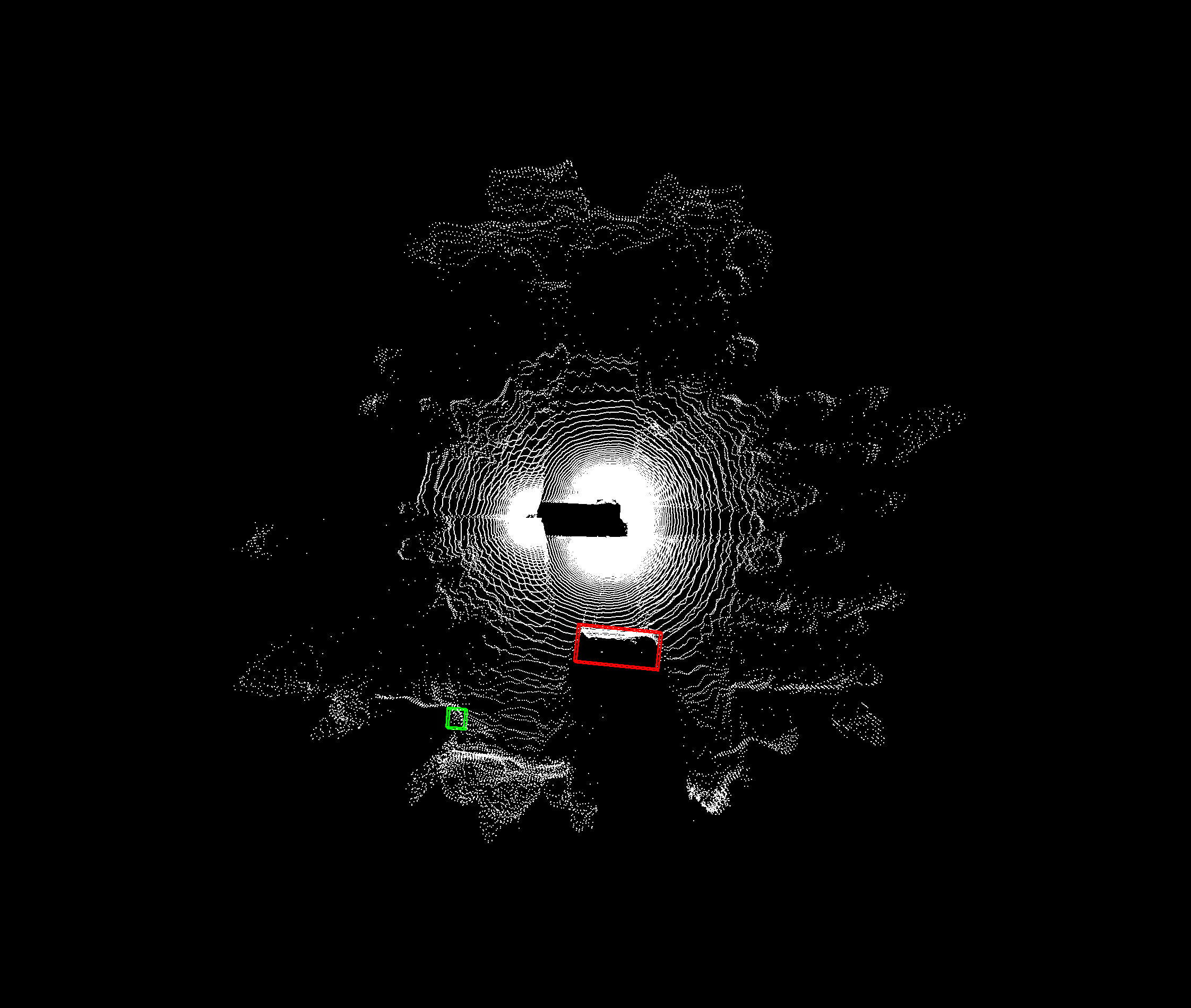} &
\tcimg{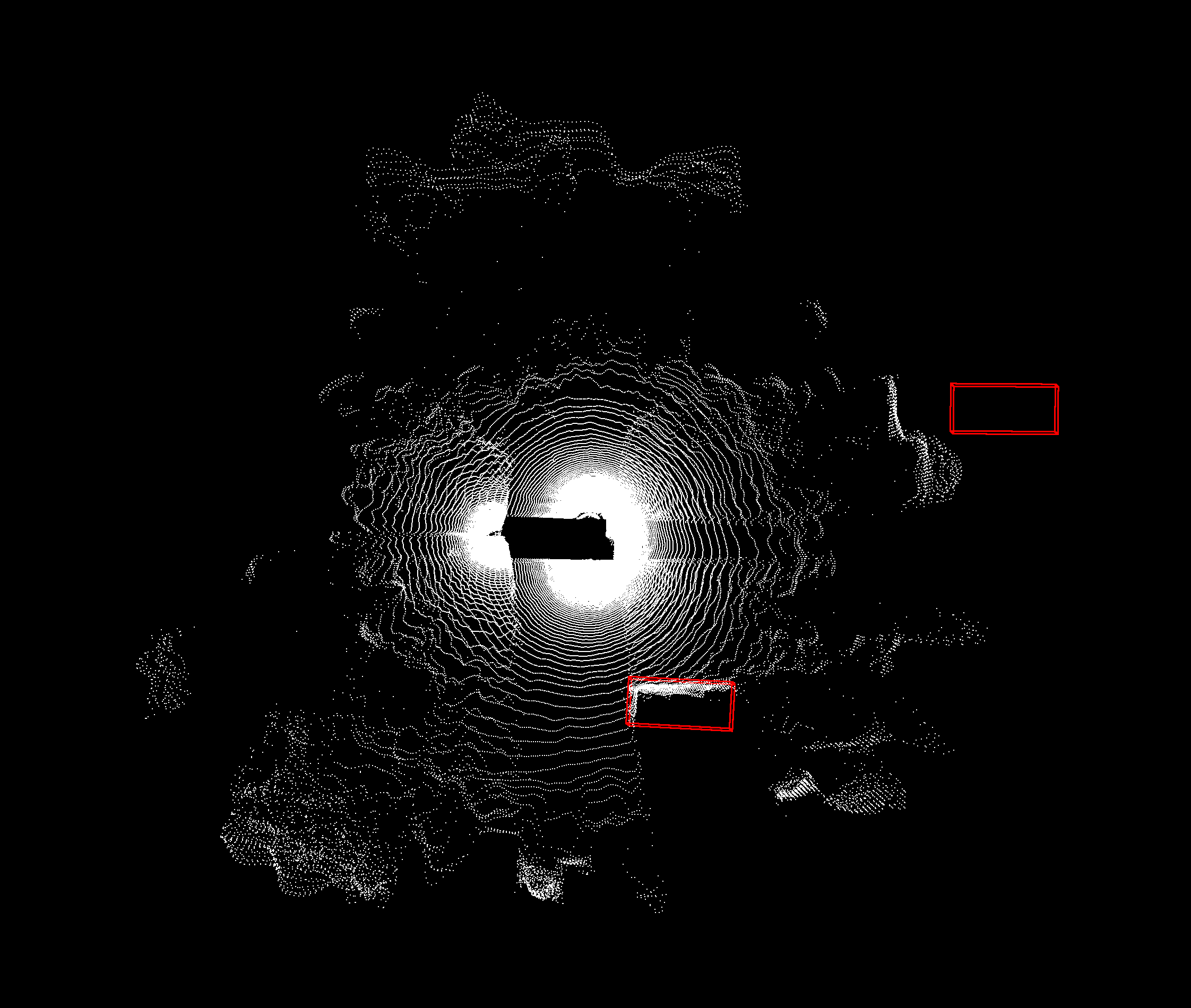} \\[-0.2em]

\tcimg{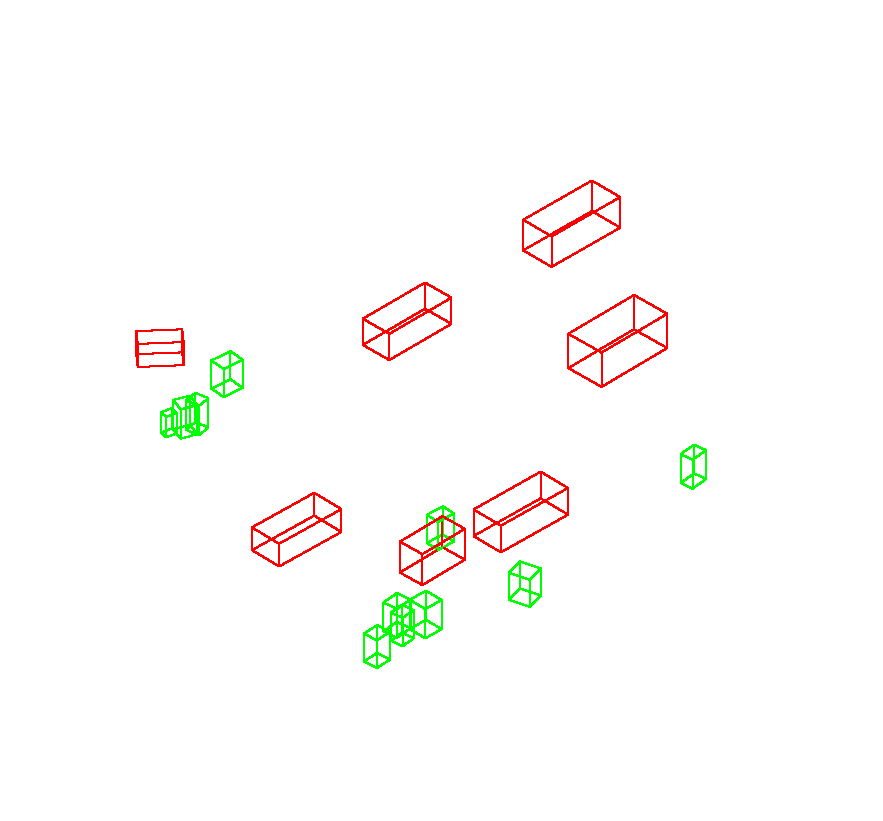} &
\tcimg{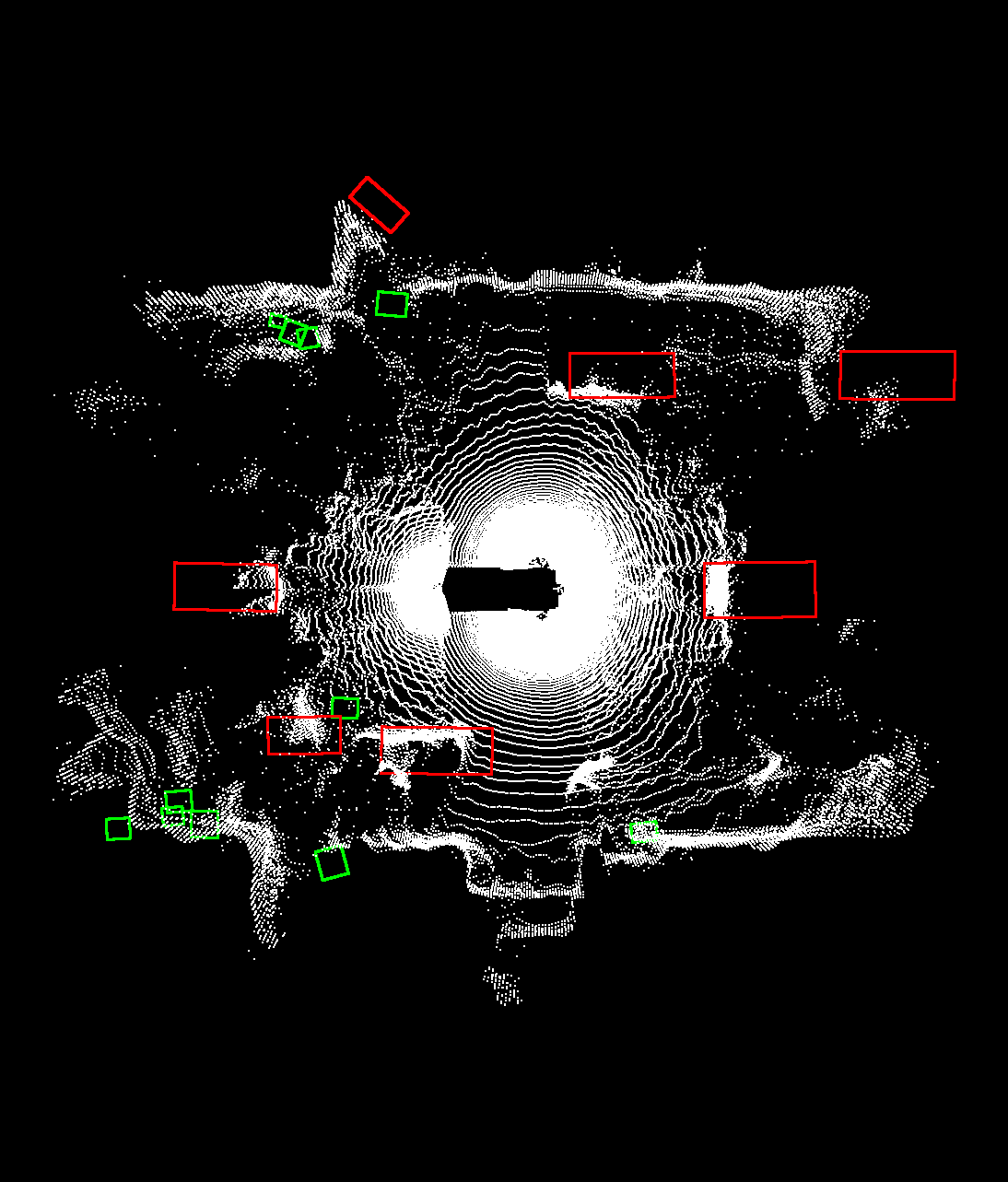} &
\tcimg{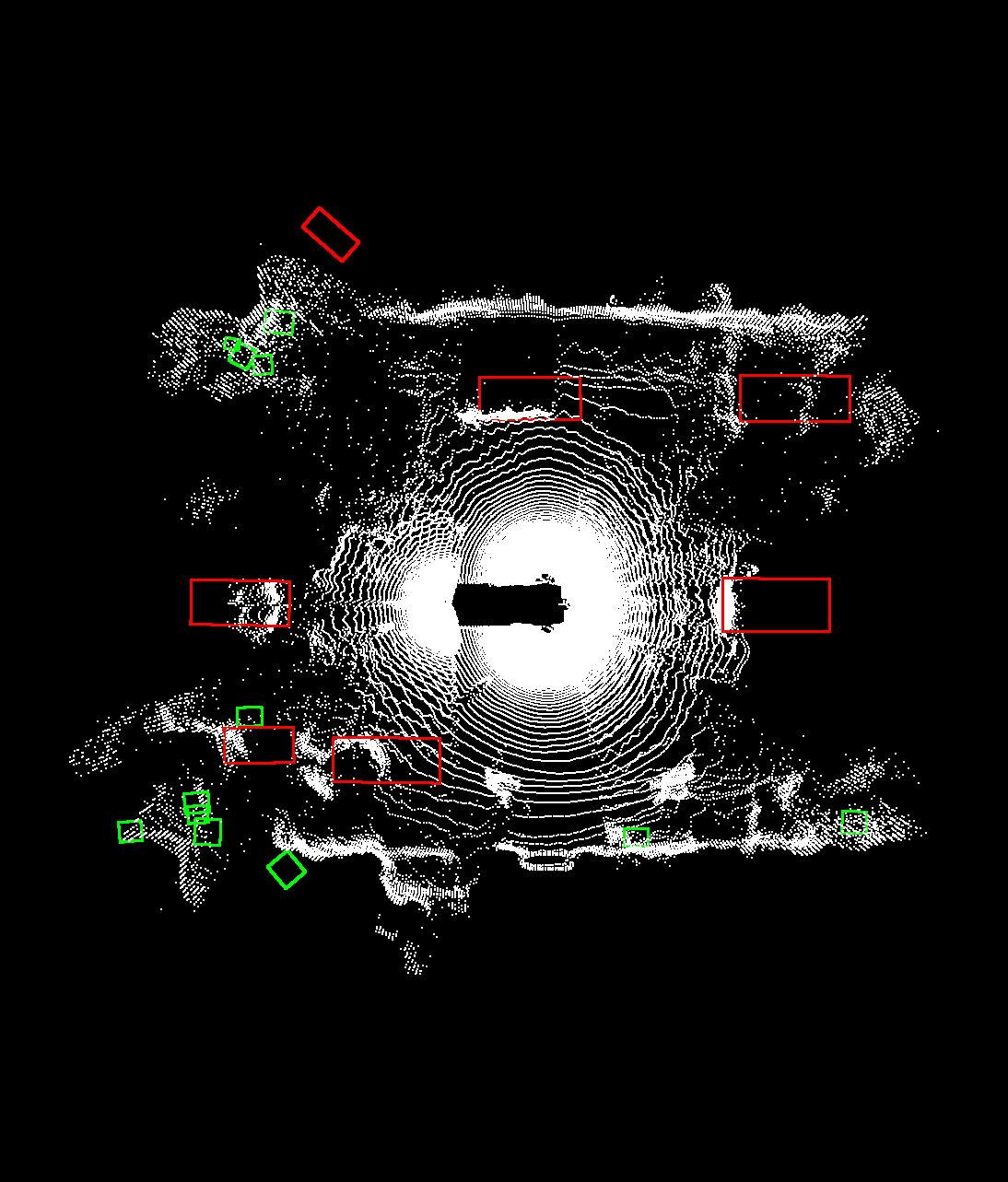} &
\tcimg{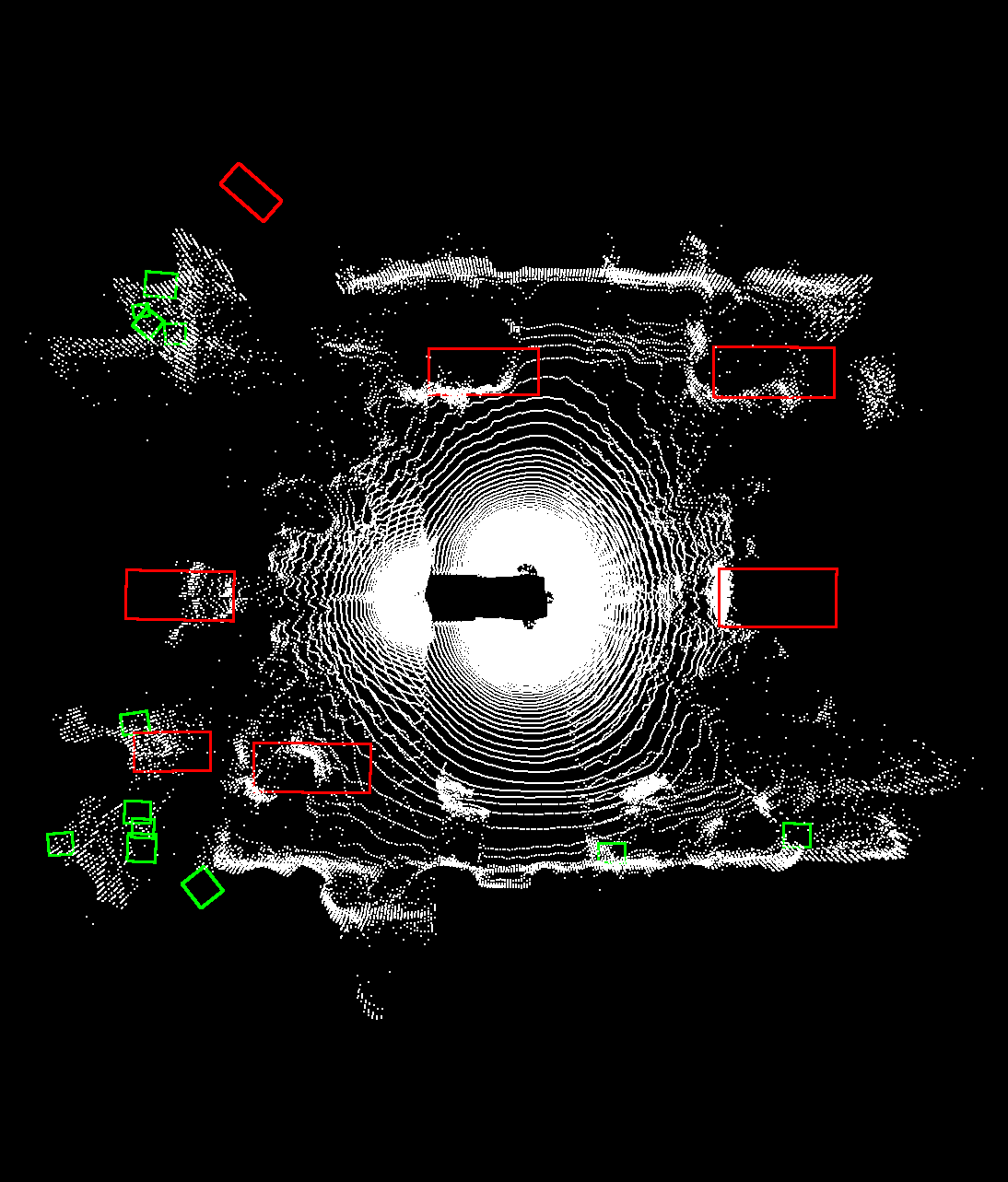} \\[-0.2em]

\tcimg{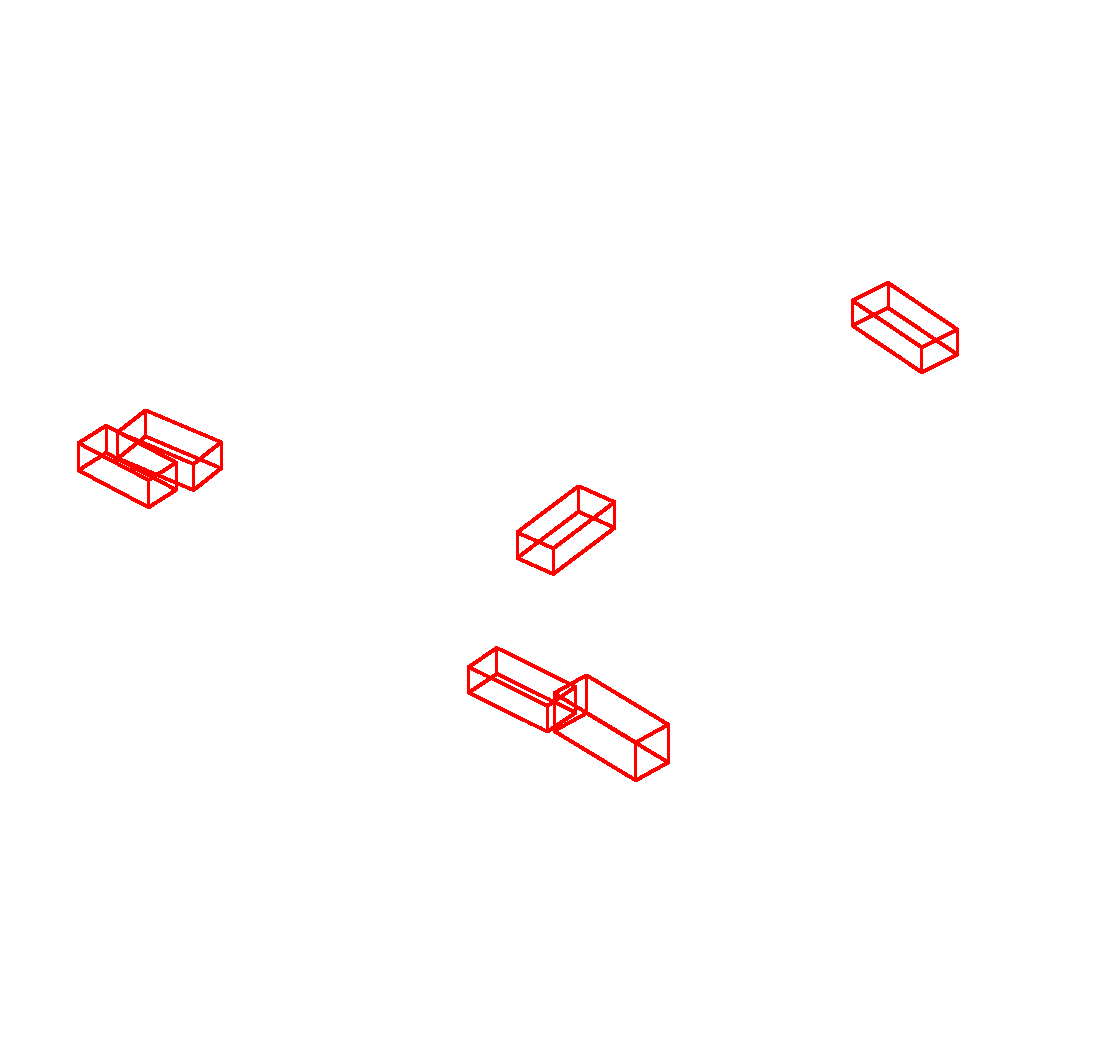} &
\tcimg{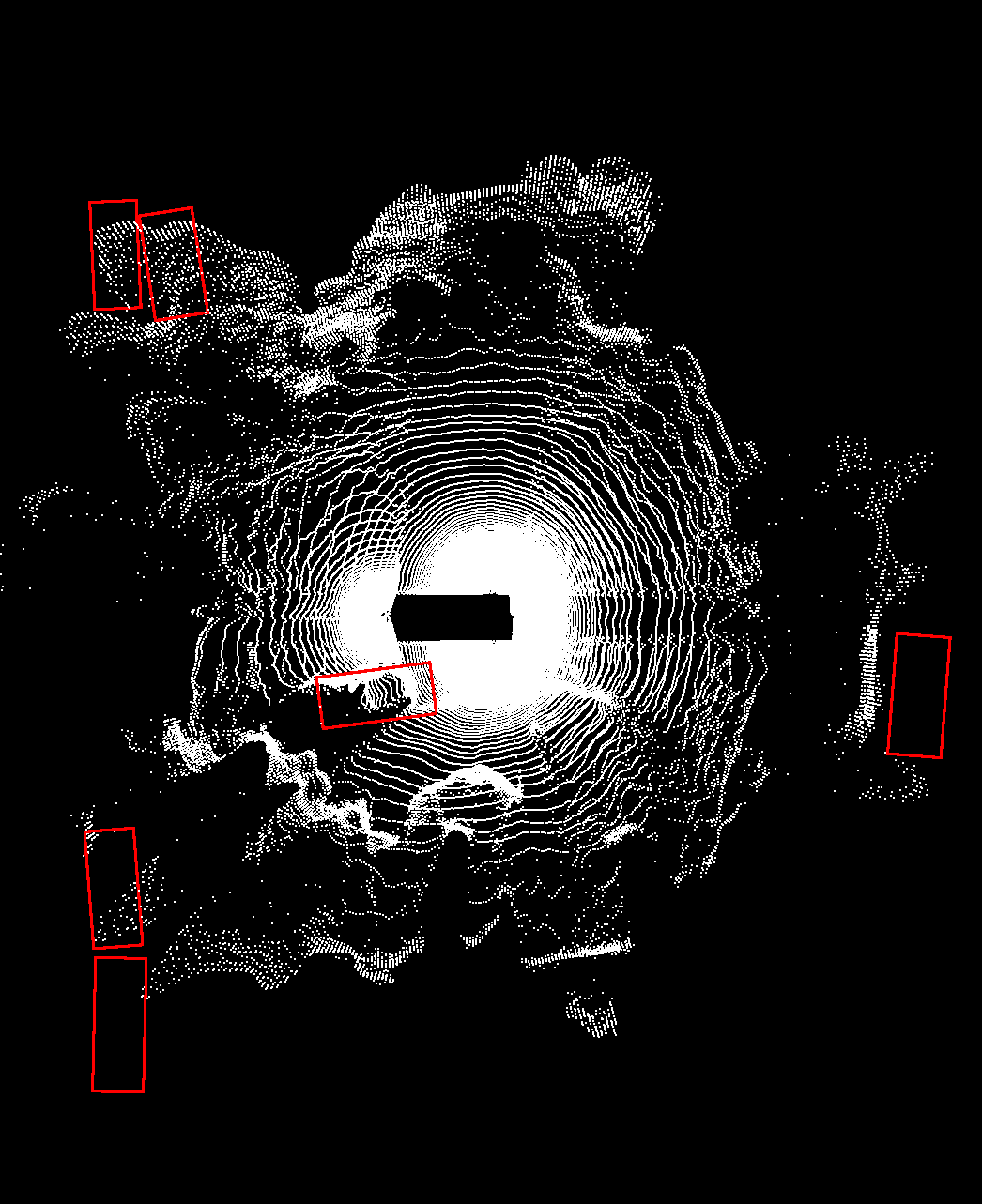} &
\tcimg{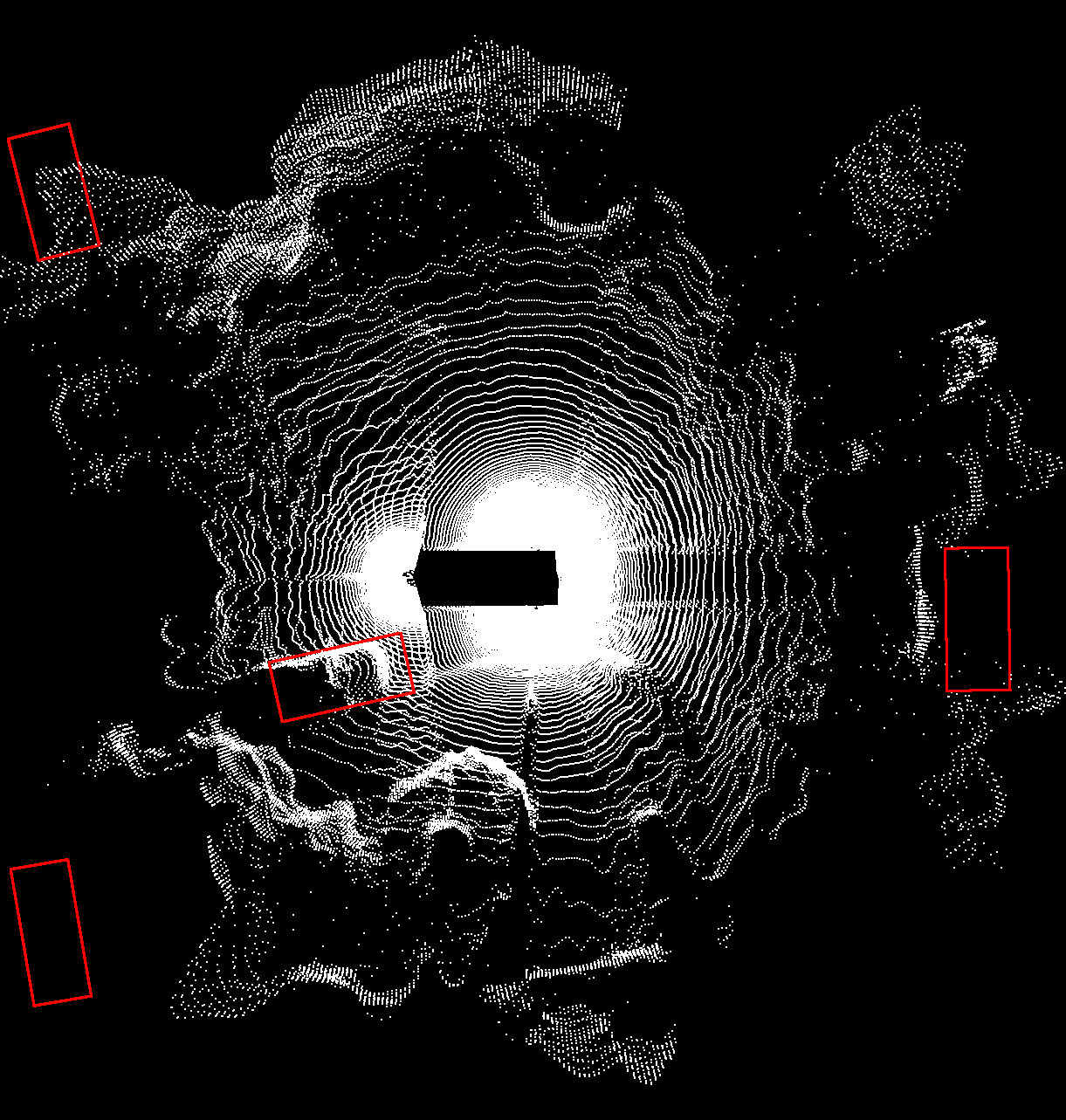} &
\tcimg{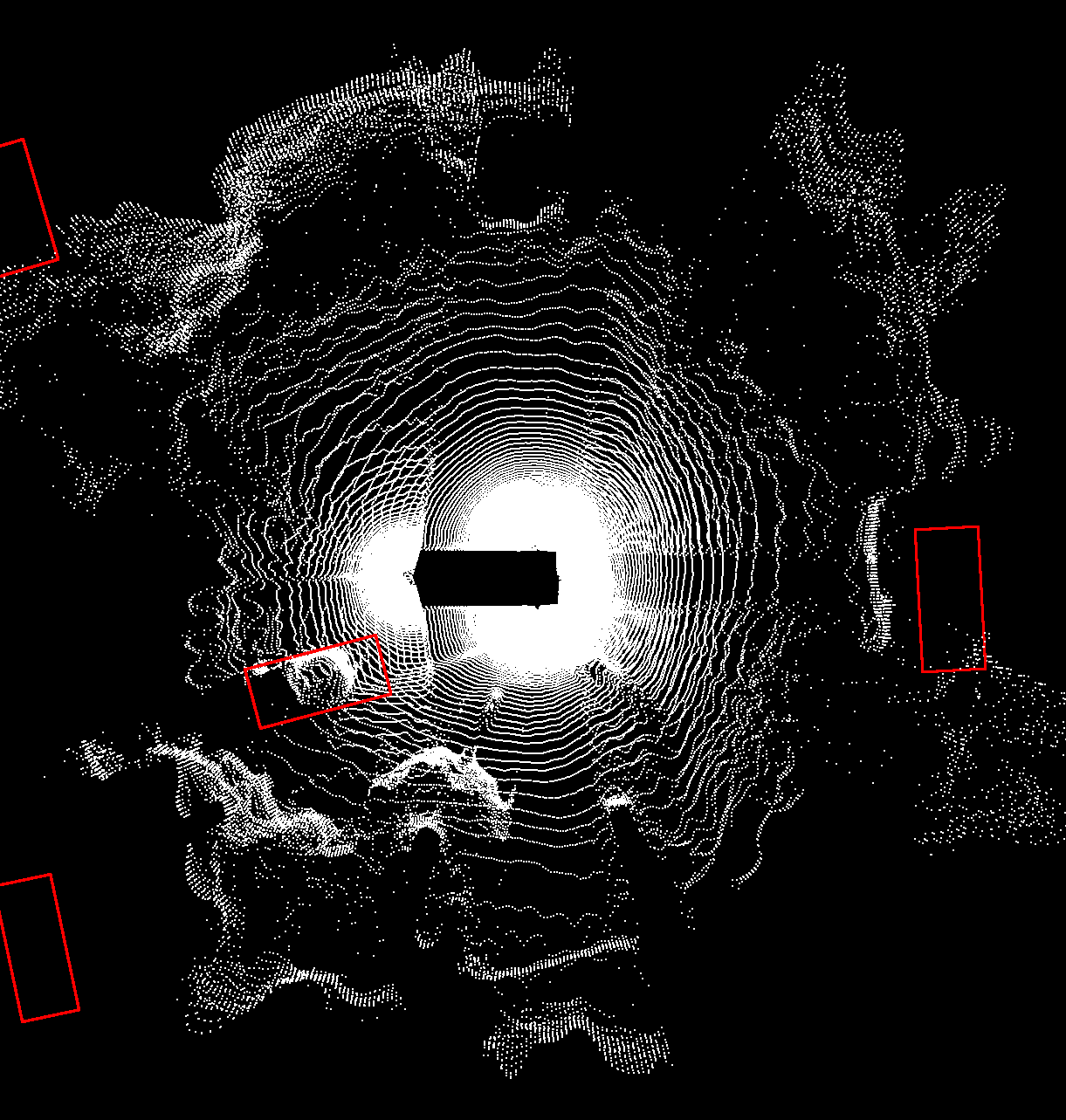} \\[-0.2em]

\tcimg{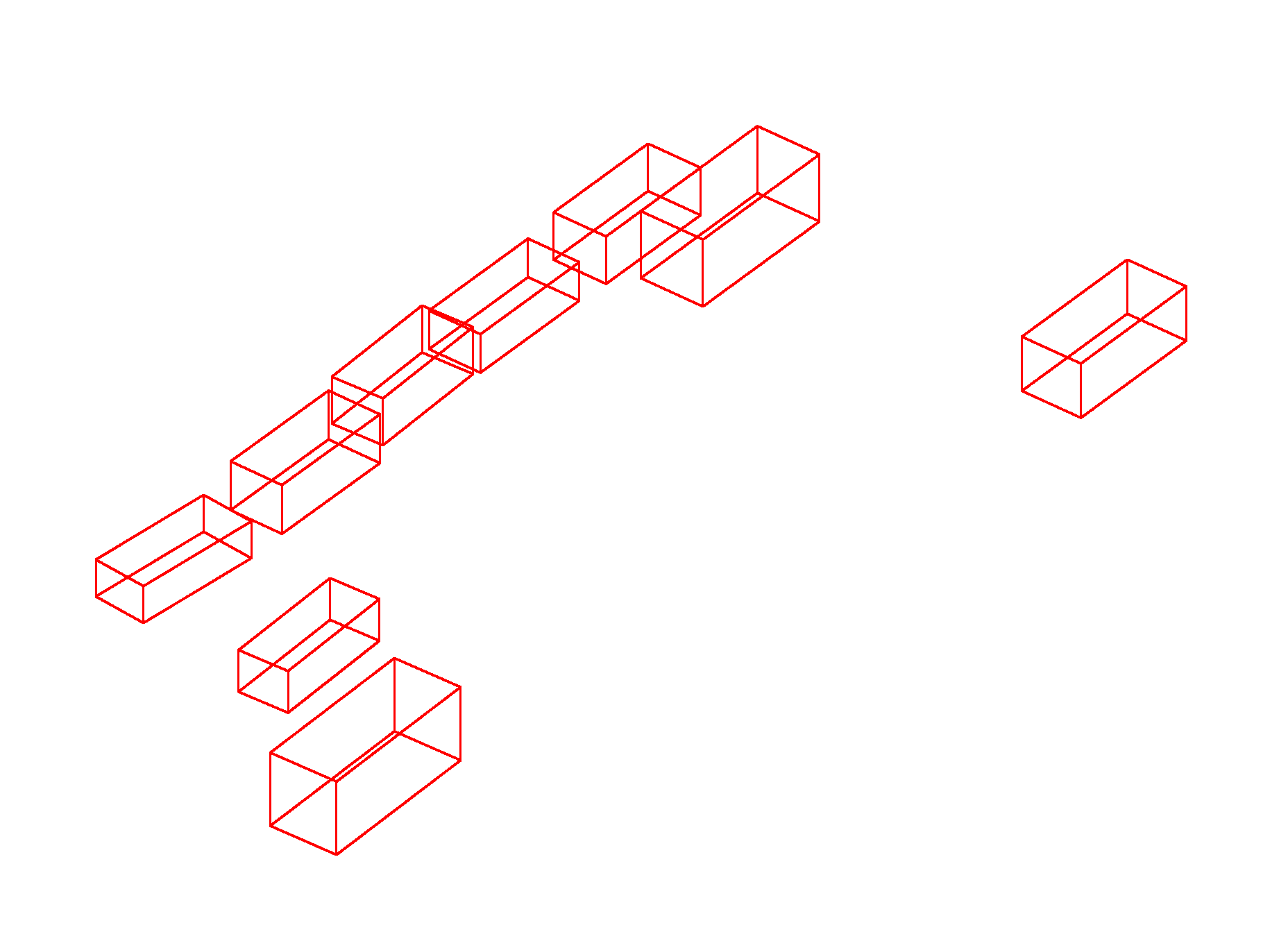} &
\tcimg{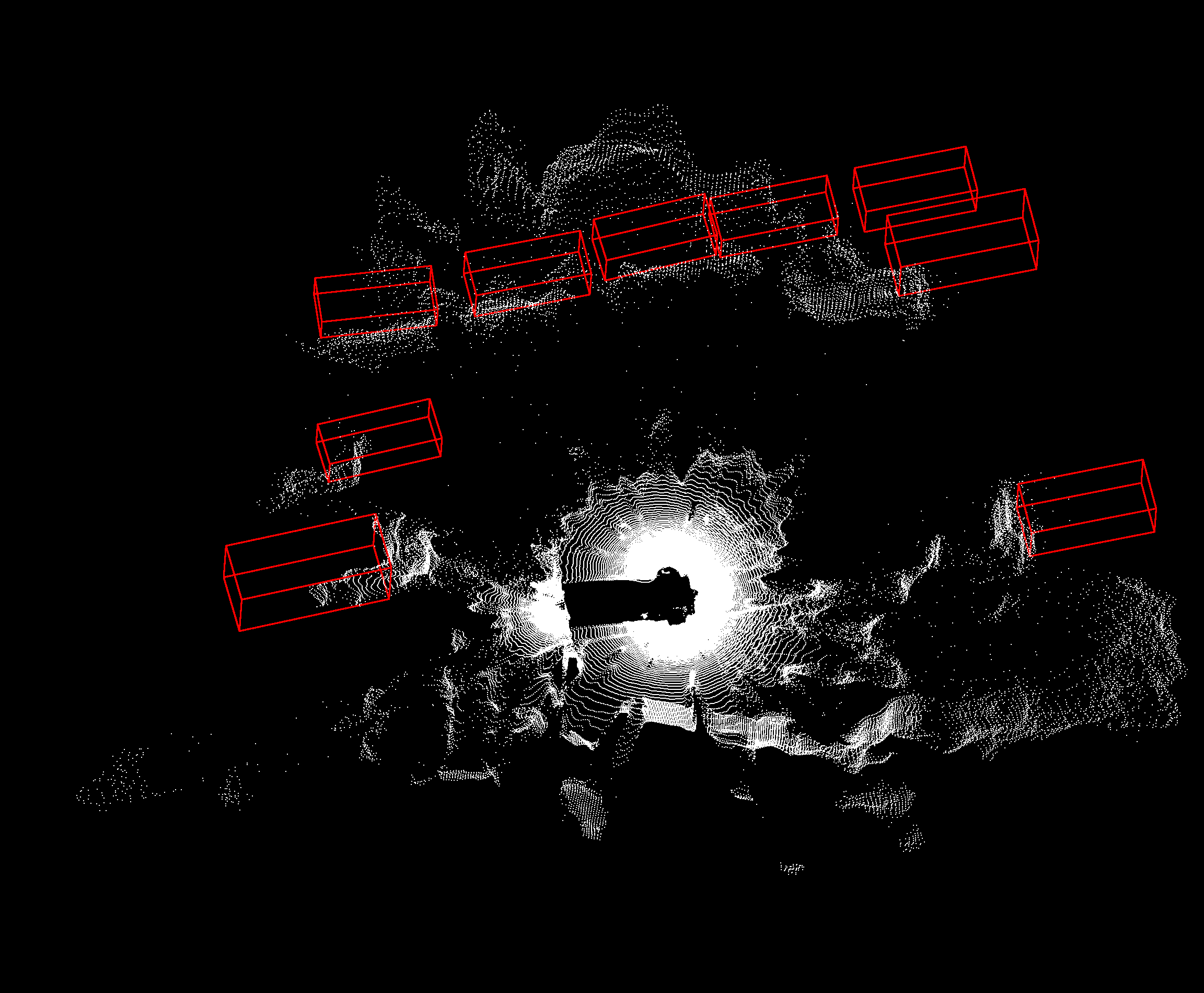} &
\tcimg{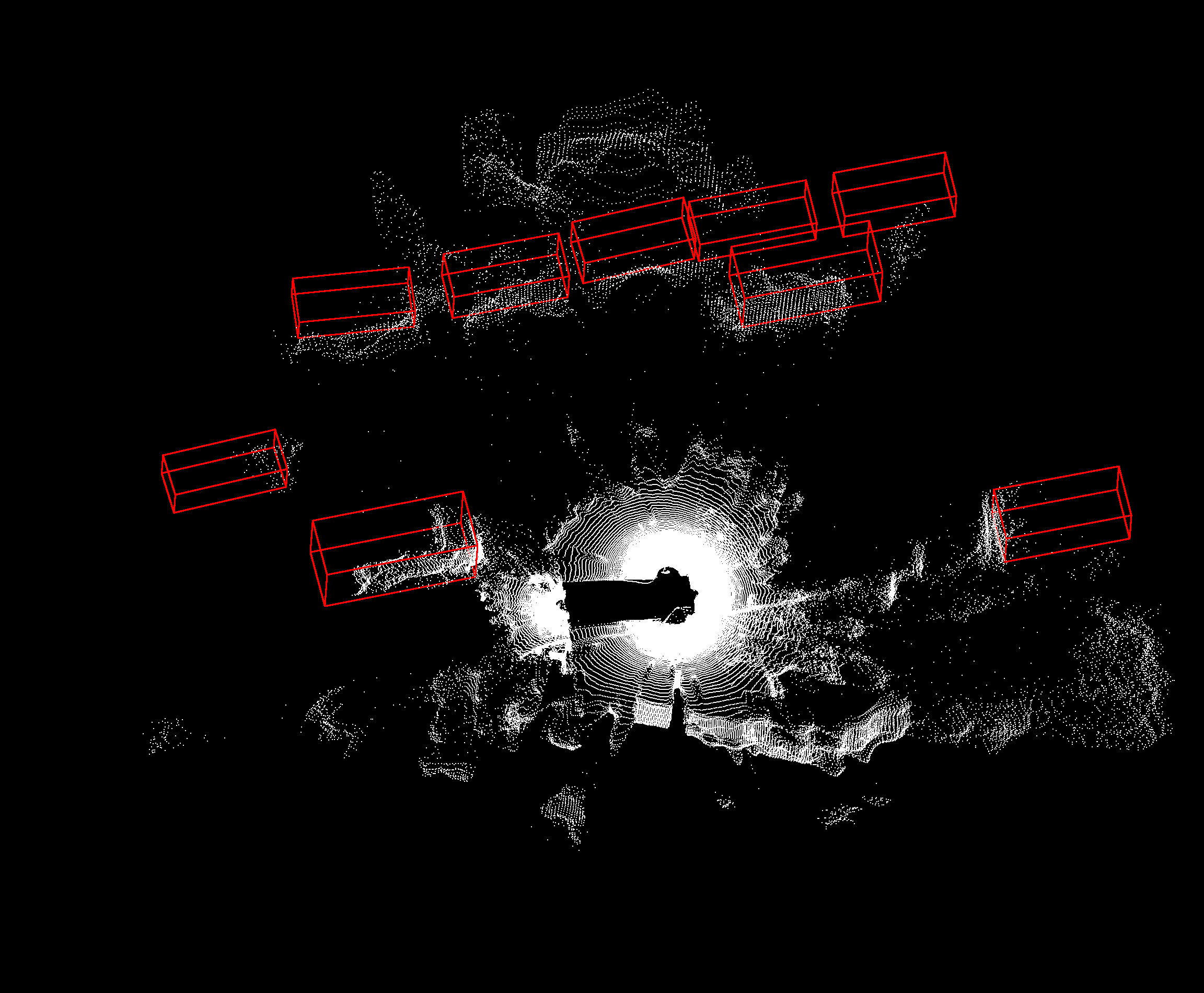} &
\tcimg{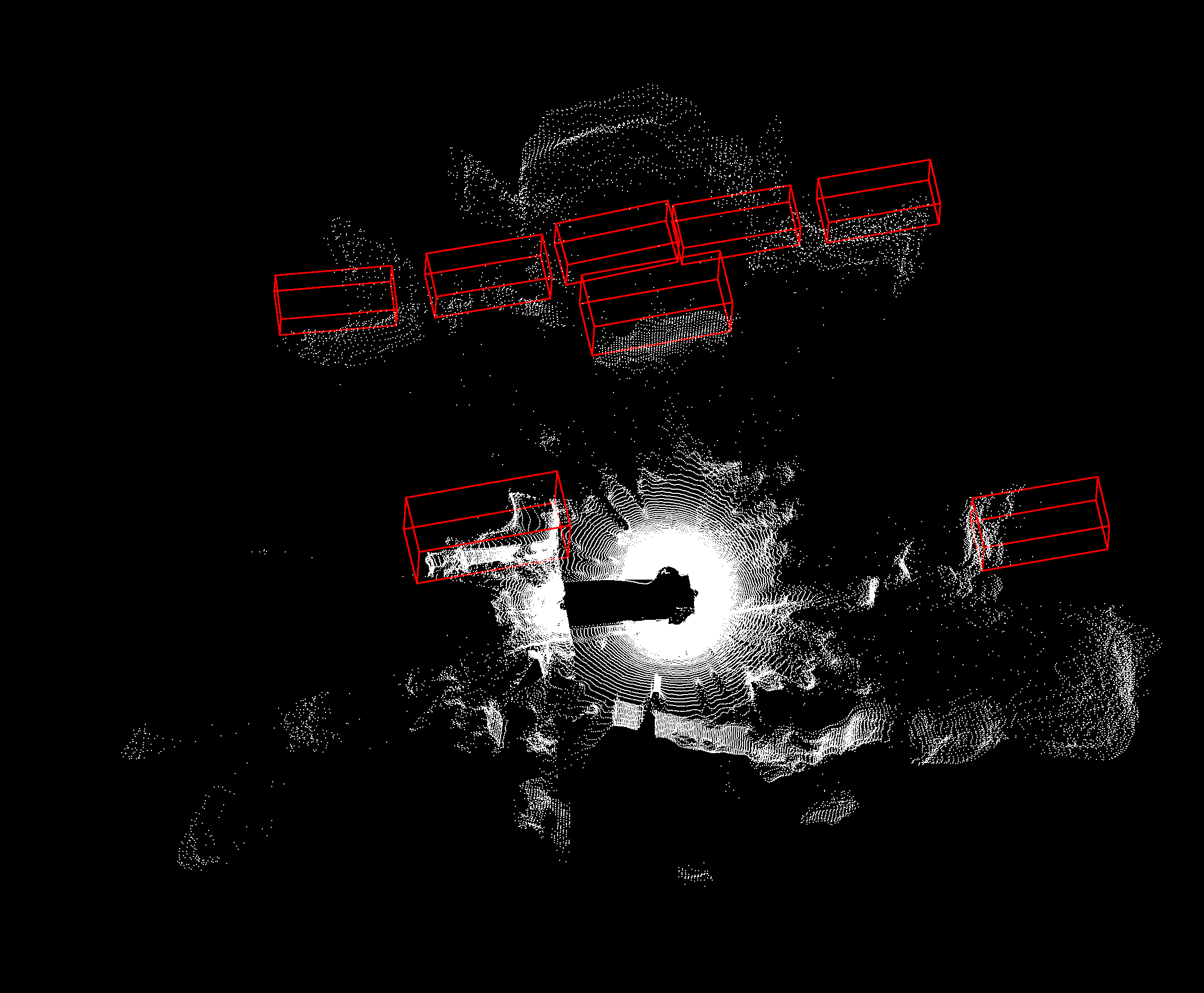}
\end{tabular}

\vspace{-0.5em}

\caption{\textbf{Temporal Consistency and Controllable Generation on Waymo.}
Qualitative results demonstrating both temporal coherence and controllable generation.
The first column shows BEV and 3D-box conditioning signals, while the remaining columns show generated LiDAR frames at different timesteps.
Across consecutive frames, \lname{} preserves stable object geometry and motion consistency while adhering to structural control signals for object placement and scene layout.}%
\label{fig:waymo_temporal_consistency}
\end{figure*}

\paragraph{Temporal Consistency.}
One of the key capabilities of our approach, enabled by the {\bf LiDAR Sequence Module}, is the modeling of long-range spatio-temporal dependencies within each sensor stream.
By jointly attending over spatial tokens across multiple time steps under a causal constraint, the module captures motion patterns and temporal context while preventing information leakage from future frames. This design enforces temporal coherence during generation and improves reasoning over evolving 3D structures.
As shown in Figure~\ref{fig:waymo_temporal_consistency}, the {\bf LiDAR Sequence Module} yields smoother predictions, more stable object representations over time, and enhanced dynamic scene understanding.

\paragraph{Controllable LiDAR Generation.} 
By conditioning the model on 3D bounding boxes and BEV maps through the {\bf ControlNet-style} branch, we enable precise object-level control during generation. These structural cues guide the generation, allowing the model to accurately place objects according to the specified layout.
As shown in Figure~\ref{fig:waymo_temporal_consistency}, this mechanism reliably regulates object presence and spatial arrangement, resulting in strong alignment between the input structural constraints and the generated LiDAR scenes.

\newpage

\end{document}